\documentclass[twoside,11pt]{article}

\usepackage{blindtext}
\usepackage{amsmath,amsthm}
\usepackage{amsfonts}
\usepackage{amssymb}
\usepackage[version=4]{mhchem}
\usepackage{stmaryrd}
\usepackage{bbold}
\usepackage{pifont}
\usepackage{algorithmic}
\usepackage{algorithm}
\usepackage{CJKutf8}
\usepackage{graphicx}
\usepackage{mathrsfs}
\usepackage{subfigure}

\usepackage{multirow}
\usepackage{booktabs}
\usepackage{float}
\usepackage{threeparttable}
\usepackage{subcaption}
\usepackage{natbib}
\usepackage{color}
\usepackage{tikz}
\usepackage[preprint]{jmlr2e}

\usepackage{lastpage}
\jmlrheading{}{2025}{1-\pageref{LastPage}}{5/25; Revised x/xx}{x/xx}{21-0000}{Qi Qin,  Erbo Li, Xingxiang Li, Yifan Sun, Wu Wang and  Chen Xu}

\ShortHeadings{LR-FFS for high-dimensional classification}{Qin, Li, Li, Sun, Wang and Xu}
\firstpageno{1}

\begin{document}

\title{Label-shift robust federated feature screening for high-dimensional classification}

\author{\name Qi Qin \email qin\_qi@ruc.edu.cn \\
       \addr Center for Applied Statistics and School of Statistics\\
       Renmin University of China\\
       Beijing, 100872, China
       \AND
       \name Erbo Li \email lear@ruc.edu.cn \\
       \addr Center for Applied Statistics and School of Statistics\\
       Renmin University of China\\
       Beijing, 100872, China
       \AND
       \name Xingxiang Li \email lxxwlm2013@xjtu.edu.cn \\
       \addr School of Mathematics and Statistics\\
       Xi'an Jiaotong University\\
       Xi'an, 710049, China
       \AND
       \name Yifan Sun \email sunyifan1984@163.com \\
       \addr Center for Applied Statistics and School of Statistics\\
       Renmin University of China\\
       Beijing Advanced Innovation Center for Future Blockchain and Privacy Computing\\
       Beijing, 100872, China
       \AND
       \name Wu Wang \email wu.wang@ruc.edu.cn \\
       \addr Center for Applied Statistics and School of Statistics\\
       Renmin University of China\\
       Beijing, 100872, China
       \AND
       \name Chen Xu \email cx3@xjtu.edu.cn \\
       \addr Department of Mathematics and Fundamental Research\\
    Peng Cheng Laboratory \\
    School of Mathematics and Statistics\\
       Xi'an Jiaotong University\\
       Xi'an, 710049, China}
\editor{My editor}

\maketitle
\begin{abstract}
Distributed and federated learning are important tools for high-dimensional classification of large datasets. To reduce computational costs and overcome the curse of dimensionality, feature screening plays a pivotal role in eliminating irrelevant features during data preprocessing. However, data heterogeneity, particularly label shifting across different clients, presents significant challenges for feature screening. This paper introduces a general framework that unifies existing screening methods and proposes a novel utility, label-shift robust federated feature screening (LR-FFS), along with its federated estimation procedure. The framework facilitates a uniform analysis of methods and systematically characterizes their behaviors under label shift conditions. Building upon this framework, LR-FFS leverages conditional distribution functions and expectations to address label shift without adding computational burdens and remains robust against model misspecification and outliers. Additionally, the federated procedure ensures computational efficiency and privacy protection while maintaining screening effectiveness comparable to centralized processing. We also provide a false discovery rate (FDR) control method for federated feature screening. Experimental results and theoretical analyses demonstrate LR-FFS's superior performance across diverse client environments, including those with varying class distributions, sample sizes, and missing categorical data. Supplementary materials are available online.
\end{abstract}


\section{Introduction}
\label{sec:introduction}
In light of recent advances in science and technology, high-dimensional data classification has become increasingly prevalent in scientific research and industrial applications \citep{fan_nonparametric_2011}. While the rapid expansion of data offers unprecedented opportunities, it also presents significant challenges \citep{fan_nonparametric_2011, fan2013mining, zhang2017multiple}:

\begin{enumerate}   
    \item \textbf{Privacy leakage}. In domains such as healthcare \citep{xu2021federated, brisimi2018federated}, data are often collected and maintained by institutions across various locations, referred to as \textit{nodes}. These datasets are highly sensitive, with strict regulations governing their use. For example, \citet{nguyen2024lightweight} utilized data from eight countries to predict sexually transmitted infections and human immunodeficiency virus, which are socially sensitive and stigmatized, highlighting significant security concerns. Even if personal information such as name and date of birth is deleted, the risk of privacy leakage still exists; for instance, a patient's faces can be reconstructed from computed tomography or magnetic resonance imaging data \citep{schwarz2019identification}. Consequently, data sharing or pooling is typically prohibited \citep{rieke2020future}.

    \item \textbf{Computational complexity}. Handling massive datasets poses significant computational challenges, as they are often too large to fit into computer memory and require significant processing time \citep{chen2020distributed, verbraeken2020survey, yu2022optimal}. 
    This issue is exacerbated when the computational capabilities of individual nodes,
    such as personal smartphones or laptops, are limited. For instance, training deep learning models on large-scale datasets such as ImageNet \citep{deng2009imagenet} can take several days, even with advanced hardware \citep{tang2020communication}. 

    \item \textbf{Data Quality}. Individual nodes often struggle with small data volumes and limited diversity, particularly in medical contexts such as imaging \citep{guan2024federated}, where the low incidence of certain diseases can restrict a single institution's ability to gather sufficient data \citep{prayitno2021systematic}. Cross-institution collaboration becomes essential in these cases, as seen in projects on brain tumor segmentation \citep{li2019privacy}, high-risk patient identification for postoperative gastric cancer recurrence \citep{feng2024robustly}, and crop disease detection \citep{mamba2023image}. Additionally, data within individual nodes are often prone to noise and outliers, which can degrade model performance. More critically, nodes may be targets of malicious or \textit{Byzantine attacks}, where compromised nodes intentionally send faulty data to disrupt the system \citep{fang_local_2020, yin_byzantine-robust_2018, jordan_communication-efficient_2018}. Therefore, robust data analysis methods are crucial in these contexts.

    \item \textbf{Statistical heterogeneity}. Heterogeneity refers to differences in data distributions across nodes. In real-world scenarios, this inherent heterogeneity in the data-generating process is widespread and can result from factors such as device variations and geographic differences. The impacts of this heterogeneity are well-documented in the literature, including issues such as unstable convergence \citep{li2020federated, mcmahan2017communication}, suboptimal model performance, and even negative outcomes \citep{li2022federated, luo2021no}.
\end{enumerate}

To leverage data from each node while ensuring data security, distributed processing and federated learning have emerged as suitable and increasingly popular approaches. In this framework, independent nodes that possess data, referred to as \textit{clients} (e.g., smartphones, hospitals), collaborate to train a global model \citep{mcmahan2017communication}. Clients communicate with a \textit{central server} (e.g., a service provider or project initiator) without sharing raw data, thereby protecting data privacy while maintaining the efficiency of statistical inference. This approach has been widely applied across various domains, including image recognition \citep{huang2022learn, shao2022federated}, medical diagnosis \citep{rieke2020future}, and wireless communication \citep{ wang2023distributed}.

When dealing with high-dimensional datasets, the computational cost is a major concern. Additionally, irrelevant features can lead to overfitting and spurious correlations. In high-dimensional data analysis, it is generally conceived that only a subset of features contributes significantly to the classification task. To mitigate computational complexity \citep{mwase2022communication, verbraeken2020survey}, it is crucial to screen out irrelevant features before conducting a formal federated analysis. This preprocessing strategy, known as \textit{feature screening} \citep{fan_ultrahigh_2009}, quantifies the relevance of each feature to the categorical response by a statistical measure, termed \textit{utility}, and subsequently removes those features with low utilities.

Traditional feature screening methods are broadly categorized into model-based and model-free procedures \citep{liu_selective_2015}. Notable model-based methods include feature annealing independence rules (FAIR) \citep{fan_high-dimensional_2008} and pairwise sure independence screening (PSIS) \citep{pan_ultrahigh-dimensional_2016}, while model-free feature screening encompasses techniques such as MV-SIS \citep{cui_model-free_2015}, fused Kolmogorov filter (FKF) \citep{mai_fused_2015}, and category-adaptive variable screening (CAVS) \citep{xie_category-adaptive_2020}. However, these methods assume that all data is stored on a single machine, making them unsuitable for distributed scenarios where communication bottlenecks exist between clients. To address this limitation, \citet{li_distributed_2020} pioneered a distributed feature screening framework based on aggregated correlation screening, allowing utility computation among clients without exchanging raw data. Building on this, \citet{li_feature_2023} proposed a robust distributed feature screening procedure based on conditional rank utility (CRU). Subsequent studies have further advanced the field by developing customized feature screening techniques tailored for distributed settings \citep{zhu_feature_2022, pang_distributed_2024, diao_distributed_2024}.

Although these model-free methods effectively address challenges such as heavy tails, noise, and outliers (Challenge 3), they fail to account for the impact of data distribution heterogeneity across different clients on screening results (Challenge 4).  \citet{kairouz2021advances, li2022federated} summarize various scenarios of heterogeneity, highlighting that differences in label distribution, commonly referred to as \textit{label shift} or \textit{label distribution skew}, are prevalent. Such disparities often arise due to factors such as individual preferences or geographic locations. For instance, pandas are primarily found in China, while kangaroos are primarily located in Australia. Real-world examples of class heterogeneity can be observed in street view data \citep{luo2019real} and natural geographic data \citep{hsu_federated_2020}. When analyzing data with label shift, it is generally assumed that features within the same class are homogeneous across clients, which holds true in areas such as cancer studies. For example, recent national and state-level U.S. data on cancer incidence for 2024 \citep{siegel2024cancer} show that lung cancer rates are three times higher in Kentucky, West Virginia, and Arkansas (75–84 per 100,000 persons) than in Utah (25 per 100,000 persons), reflecting historical differences in smoking rates. Similar differences are observed in cervical cancer and melanoma incidence. However, despite geographical variations, the features associated with a specific type of cancer remain consistent. In this paper, we propose a novel utility called \textit{Label-shift Robust Federated Feature Screening (LR-FFS)}, specifically designed to manage distributed data with potential label shifts, addressing a critical gap in the existing literature.

\begin{figure}[H]
	\centering
	\includegraphics[width=0.98\textwidth]{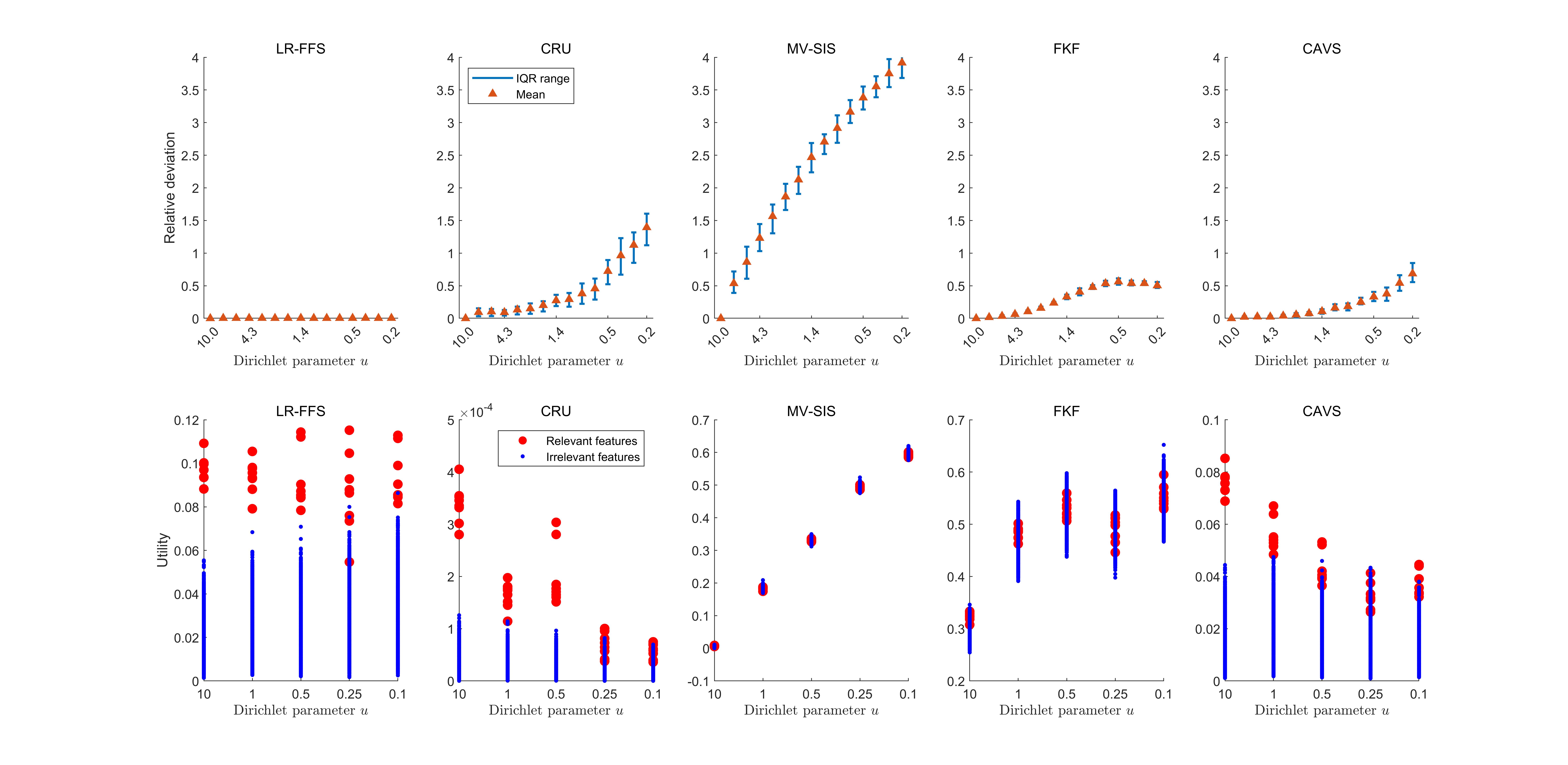}
	\caption{Impact of label shift on feature screening methods. First row: Relative deviation of utility estimates across heterogeneity levels. Second row: Utility distributions for relevant (red circles) and irrelevant (blue dots) features at selected heterogeneity levels.}    
	\label{Fig:toy example}
\end{figure}

To demonstrate the impacts of label shift on existing screening methods, we conduct a simulation study with $30$ clients, $10,000$ features, and five classes (detailed in Example \ref{exa:example2}, Setting (b)). The first eight features are relevant. We control the heterogeneity of $Y$ between clients using a Dirichlet distribution parameter $u$, where $u$ close to zero indicates increased heterogeneity. The first row of Figure \ref{Fig:toy example} presents the IQR (interquartile ranges) and mean values of relative deviations for relevant features, and the second row presents the utility distributions for both relevant (red triangles) and irrelevant (blue circles) features across five selected parameter values. The relative deviation is defined as the absolute value of the logarithmic difference $|\log(\hat{\omega}_{\text{distributed}}) - \log(\hat{\omega}_{\text{pooling}})|$, between distributed and pooled data estimates of utility values $\hat{\omega}$ obtained by different methods for the relevant feature $X$. The results show that conventional methods (MV-SIS, CRU, CAVS, FKF) exhibit a growing relative deviation in the utility estimates as heterogeneity increases ($u$ approaches 0.2), MV-SIS showing the most severe degradation followed by CRU. In contrast, our proposed LR-FFS maintains near-zero deviation across all heterogeneity levels while preserving stable separation between relevant (red circles) and irrelevant (blue dots) features---a capability that deteriorates sharply for other methods under high heterogeneity (Figure \ref{Fig:toy example}).

This performance divergence stems from fundamental limitations: label shift induces client-specific estimation bias that diverges from pooled optimal values, particularly when certain classes are underrepresented at individual clients (\cite{zhang2022federated}). Although PSIS and FAIR show label-shift robustness, their susceptibility to outliers (Challenge 3) limits practical applications. The complete breakdown of the original utility rankings (Figure \ref{Fig:toy example}, second row) confirms that existing methods cannot maintain reliable feature screening performance under label shifts, a critical weakness addressed by LR-FFS's design. Section \ref{sec:methodology} provides formal analysis of these phenomena and their implications for high-dimensional federated learning.

Our paper's main contributions are threefold. First, we propose a general distributed variable screening framework that unifies existing methods such as CRU, MV-SIS, and CAVS as special cases. This framework allows for a unified analysis and implementation of these methods, enabling the simultaneous study of their large-sample properties. 
Second, we introduce a novel utility, label-shift robust federated feature screening (LR-FFS), and the corresponding distributed estimation procedure for accurately quantifying the marginal importance of numerical features in classification problems with label-shift. In addition, we present a distributed framework algorithm for false discovery rate (FDR) control based on feature permutation. As detailed in Section \ref{subsec:lrffs}, we define a class utility for each classification level based on the conditional expectation of the conditional distribution, with LR-FFS representing the maximum value within this series of class utilities. This utility is model-free and insensitive to class distributions, outliers, and model misspecification. Even in the presence of label shift, each client shares the same estimation target, ensuring consistency between the aggregated and pooled results without compromising computational accuracy. The simple structure of LR-FFS facilitates distributive estimation using a natural unbiased estimator across clients, with one-shot aggregation enabling the derivation of global values while maintaining communication efficiency and data privacy. Third, we establish the convergence rates, sure screening properties, and FDR control properties for both the general screening framework and LR-FFS. The convergence rate of LR-FFS is comparable to that of estimators with access to all data across clients. Numerical examples further illustrate the robust performance of LR-FFS with finite samples.

The rest of this article is organized as follows: Section \ref{sec:methodology} analyzes the impact of label shift on existing feature screening methods in classification problems and proposes LR-FFS, along with its corresponding distributed procedure and FDR control procedure. Section \ref{sec:theorem} provides a theoretical demonstration of the estimation efficiency and robustness of LR-FFS and the general framework against label shift and outliers. Section \ref{sec:simulation} showcases the advantages of our method through numerical simulations and a real-world data example. Section \ref{sec:conclusion} concludes the paper, with theorem proofs and additional details provided in the Appendix.

\section{Methodology}\label{sec:methodology}
Before developing our methodology, we first provide background material and introduce key notations in Section \ref{subsec:bachgroud}. Section \ref{subsec:general} presents a general feature screening framework and analyzes the effects of label shift on the utility values of existing methods. Finally, Section \ref{subsec:lrffs} introduces the novel federated feature screening method, LR-FFS, and explains how it effectively addresses the issue of label shifting.

\subsection{Background and notations}\label{subsec:bachgroud}
Let $Y\in \left\{ y_1,\cdots ,y_R \right\}$ be a categorical response with $R$ classes, and let $ \boldsymbol{X}=\left( X_1,\cdots ,X_p \right) ^T$ be a vector of $p$ numerical features. Suppose the full dataset $\mathcal{D}$ is naturally partitioned into $m$ data segments $\{{\mathcal{D}_l}\}_{l=1}^m$, each residing on one of $m$ clients that process data separately and independently. The data segment $\mathcal{D}_l=\{(\boldsymbol{X}_i^l,Y_i^l)\}_{i=1}^{n_l}$ contains $n_l$ observations of $(\boldsymbol{X},Y)$, with the total number of observations across all clients given by $\sum_{l=1}^m n_l = N$, and where $n_l\ll p$.  Specifically, let $F(Y\mid \boldsymbol{X})$ denote the conditional distribution function of $Y$ given $\boldsymbol{X}$. We define the index set of relevant features across the clients as: 
$$\mathcal{A}=:\{1\le j \le p: F(y_r\mid \boldsymbol{X}) \text{ functionally depends on } X_j \text{ for some } r=1,\cdots, R\}$$
and the index set of irrelevant features as $\mathcal{I} = \{1,\cdots,p\} \backslash \mathcal{A}$.

Our goal is to screen out most irrelevant features with indices in $\mathcal{I}$, particularly focusing on scenarios with class heterogeneity among clients. For example, this is relevant when identifying pathogenic genes and establishing unified drug regimens, as discussed in Section \ref{sec:introduction}. To better articulate our research problem, our investigation is conducted under the following settings:

\begin{itemize}
    \item[S1] (Sparsity) Only a few features are relevant to the response variable, with $s=\left|\mathcal{A}\right|\ll p$, where $\left|\mathcal{A}\right|$ denotes the number of elements in the set $\mathcal{A}$.
    \item[S2] (Heterogeneity)  Suppose the $l$-th client has $n_l$ samples  $\{(\boldsymbol{X}_i^l,Y_i^l)\}_{i=1}^{n_l}$ drawn from the joint distribution $P_l(\boldsymbol{X},Y)=P_l(\boldsymbol{X}\mid Y)P_l(Y)$, where $P_l(\boldsymbol{X}\mid Y)$ is the conditional distribution function of $\boldsymbol{X}$ given $Y$ on the $l$-th client. The marginal distribution of the response, $P_l(Y)$, varies across clients, while the conditional distribution $P_l(\boldsymbol{X}\mid Y)$ remains constant across all clients, denoted simply as $P(\boldsymbol{X}\mid Y)$.
\end{itemize}
 
These settings relax the strict IID assumptions of data across clients, allowing the marginal distribution of the response to be heterogeneous across clients, i.e., exhibiting label shift. S1 is a common assumption in high-dimensional data analysis, where a significant portion of the data may consist of redundant information, necessitating feature screening or selection during the preprocessing stage. S2 defines label shift and imposes certain requirements on participating clients: only clients with the same conditional distribution of features can participate in the distributed system; otherwise, the effectiveness of feature screening may be compromised.

\subsection{A general framework for feature screening} \label{subsec:general}
To infer the relevance of $X_j$ to the response variable $Y$, existing utilities can take various forms but share profound underlying connections. Many screening utilities can be expressed as $\mathbb{E} \left( g\left( X_j, Y \right) \right)$ \citep{cui_model-free_2015,xie_category-adaptive_2020, li_distributed_2020,li_feature_2023}, where $g(X_j, Y)$ is a function typically related to $F(X_j)$, which is the distribution function of $X_j$. In a distributed framework, the impact of label shifting is subtle and complex. To illustrate this effect, we decompose the utility using the law of iterated expectations into two components for a specific category $y_r$:
\begin{align*}
\mathbb{E} \left( g\left( X_j,Y \right) \right)&=\sum_{y\in \left\{ y_1,\cdots ,y_R \right\}}^{}{\mathbb{E} \left( g\left( X_j,Y \right) \mid Y=y \right) P\left( Y=y \right)}\\
&=\mathbb{E} _{Y=y_r}\left( g\left( X_j,Y \right) \right) P\left( Y=y_r \right) +\mathbb{E} _{Y\ne y_r}\left( g\left( X_j,Y \right) \right) P\left( Y\ne y_r \right),
\end{align*}
where $\mathbb{E}_{Y=y_r} \left( g\left( X_j, Y \right) \right)$ and $\mathbb{E}_{Y\ne y_r} \left( g\left( X_j, Y \right) \right) $ denote the conditional expectations of $g(X_j, Y)$ given $ Y=y_r $ and $Y\ne y_r$, respectively. Specifically, we can further derive the following formula:
\begin{align*}
F\left( X_j \right)=F_{Y=y_r}\left( X_j \right) P\left( Y=y_r \right) +F_{Y\ne y_r}\left( X_j \right) P\left( Y\ne y_r \right). 
\end{align*}
where $F_{Y=y_r}(X_j) $ and $F_{Y\ne y_r}(X_j)$ represent the conditional distribution functions given $Y=y_r$ and $Y\ne y_r$, respectively.

These decompositions reveal how the proportion of  $Y = y_r$ affects the utility values expressed as expectations of $g(X_j, Y)$ or functions of $F(X_j)$. When label shift occurs, discrepancies between client-specific and overall target utility functions can result in estimation biases. Although we categorize the response values into $Y = y_r$ and $Y \neq y_r$ to illustrate this effect, the impact of label shifts on estimation can be more intricate, involving various combinations of response proportions across clients, resulting in $R^m$ types. This multitude of combinations poses significant challenges in practice.
Inspired by these decompositions, we propose a statistic based on conditional distributions and conditional expectations to mitigate this complexity and reduce the impact of class proportions. We integrate utilities that fit this form within a unified framework as follows:

\begin{equation}
\omega_j^{(d)}=\sum_{r=1}^R{\zeta_r {\omega_{j,r,d}^k}}, \label{equ:utility}
\end{equation}
where $\omega_{j,r,d}=\left| \mathbb{E}_{Y=y_r}\left( \left( F_{Y\ne y_r}\left( X_j \right) -F_{Y=y_r}\left( X_j \right) \right) ^d \right) \right| $ is the utility value for the $j$-th feature in category $Y=y_r$. Here, $d$ characterizes the order of the difference, $k$ is an exponent, and $\zeta_r$ are weight parameters typically related to the proportion of $Y=y_r$, which we denote by $\pi_r$. 

Similar to the Kolmogorov–Smirnov distance, the utility $\omega_{j,r,d}$ quantifies whether samples from $Y = y_r$ and $Y \neq y_r$ originate from the same distribution. Analogous to the KF and FKF, a small absolute difference $\left| F_{Y=y_r}(x) - F_{Y \ne y_r}(x) \right|$ suggests that $X_j$ is unrelated to $Y$. Proposition \ref{pro:equal} ensures that computing the expectation conditional on $Y \neq y_r$ yields identical results, thereby demonstrating the robustness of the proposed framework under label-shift conditions.

\begin{proposition}\label{pro:equal} 
	For any $d\ge 1$ and category $y_r$, the utility under two different probability measures $\mathbb{P}_{Y=y_r}$ and $\mathbb{P}_{Y \neq y_r}$ are the same:
	$$\mathbb{E}_{Y=y_{r}}\left[ F_{Y \neq y_{r}}(X)-F_{Y=y_{r}}(X)\right]^d = \mathbb{E}_{Y\ne y_{r}}\left[ F_{Y \neq y_{r}}(X)-F_{Y=y_{r}}(X)\right]^d.$$
\end{proposition}

By varying the parameters $k$, $d$, and the weights $\zeta_r$, many existing utilities can be included as special cases within our general framework, as the following proposition suggests.

\begin{proposition}\label{pro:special case}
	Existing utilities can be represented as special cases of the general framework:
	\begin{itemize}
		\item for CRU \citep{li_feature_2023}, set $\zeta_r = \left[P(Y=y_r)(1-P(Y=y_r))\right]^2 $ and $d=1, k=2$.
		\item for MV-SIS \citep{cui_model-free_2015}, set $\zeta_r = P(Y=y_r)(1-P(Y=y_r))^2$ and $d=2, k=1$.
		\item for CAVS \citep{xie_category-adaptive_2020}, set $\zeta_r = (1-P(Y=y_r))$ and $d=1, k=1$.
	\end{itemize}
\end{proposition}

As Proposition \ref{pro:special case} shows, different utilities employ different weights $\zeta_r$ to aggregate utility values from specific categories. For example, the weights for CAVS are $\zeta_r=1-P(Y=y_r)$, which emphasizes the utilities of categories with lower proportions. In contrast, CRU uses weights that are the square of the variance of $I(Y=y_r)$, favoring categories with proportions closer to $0.5$.
Regarding the order of the difference parameter $d$, \citet{cui_model-free_2015} investigated the second-order difference $d=2$, whereas \citet{li_feature_2023} and \citet{xie_category-adaptive_2020} focused on the first-order difference $d=1$, of the distributions. When $ d = 2$, the computational complexity for estimating $\omega_{j,r,2}$  using  U-statistics is typically $O(N^3 p)$, which is significantly higher than the complexity of estimating $\omega_{j,r,1}$, which is $O(N^2 p)$. For $d > 2$, the computational burden increases further. Therefore, to enhance computational efficiency, we focus on utilities with $d = 1$. For clarity, we denote $\omega_{j}$ and $\omega_{j, r}$ as utility values using the first-order difference $d=1$ in the following sections. The proofs of Proposition \ref{pro:special case} are provided in Appendix \ref{sec:conversion}.

As mentioned in Proposition \ref{pro:special case}, existing methods are closely tied to class proportions. In a distributed framework, when class proportions differ across clients due to label shift, and the original distributed estimation procedures are still applied, the effects of label shift become apparent. This leads us to the critical question: \textit{In the presence of label shifting, how can we ensure that different clients have the same goal, i.e., that the utility is insensitive to class proportions?}

\subsection{LR-FFS utility}\label{subsec:lrffs}
To better mitigate the impact of label shift, we adopt a special weight, $\zeta_r=I(\omega_{j, r}=\max_{r_1} \omega_{j, r_1})$, and propose a novel utility function, \textit{Label-shift robust federated feature screening} (LR-FFS). The utility of LR-FFS is formally defined as follows:
	\begin{equation*}
	\begin{split}
	\omega_j &=\max_r \omega_{j,r} =\max_r\left|\mathbb{E}_{Y=y_{r}}\left(F_{Y \neq y_{r}}\left(X_{j}\right)\right)-\mathbb{E}_{Y= y_{r}}\left(F_{Y = y_{r}}\left(X_{j}\right)\right)\right|\\ &= \max_r \left| \mathbb{E}_{Y=y_{r}}\left(F_{Y \neq y_{r}}\left(X_{j}\right)\right) - 1/2 \right|,
	\end{split}
	\end{equation*}
	where the third equality follows from $ \mathbb{E}_{Y= y_{r}}\left(F_{Y =y_{r}}\left(X_{j}\right)\right)=1/2 $. 
	
    LR-FFS's insensitivity to label shifting can be explained from two perspectives: the choice of statistics and the selection of coefficients. First, our statistic $\omega_{j}$ is derived from the conditional expectation of the conditional distribution function, which aids in identifying and mitigating the impact of label shifts. Proposition \ref{pro:consistent case} demonstrates that LR-FFS is robust to variations in the proportion of $Y = y_r$ under regular conditions. However, variations in the proportions of the remaining $R-1$ categories, other than $Y = y_r$ can still affect the utility value. To address this, we merge the remaining $R-1$ categories into a single category $Y \neq y_r$ and focus on the difference between the distributions of $X_j$ conditional on $Y = y_r$ and $Y \neq y_r$. Moreover, the weights in LR-FFS, specifically $\zeta_r = I(\omega_{j, r} = \max_{r_1} \omega_{j, r_1})$, are independent of category proportions, which minimizes the impact of label shifts. As discussed in Subsection \ref{subsec:general}, the weights used in existing feature screening methods, such as CRU and CAVS, are functions of category proportions, making them vulnerable to label shifts. In contrast, LR-FFS's design inherently mitigates this vulnerability by ensuring that the utility estimation remains consistent across clients, regardless of shifts in label distributions.

    \begin{proposition}\label{pro:consistent case} 
        When the proportion of categories other than $Y = y_r$ remains at a fixed ratio among the remaining $R-1$ categories, $\mathbb{E}_{Y=y_r}\left( F_{Y\ne y_r}\left( X_j \right) \right) $ is independent of the proportion of $Y = y_r$.
    \end{proposition}
    
    \begin{remark}
        In a simple three-class scenario, when the proportions of $Y = y_2$ and $Y = y_3$ are fixed at a certain ratio, $\omega_{j,1}$ is not influenced by the proportion of $Y = y_1$. This independence from the proportion of $Y=y_r$ extends more broadly under conditions where the conditional distributions of some categories are identical.
    \end{remark}

\begin{remark}

It is important to emphasize that our objective is to mitigate, rather than completely eliminate, the impact of label shift. This approach is consistent with principles found in \textit{client drift mitigation} \citep{karimireddy2020scaffold, li2020federated, durmus2021federated, luo2021no} in classical federated learning, where local objectives are adjusted to align local models more closely with the global model. This approach represents a delicate balance between reducing the impact of label shift and maintaining estimation accuracy.
To further explore this balance, we introduce an additional utility named LR-FFS-PAIR, defined as $\omega_j = \max_{r,k}\left|\mathbb{E}_{Y=y_{r}}\left(F_{Y = y_{k}}\left(X_{j}\right)\right)-\frac{1}{2}\right|$. LR-FFS-PAIR considers the maximum pairwise contrast between the distributions of $X_j$ under each pair of categories of $Y$. This method sacrifices some estimation accuracy by effectively reducing the sample size to better mitigate the impact of heterogeneity, while also introducing additional computational burden. Detailed results of this approach are presented in the supplementary materials for completeness.
 \end{remark}

To deepen the understanding of the proposed statistics, Section \ref{subsec:mwtest} explores the relationship between $ \omega_{j,r}=\left|\mathbb{E}_{Y=y_{r}}\left(F_{Y \neq y_{r}}\left(X_{j}\right)\right)-1/2\right|$ and the Mann–Whitney test.

\subsubsection{Connection to Mann-Whitney test}\label{subsec:mwtest}
First, we will delve into estimating the quantity $\gamma_{j,r} = \mathbb{E} _{Y=y_r}\left( F_{Y\ne y_r}\left( X_j \right) \right)$. Let a random sample $\{(\boldsymbol{X}_i,Y_i)\}_{i=1}^{N}$ of size $N$ be drawn from the population $\{(\boldsymbol{X},Y)\}$. Notice that:
\begin{equation}\label{equ:compare} 
\begin{aligned}
\mathbb{E} _{Y=y_r}\left( F_{Y\ne y_r}\left( X_j \right) \right) 
&=\int \mathbb{E}_{X_j^\prime}\left( I(X_j^\prime<X_j)\mid Y_j^\prime \ne y_r  \right)f(X_j\mid Y = y_r) \mathrm{d} X_j\\
&=\int I(X_j^\prime<X_j) f(X_j^\prime \mid Y \ne y_r)  f(X_j\mid Y = y_r) \mathrm{d} X_j\mathrm{d} X_j^\prime \\
&=P\left( X_{j,i_1}<X_{j,i_2}\mid Y_{i_1} \ne y_r,Y_{i_2}= y_r \right).  
\end{aligned}
\end{equation}

Referring to the transformation in Equation \ref{equ:compare} and defining $A_r=\left\{ j:Y_j\ne y_r \right\}$, $B_r=\left\{ j:Y_j=y_r \right\}$, $\gamma_{j,r}$ can be directly estimated by:
\begin{equation}\label{equ:estimate}
\hat{\gamma}_{j,r} =\frac{\sum_{i_1\in A_r}{\sum_{i_2\in B_r}{I\left( X_{j,i_1}<X_{j,i_2} \right)}}}{|A_r|\times |B_r|},
\end{equation}
where $|C|$  denotes the number of elements in the set $C$. $X_{j,i}$ denotes the $j$-th feature of the $i$-th sample. The estimator $\hat{\gamma}_{j,r}$ is essentially the Mann–Whitney statistic to test whether samples from $Y=y_r$ and $ Y \neq y_r$ come from the same distribution. When $X_j$ and $Y$ are statistically independent, straightforward computation yields:
$$\mathbb{E}\left(\hat{\gamma}_{j,r}\right)=\frac{\sum_{i_1\in A_r}{\sum_{i_2\in B_r}  \mathbb{E}\left({I\left( X_{j,i_1}<X_{j,i_2} \right)}\right)} }{|A_r|\times |B_r|}=\frac{1}{2}.$$

Therefore, the quantity $1/2$ in $\left|\mathbb{E}_{Y=y_{r}}\left(F_{Y \neq y_{r}}\left(X_{j}\right)\right)-1/2\right|$ offers another perspective: a significant deviation of $\gamma_{j,r}$ from $1/2$ indicates a stronger evidence of a relationship between $X_j$ and $Y$.
 
\subsection{Federated Feature Screening}\label{subsec:federated feature screening}
Following the problem setup in Section \ref{subsec:bachgroud}, we now discuss how to estimate the LR-FFS utility in the federated setting under possible label shift. Here, $\gamma_{j,r} =\mathbb{E}_{Y=y_r}\left(F_{Y \neq y_r}\left(X_j\right)\right)$ is the key statistic. To estimate $\gamma_{j,r}$, we adopt a one-shot aggregation (OSA) approach to obtain robust global estimates with minimal inter-machine communication costs \citep{huang2019distributed, li_feature_2023}.

We first decompose $\gamma_{j,r}$ into two components and estimate them separately: $\gamma_{j,r} = \frac{U_{j,r}}{\theta_r}$, where
\begin{align*}
&\theta_{r} =\pi_r(1-\pi_r)=  \mathbb{E}(I(Y_{i_2}=y_r)I(Y_{i_1}\ne y_r)),\\
&U_{j,r} =\gamma_{j, r}\pi_r(1-\pi_r)= \mathbb{E} \left( I\left( X_{j,i_1}<X_{j,i_2} \right) I\left( Y_{i_1} \ne y_r \right) I\left( Y_{i_2} =  y_r \right) \right).
\end{align*}

We begin with estimating $U_{j,r}$ first. Let $\mathcal{S}_l$ denote the index set of observations in $\mathcal{D}_l$. On the $l$-th client, $U_{j,r}$ can be estimated using a binary U statistic:
\begin{equation}
\hat{U}_{j,r}^l = \frac{\sum_{i_1\ne i_2 \in \mathcal{S}_l}{I\left( X_{j,i_1}<X_{j,i_2} \right) I\left( Y_{i_1} \ne y_r \right) I\left( Y_{i_2}= y_r \right)}}{n_l(n_l-1)}.
\end{equation}
Each client then transmits $\hat{U}_{j,r}^l$ to the central server, which aggregates $\hat{U}_{j,r}^l$ across clients using a weighted average $\bar{U}_{j,r} = \sum_l h_l \hat{U}_{j,r}^l / \sum_l h_l$, where $h_l = \lfloor n_l / 2 \rfloor$ is the effective sample size of the $l$-th client \citep{chen2021distributed, chen2023distributed}. It is crucial to note that $\bar{U}_{j,r}$ is a biased estimator of $U_{j,r}$ due to label shift, and obtaining unbiased estimators of $U_{j,r}$ based on local clients' data is not feasible. 

Fortunately, it is possible to correct this bias and design a consistent estimator of $\gamma_{j,r}$. Before delving into the details, we need to introduce some additional notations. Denote $\pi_r^l$ as the proportion of $Y = y_r$ on the $l$-th client and $\pi_r^*\in (0,1/2)$ as the solution to the following equation:
\begin{equation}
\sum_{l=1}^m{h_l }{\pi_r}^l\left( 1-{\pi_r}^l \right) =\sum_{l=1}^m{h_l}{\pi_r}^*\left( 1-{\pi_r}^* \right) \label{equ:bridge},
\end{equation}
which is a classic quadratic equation, and $\pi_r^*$ has a closed-form solution. We also denote $\theta_{r}^*= \pi_r^*(1-\pi_r^*)$ and $U_{j,r}^* = \mathbb{E}_{Y=y_r}\left( F_{Y\ne y_r}\left( X_j\right)\right)\pi_r^*(1-\pi_r^*)$. Note that $\gamma_{j,r} = U_{j,r}^*/\theta_{r}^*$. Some algebra shows that:
\begin{equation*}
\mathbb{E}\left(\bar{U}_{j,r} \right)=\mathbb{E}_{Y=y_r}\left( F_{Y\ne y_r}\left( X_j\right)\right)\frac{\sum_{l=1}^m h_l \pi_r^l(1-\pi_r^l)}{\sum_{l=1}^m h_l} = {U}_{j,r}^*.
\end{equation*}

Therefore, if we can consistently estimate $\theta_{r}^*$, we can correct the bias of $\bar{U}_{j,r}$ and obtain a consistent estimator of $\gamma_{j,r}$. Similar to the estimator $\bar{U}_{j,r}$, we estimate $\theta_r^*$ using a weighted U statistic $\bar{\theta}_r = \frac{\sum_{l=1}^m h_l \hat{\theta}_{r}^l}{\sum_{l=1}^m h_l}$, where
$$\hat{\theta}_{r}^l= \frac{\sum_{i_1\ne i_2 \in \mathcal{S}_l}{{I\left( Y_{i_1}\ne y_r\right)I\left( Y_{i_2}= y_r\right)}}}{n_l(n_l-1)}.$$
It is straightforward to check that $\mathbb{E}\left(\bar{\theta}_{r} \right) = \theta^*$. Lastly, we define the estimator of $\gamma_{j,r}$ as $\bar{\gamma}_{j,r} = \bar{U}_{j,r} / \bar{\theta}_r$.

Through this procedure, we can obtain an estimate of $\omega_{j}$. In practical implementation, we employ an equivalent but more interpretable algorithm for LR-FFS: each client $l$ uploads their local estimates $\hat{\gamma}_{j,r}^l$ with class-proportion-weighted sample size weights $\lambda_{l,r}$ to the server. The server then computes the global estimates $\bar{\gamma}_{j,r}$ and $\bar{\omega}_{j,r}$ through proper weighted aggregation, as detailed in Algorithm \ref{alg:procedure}. This modified implementation serves two key purposes: (1) it enhances interpretability of the federated learning process, and (2) it explicitly demonstrates the method's dual functionality for both componentwise estimation \citep{li_distributed_2020} and weighted statistical averaging. The mathematical equivalence between these algorithmic variants is formally established in Proposition \ref{ref:similar}.

\begin{proposition}\label{ref:similar}
To estimate the numerator of the aggregated parameters $\bar{\theta}_{r}$ and $\bar{U}_{j,r}$ we consider the aggregation of $\sum_{l=1}^{m} \lambda_{l,r} \hat{\gamma}_{j, r}^{l}/\sum_{l=1}^m \lambda_{l,r}$ from Step 2 in Algorithm \ref{alg:procedure}, specifically, $\sum_{l=1}^m h_l \hat{\theta}_{r}^l = \lambda_{l,r} \hat{\gamma}_{j, r}^{l}$ and $\sum_{l=1}^m h_l \hat{U}_{j,r}^l = \sum_{l=1}^m \lambda_{l,r}$.
\end{proposition}

\begin{algorithm}[H]
	\caption{Practical federated feature screening for LR-FFS.}
	\label{alg:procedure} 
	\renewcommand{\algorithmicrequire}{\textbf{Input:}}
	\renewcommand{\algorithmicensure}{\textbf{Output:}}
  \scriptsize{
	\begin{algorithmic}
		\REQUIRE $\{(\boldsymbol{X}_i^l,Y_i^l)\}_{i=1}^{n_l}$ %%input
		\ENSURE  the estimated screening utilities $\{\bar{\omega}_j\},j=1,\cdots,p$
		
		\FOR{each feature $j \in \{1,\cdots,p\}$ \textit{\textbf{in parallel}}}
		\FOR{each client $l \in \{1,\cdots,m\}$ \textit{\textbf{in parallel}}}
		\STATE \textbf{Step 1:} Client $C_l$ does:
		\FOR{each category $r \in \{1,\cdots,R\}$}
		\STATE  $\hat{\gamma}_{j, r}^l \leftarrow \frac{\sum_{i_1\in A_r^l}{\sum_{i_2\in B_r^l}{I\left( X_{i_1}<X_{i_2} \right)}}}{|A_r^l|\times |B_r^l|}$, where $A_r^l$ and $B_r^l$  represent the sets of $A_r$ and $B_r$ on the $l$-th client respectively defined in Equation \ref{equ:estimate}
		\STATE The weights calculated in the second step are obtained as: $\lambda _{l,r} \leftarrow \frac{\lfloor n_l/2 \rfloor}{n_l\left( n_l-1 \right)}|A_r^l||B_r^l|$
		\ENDFOR
		\STATE upload$_{C_l \rightarrow S}$ $\{\hat{\gamma}_{j, r}^l,\lambda _{l,r}\}$, $r=1,\cdots,R$
		\ENDFOR
		\STATE \textbf{Step 2:} Central Server $S$ does:
		\STATE $\bar{\omega}_j \leftarrow 0$
		\FOR{each category $r \in \{1,\cdots,R\}$}
		\STATE $\bar{\gamma}_{j,r} \leftarrow \sum_{l=1}^m (\hat{\gamma}_{j, r}^l \lambda_{l,r})/\sum_{l=1}^m \lambda _{l,r}$
		\STATE $\bar{\omega}_{j,r} \leftarrow |\bar{\gamma}_{j,r}-1/2|$
		\STATE $\bar{\omega}_{j}  \leftarrow \max\{\bar{\omega}_{j,r},\bar{\omega}_{j}\}$
		\ENDFOR
		\ENDFOR
		\RETURN $\{\bar{\omega}_{j}\}$, 	$j=1,\cdots,p$
	\end{algorithmic}
 }
\end{algorithm}

After aggregating the utility values of all the features at the central server, we screen features by retaining a set of key features where:
\begin{equation}
\hat{\mathcal{A}}=\left\{1 \leq j \leq p:\bar{\omega}_{j}>\delta \right\} ,\label{equ:threshod}
\end{equation}
where $\delta>0$ is a user-specified screening threshold.

\begin{remark}
    In the federated setting, beyond common issues such as outliers and noisy data, an extreme situation involves malicious client attacks. Although LR-FFS is not explicitly designed to prevent such attacks, it does not weight $\omega_{j,r}$ based on category proportions, thereby reducing the impact of errors introduced by malicious clients.  Furthermore, the aggregated nature of the $\omega_{j,r}$ estimator in Algorithm \ref{alg:procedure} supports the adoption of robust aggregation methods, such as the \textit{median of mean}, which can further enhance resilience against malicious attacks. This approach could be explored in future research.
\end{remark}

The computational complexity for each client in Algorithm \ref{alg:procedure} is $O(n_l^2 p)$, which is comparable to the corresponding step in \citet{li_feature_2023}. Therefore, addressing label shifts does not introduce additional computational burdens. The proposed LR-FFS framework ensures strong privacy preservation by relying solely on the exchange of highly processed summary statistics, eliminating any need for raw data sharing among clients.
 Table \ref{tab:table_advantage} summarizes the computational complexity, transmission cost, and robustness of existing methods. LR-FFS distinguishes itself with its simplicity and robustness, offering significant advantages in scenarios with large $N$ and $p$.

\begin{figure}[H]
	\centering
	\subfigure
	{
		\begin{minipage}[b]{.45\linewidth}
			\centering
			\includegraphics[scale=0.05]{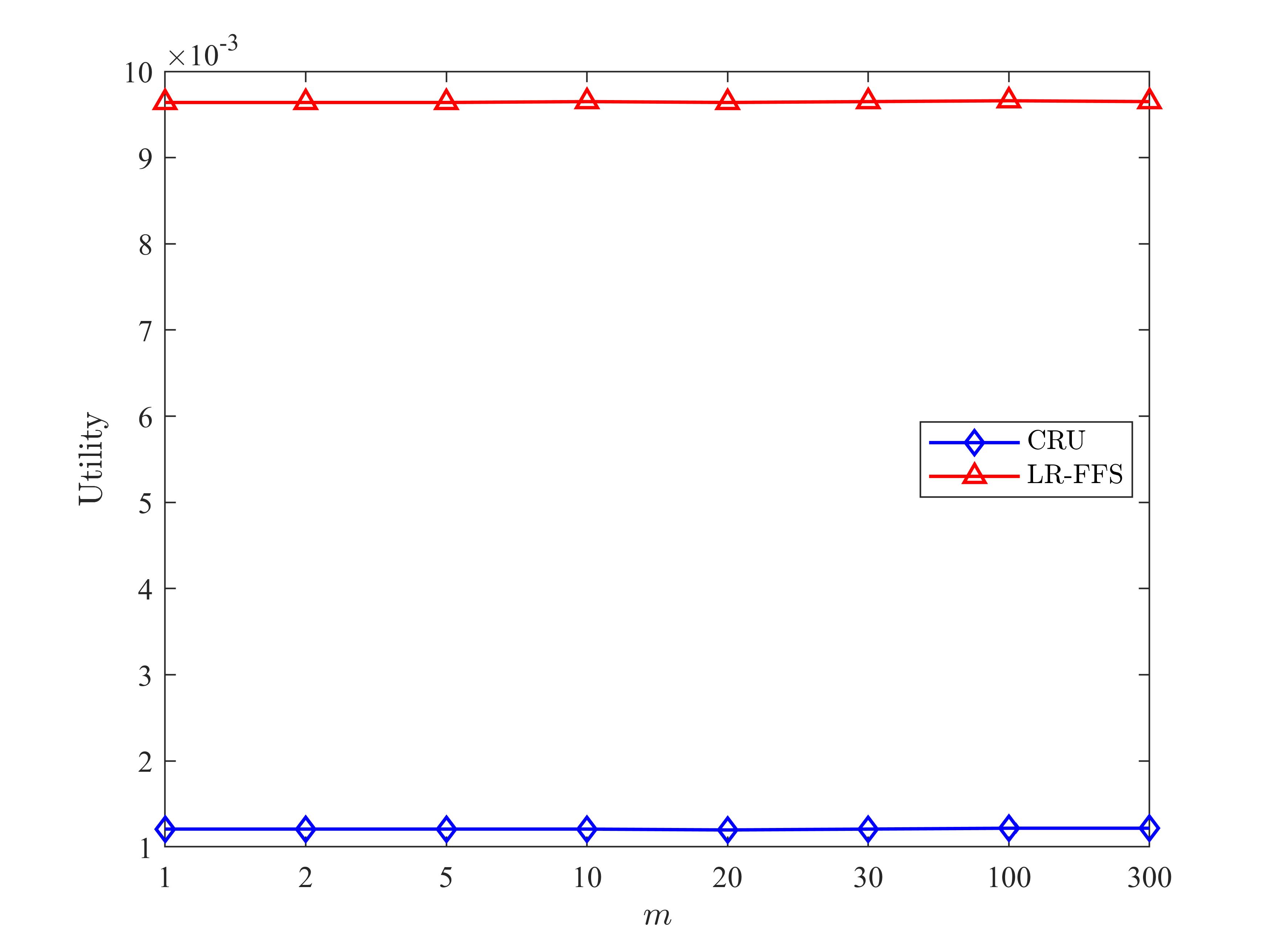}
		\end{minipage}
	}
	\subfigure
	{
		\begin{minipage}[b]{.45\linewidth}
			\centering
			\includegraphics[scale=0.05]{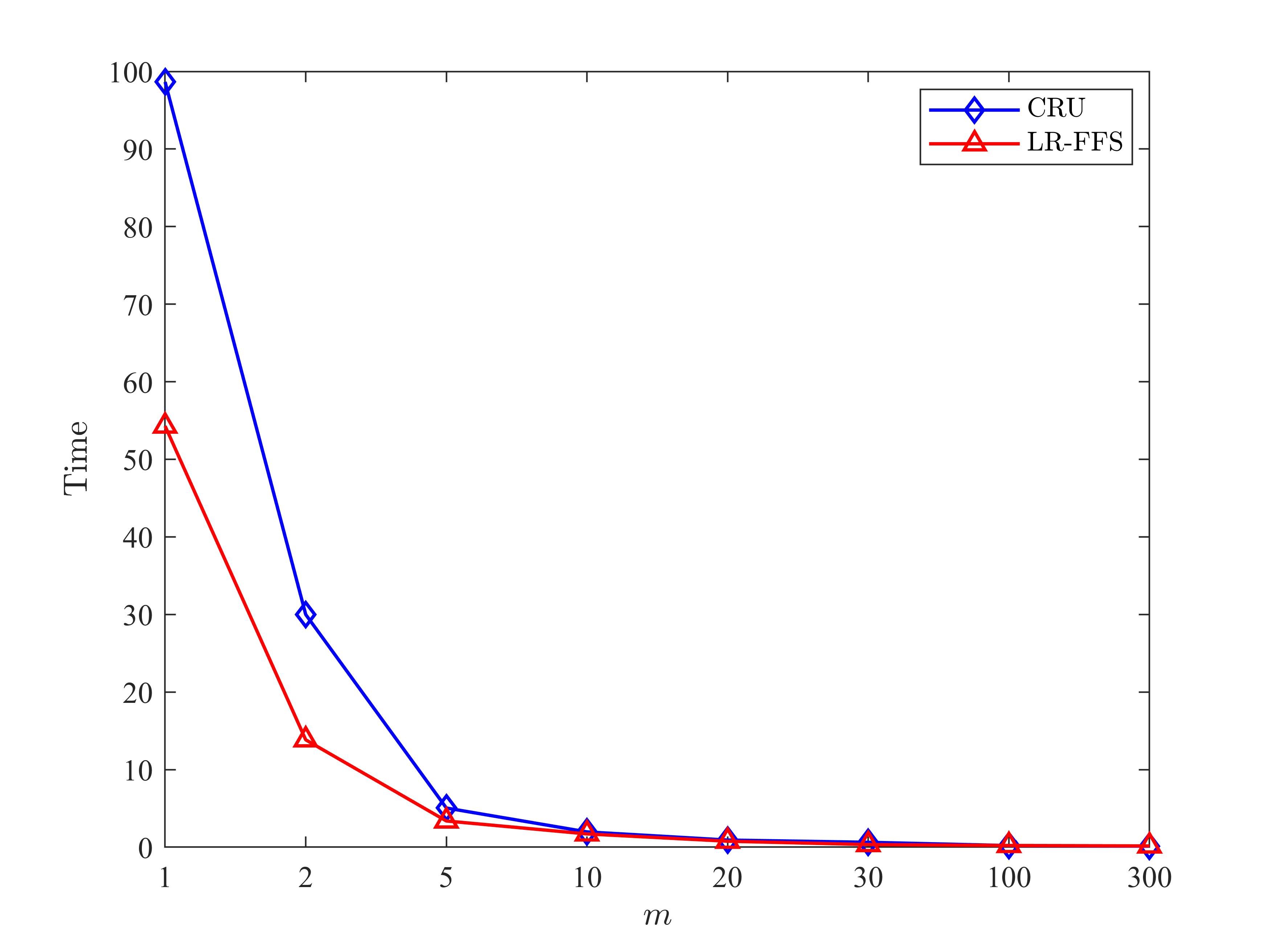}
		\end{minipage}
	}
	\caption{Simulation results for Example \ref{exa:example1}, left plot displays utility values while right plot shows their time consumption. The horizontal axis indicates the number of segments.}
	\label{fig:figure_time}
    
\end{figure}

Example \ref{exa:example1} demonstrates the computational efficiency of LR-FFS by partitioning the dataset into IID equally sized subsets. As shown in Figure \ref{fig:figure_time}, under this ``divide-and-conquer'' approach, computational efficiency significantly improves (right panel) while maintaining accuracy (left panel) as the number of partitions $m $ increases \footnote{The utility values cannot be directly compared across methods due to scale incompatibility.} . Furthermore, by estimating $ \omega_{j,r}$ within the framework \ref{equ:utility}, label-shift robust estimates can also be obtained based on other methods, as the estimation of $\pi_r$ or $\zeta_r$ remains unaffected by label shift. The corresponding algorithmic procedures are detailed in Appendix \ref{subsec:general_procedure}. Our MATLAB code repository is available on \url{https://github.com/Kee-Qin/LR-FFS}.

\begin{table}[!htbp]
	\centering
	\caption{OSA with common classification-based screening utilities.}
 \scriptsize{
	% \resizebox{\textwidth}{!}{
		\begin{tabular}{cccccc}
			\toprule
			\multicolumn{1}{c}{\multirow{2}[2]{*}{Utility}} & \multicolumn{1}{c}{\multirow{2}[2]{*}{OSA local complexity}} & \multicolumn{2}{c}{Robustness} & \multicolumn{1}{c}{\multirow{2}[2]{*}{Privacy-preservation}}&\multicolumn{1}{c}{\multirow{2}[2]{*}{Communication cost}} \\
			&       & \multicolumn{1}{c}{outliers} & \multicolumn{1}{c}{label shift} &  \\
			\midrule
			CRU   & $O(n_l^2p)$ & \ding{51}     & \ding{55}     & \ding{51}&$mR(p+1)$ \\
			FAIR   & $O(n_1p)$ & \ding{55}     & \ding{51}     & \ding{51}&$mR(2p+1)$ \\
			MV-SIS    &$O(n_l^3p)$& \ding{51}     & \ding{55}    & \ding{51}& $mR(2p+1)$ \\
			PSIS  & $O(n_lp)$ & \ding{55}     & \ding{51}     & \ding{51}& $mR(p+1)+mp$ \\
			FKF   & $O(n_lp)$ & \ding{51}     & \ding{51}     & \ding{55} & $>mR(Np+1)$ \\
			CAVS  & $O(n_l^2p)$ & \ding{51}     & \ding{55}      & \ding{51} &$mR(p+1)$\\
			LR-FFS & $O(n_l^2p)$ & \ding{51}     & \ding{51}     & \ding{51}&$mR(p+1)$\\
			\bottomrule
	\end{tabular}
 }
	\label{tab:table_advantage}%
\end{table}%

\begin{example}\label{exa:example1}
	We evaluate the distributed estimator by assessing its computational accuracy and efficiency. For this purpose, $N = 3000$ random copies of $(\boldsymbol{X},Y)$ are generated independently, where $Y\in \{1,2\}$ with $P(Y=1)=P(Y=2)=0.5$. Conditioned on $Y$, $X	\sim N(0.35,1)$ if $Y = 1$, and $X\sim N(0,1)$ if $Y = 2$. We partition the $N$ samples into $ m = (1, 2, 5, 10, 20, 30, 100) $ segments equally. The evaluation results are depicted in Figure \ref{fig:figure_time}.
\end{example}

\subsubsection{FDR control}\label{subsec:FDR_control}
In this subsection, we consider how to achieve more precise control over the FDR, referring to \citet{zhu2011model} and \citet{tong2023model} for the introduction of the FDR method. Specifically, for each feature $X_j$, we independently shuffle (permute) the data held by each client to create a ``pseudo'' feature $X_j^\prime$. Using the federated feature screening process, we compute the utilities for both $X_j$ and $X_j^\prime$, denoted as ${\omega}_j$ and ${\omega}_j^\prime$, respectively. Then, a new marginal utility $\phi_j$ that characterizes the relationship between $X_j$ and the response variable $Y$ can be defined as $\phi_j = {\omega}_j- {\omega}_j^\prime$.

The generated ``pseudo'' feature $X_j^\prime$ is independent of $Y$, thus the utility $\omega_j^\prime$ should be close to zero. If $X_j$ is a relevant feature, the value of $\phi_j$ should be significantly large; otherwise, it should be close to zero, with the probabilities of $\phi_j$ being positive or negative being approximately equal. This property is referred to as the \textit{marginal symmetry property} \citep{guo2023threshold}.

Given a threshold $\delta > 0$, we define the estimated set $\mathcal{A}$ as $\hat{\mathcal{A}}(\delta) = \{1\le j\le p : \hat{\phi}_j \geq \delta\}$ where $\hat{\phi}_j$ can be estimated by $\bar{\omega}_j- \bar{\omega}_j^\prime$. The FDP of $\hat{\mathcal{A}}(\delta)$ is then given by:
\begin{equation*}
    \textrm{FDP}[\hat{\mathcal{A}}(\delta)] = \frac{\left|\hat{\mathcal{A}}(\delta)\cap \mathcal{I}\right|}{\left|\hat{\mathcal{A}}(\delta) \right| \lor  1}=\frac{\left|\left\{j\in \mathcal{I}:\hat{\phi}_j \ge \delta\right\}\right|}{\left|\left\{j:\hat{\phi}_j \ge \delta\right\} \lor  1\right|},
\end{equation*}
where $a \lor b = \max\{a,b\}$. Then the FDR of $\hat{\mathcal{A}}(\delta)$ is $\textrm{FDR}[\hat{\mathcal{A}}(\delta)] = \mathbb{E}\{\textrm{FDP}[\hat{\mathcal{A}}(\delta)]\}$. In practice, since $\mathcal{I}$ is unknown, and inspired by the \textit{marginal symmetry property}, we have:
\begin{equation*}
    \left|\left\{j\in \mathcal{I}:\hat{\phi}_j \ge \delta\right\}\right|\approx \left|\left\{j\in \mathcal{I}:\hat{\phi}_j \le -\delta\right\}\right| \le \left|\left\{j:\hat{\phi}_j \le -\delta\right\}\right|.
\end{equation*}
To this end, the discussion leads to a conservative estimation of $\textrm{FDP}[\hat{\mathcal{A}}(\delta)]$ as follows:
\begin{equation*}
    \widehat{\textrm{FDP}}[\hat{\mathcal{A}}(\delta)] = \frac{\left|\left\{j:\hat{\phi}_j \le -\delta\right\}\right|}{\left|\left\{j:\hat{\phi}_j \ge \delta\right\} \lor  1\right|},
\end{equation*}
which motivates the threshold $\delta$ to be chosen by: 
\begin{equation}\label{equ:threshold_selection}
    \hat{\delta} = \inf\left\{\delta>0: \frac{1+\left|\left\{j:\hat{\phi}_j \le -\delta\right\}\right|}{\left|\left\{j:\hat{\phi}_j \ge \delta\right\} \lor  1\right|}<\alpha\right\}
\end{equation} 
under a pre-given level $\alpha$. The extra term $1$ in the numerator makes the choice of $\delta$ more conservative. Theorem \ref{the:FDRcontrol} provides a theoretical property about the estimated active set $\widehat{\mathcal{A}}$.

We employ permutation methods to construct ``pseudo'' features that are independent of $Y$ while preserving the same distribution as the original features. A related approach involves the construction of knockoff features \citep{barber2015controlling}, which ensures that these ``pseudo'' features are correlated with the original features to maintain exchangeability. This technique offers improved control over the FDR in feature selection. However, it cannot be directly applied to high-dimensional problems due to the requirement that $2p < n$. \citet{liu2022model} and \citet{pang_distributed_2024} have extended this approach to high-dimensional settings for non-distributed and distributed feature screening, respectively. They addressed the dimensionality constraint by initially screening down to $d$ features to ensure $2d < n_l $ before constructing knockoff features. Nevertheless, this method incurs additional computational costs. A notable drawback occurs when sample sizes are sparse across some clients; to satisfy the $2d < \min{n_l}$ condition, many relevant features may be excluded, which can be counterproductive.

\section{Theoretical Properties}\label{sec:theorem}
In this section, we analyze the asymptotic properties of the general variable screening framework and LR-FFS. To address issues arising from label shifts, we define $\vartheta_r = \frac{\pi_r^*(1-\pi_r^*)}{\pi_r(1-\pi_r)} $, which quantifies the degree of label shifts for category $y_r$, where $\pi_r^* $ is defined in Equation \ref{equ:bridge}. The interpretation of $\vartheta_r$ is provided in Remark \ref{rem:heterogeneity_factor}. The following conditions are necessary to ensure the sure screening and ranking consistency properties of the proposed procedure:
\begin{itemize}
	\item[(C1)] There exist three positive constants, $b_1$, $b_2$ and $b_3$ such that $b_1/R \le \min_r \pi_r \le \max_r \pi_r \le  1-b_2/R$, and  $\min_r\vartheta_r \ge b_3$.
    \item[(C2)] There exist positive constants $c>0$ and $0 \leq \kappa<1 / 2$ such that $\min _{j \in \mathcal{A}}\omega_{j} \geq 2 c N^{-\kappa}.$
	\item[(C3)] The number of classes $R=O(N^\xi)$, for some $\xi >0$, satisfying $\kappa +2\xi <\frac{1}{2}$.
	\item[(C4)] $\min _{j \in \mathcal{A}}\omega_{j}-\max _{j \in \mathcal{I}}\omega_{j} \geq 2 c N^{-\eta}$ for some $\eta \in\left(\kappa, \frac{1}{2}\right).$
\end{itemize}

Condition (C1) requires that the proportion of each category is neither too large nor too small, while also imposing restrictions on the degree of label shifts. This condition relaxes the IID requirements on the clients' data, allowing for scenarios where some clients may have very limited or missing data for certain categories, provided that other clients possess sufficient samples. For example, consider a scenario with three clients and three categories, where each client has equal sample sizes for the category it possesses. Specifically, client 1 has data only for categories 2 and 3, client 2 has data for categories 1 and 3, and client 3 has data for categories 1 and 2. In this setting, Condition (C1) is easily satisfied.

\begin{remark} \label{rem:heterogeneity_factor}
For category $y_r$, when there is no heterogeneity in this category, i.e., $\pi_r^1 = \cdots = \pi_r^m = \pi_r$, then $\vartheta_r = 1 $. When the data for category $r$ exists solely on one client, $\vartheta_r = 0$. As $\vartheta_r$ decreases, the degree of category heterogeneity among clients increases.
\end{remark}

Conditions (C2–C3) are similar to those in \citet{cui_model-free_2015} and \citet{xie_category-adaptive_2020}. Condition (C2) allows the minimum true signal to be on the order of $N^{-\kappa}$. Condition (C3) permits the number of classes for the response to diverge as $N$ increases. Condition (C4), which aligns with the setting in \citet{li_distributed_2020}, ensures that the active and inactive predictors can be well separated at the population level. 

\begin{proposition}\label{pro:omega_bound}
	Suppose Conditions (C1) and (C3) hold. For any constant $c_1>0$ and $r=1, \cdots, R$, there exists $c_2>0$ such that:
	\begin{equation}
	P\left(\max _{1 \leq j \leq p}\left|\bar{\omega}_{j, r}-\omega_{j, r}\right| \geq c_1 N^{-\kappa}\right) \leq 6p \exp \left( -c_2 N^{1-2\kappa-4\xi } \right),
	\end{equation}
\end{proposition}

Proposition \ref{pro:omega_bound} demonstrates that the estimator $\bar{\omega}_{j, r}$ is uniformly consistent, even if the number of features increases exponentially with the sample size, satisfying $\log(p) = O(N^\varrho)$ for some $\varrho\in \left(0,1-2\kappa-4\xi\right)$. The constant $c_2$ encapsulates information about the heterogeneity of category distributions and is positively related to  $\min_r\vartheta_r$. The error bound matches the efficiency of classic single-machine feature screening and is comparable to the efficiency of distributed feature screening in \citet{li_feature_2023}. Notably, label shift does not affect the convergence rate of the estimator. Under Condition (C3), there is a slight difference in the error bound compared to $O(N^{1-2\kappa-\xi})$ as reported in \citet{cui_model-free_2015} and \citet{xie_category-adaptive_2020}. However, this discrepancy becomes negligible when the sample size $N$ is sufficiently large.

\begin{proposition}\label{pro:variance}
    The variances of $\bar{\theta}_{r}$ and $\bar{U}_{j,r}$ can be expanded as
    \begin{align*}
        \max_{r} \mathrm{var}(\bar{\theta}_{r}) &= O(N^{-1})+O(m N^{-2}),\\
        \max_{j,r} \mathrm{var}(\bar{U}_{j,r}) &= O(N^{-1})+O(m N^{-2}).
    \end{align*}
Moreover, under the condition(C1), (C3) and $m = O(N)$, the mean squared error of $\bar{\omega}_{j,r}$ has the
following uniform order:
$$\max_{j,r}\mathrm{MSE}(\bar{\omega}_{j,r}) = \mathbb{E}(\bar{\omega}_{j,r}-{\omega}_{j,r})^2=O(N^{4\xi-1}).$$
\end{proposition}

Proposition \ref{pro:variance} confirms that $ \bar{\omega}_{j,r}$ attains the same mean squared error rate achievable by a centralized estimator, validating the efficacy of the federated approach. With Propositions \ref{pro:omega_bound} and \ref{pro:variance}, we derive a probability error bound for the estimator $\bar{\omega}_{j}$ in the following theorem.

\begin{theorem}[Sure screening property for LR-FFS]\label{the:surescreening}	

Following the notations and conditions of Proposition \ref{pro:omega_bound}, and if $\delta =cN^{-\eta}$ as in the threshold definition \ref{equ:threshod}, for any constant $c_3>0$, there exists $c_4>0$ such that:

    \begin{equation}
    P\left(\max _{1 \leq j \leq p}\left|\bar{\omega}_{j}-\omega_{j}\right| \geq c_3 N^{-\kappa}\right) \leq 6pR \exp \left( -c_4 N^{1-2\kappa-4\xi } \right),
    \end{equation}
    Moreover, under condition (C2), we have
    \begin{equation}
    P\left(\mathcal{A} \subset \hat{\mathcal{A}}\right) \geq 1-6sR \exp \left( -c_5 N^{1-2\kappa-4\xi } \right),
    \end{equation}
    where $c_5$ is some positive constant and $s=\left|\mathcal{A}\right|$ is the true model size.
\end{theorem}

\begin{theorem}[Ranking consistency property for LR-FFS]\label{the:rank}
Continuing with the assumptions of Theorem \ref{the:surescreening}, and additionally assuming Condition (C4) holds, there exists a constant $c_{6}>0$ such that:
    \begin{equation}
    P\left(\min_{j \in \mathcal{A}}\bar{\omega}_{j}>\max_{j \in \mathcal{I}}\bar{\omega}_{j}\right) \geq 1-6pR \exp \left( -c_{6} N^{1-2\eta-4\xi } \right).
    \end{equation}
\end{theorem}

\begin{theorem}[Controlling false discovery rate for LR-FFS]\label{the:FDRcontrol}	
Continuing with the assumptions of Theorem \ref{the:rank}, there exists a constant $c_{7}>0$ such that:
    \begin{equation}
P\left\{\left|\hat{\mathcal{A}}\right| \leq\left(c / 2\right)^{-1} N^\kappa \sum_{j=1}^p \omega_{j}\right\} \geq 1-6pR \exp \left( -c_{7} N^{1-2\kappa-4\xi } \right).
    \end{equation}
\end{theorem}

In Theorems \ref{the:surescreening}, and \ref{the:rank}, the minimal signal strength for $\omega_j$ aligns with the commonly used feature identifiability condition found in the literature. Our approach does not impose restrictions on the moments of the features, making it robust against heavy-tailed distributions. Compared to CRU, LR-FFS can accommodate heterogeneity in the response's distribution across clients. When the total sample size $N = \sum_{l=1}^m n_l $ is large, LR-FFS can eliminate most irrelevant features and retain all relevant ones with high probability, ensuring the sure screening property. Our convergence rate matches that of single-machine screening methods, demonstrating the distributed method's efficiency.

When Condition (C4) holds, a gap arises between the utilities of active and inactive features. We prove a theoretical result stronger than the sure screening property: when $\log(p)=o(N^{1-2\eta-4\xi})$, relevant features can be uniformly ranked above irrelevant ones through LR-FFS, with the probability tending to 1 (Theorem \ref{the:rank}). Consequently, there exists an ideal threshold to distinguish between active and inactive features.

Theorem \ref{the:FDRcontrol} shows that with high probability, the number of selected variables is bounded by $ O(N^\kappa \sum_{j=1}^p \omega_{j}) $. If $\sum_{j=1}^p \omega_{j}$ is of polynomial order in $N$, LR-FFS controls the number of selected features $\left|\hat{\mathcal{A}}\right|$ to be polynomial in $N$, even when $p$ grows exponentially. We prove that LR-FFS can control the FDR at a given threshold level $\delta =cN^{-\eta}$, where $c$ is a constant. However, determining an appropriate value of $\delta$ is not straightforward in practice. In Subsection \ref{subsec:FDR_control}, we provide a detailed procedure for distributed FDR control, with Theorem \ref{the:FDRalpha} offering theoretical guarantees for this procedure.

\begin{theorem}\label{the:FDRalpha}

For any $j\in \mathcal{I}$, define $\phi^*_j = \mathbb{I}(\phi_j<0)$. If there exists a sequence $c_n \to \infty$ as $n \to \infty$, such that $\mathbb{E}\phi^*_j = 0.5+o(c_n^{-1})$ and $c_n/p \to 0$ as $(n,p)\to \infty$, then for any $\alpha \in (0,1)$, the threshold $\hat{\delta} $ selected in Equation \ref{equ:threshold_selection} and corresponding estimated set  $\widehat{\mathcal{A}} = \{1\le j\le p : \hat{\phi}_j \geq \hat{\delta} \}$ satisfy:

    \begin{equation*}
        \textrm{FDR}[\widehat{\mathcal{A}}] = \mathbb{E}\left[\frac{\left| \mathcal{I} \cap \widehat{\mathcal{A}}\right|}{\left|\widehat{\mathcal{A}} \right|\lor  1}\right]\le \alpha + o(1).
    \end{equation*}
\end{theorem}

The assumptions of Theorem  \ref{the:FDRalpha} are the same as those of Theorem 2 in \citet{tong2023model}. Under mild conditions, it can effectively control the FDR at a given $\alpha$ level. These conditions require that the growth rate of $p$ is faster than $c_n$, which is easily satisfied in high-dimensional settings.

Next, we use Theorems \ref{the:surescreening_general} and \ref{the:rank_general} to justify the screening effectiveness of the general framework:
$$\bar{\omega}_j^{(d)} = \sum_{r=1}^R \bar{\zeta_r} \bar{\omega}_{j,r,d}^k.$$
When $ d > 1 $, the estimates of $\omega_{j,r,d} $, $ \bar{\omega}_{j,r,d} $, as well as the estimate of $ \zeta_r $, $\bar{\zeta_r}$, are provided in Appendix \ref{sec:general}. Specifically, when $ d = 1$, the estimate of $\omega_{j,r,d}$ or $\omega_{j,r}$ is given in Algorithm \ref{alg:procedure}.

\begin{theorem}[Sure screening property for the general framework]\label{the:surescreening_general}
 Suppose the number of classes $R$ is fixed and $\zeta_r$ is a continuous function of the category proportions. If $\delta =c N^{-\eta}$ in the threshold definition \ref{equ:threshod} and Condition (C1) holds, for any constant $c_8>0$, there exists $c_9>0$ such that:
	\begin{equation}
	P\left(\max _{1 \leq j \leq p}\left|\bar{\omega}_{j}^{(d)}-\omega_{j}^{(d)}\right| \geq c_8 N^{-\kappa}\right) \leq 12(2^d-1)pR\exp \left( -c_9 N^{1-2\kappa}\right),
	\end{equation}
	Moreover, under condition (C2), we have:
	\begin{equation}
	P\left(\mathcal{A} \subset \hat{\mathcal{A}}\right) \geq  1-12(2^d-1)sR\exp \left( -c_{10} N^{1-2\kappa}\right),
	\end{equation}
	where $c_{10}$ is a positive constant and $s=\left|\mathcal{A}\right| $ is the true model size.	
\end{theorem}

\begin{theorem}[Ranking consistency property for the general framework]\label{the:rank_general}
Assuming that the conditions of Theorem \ref{the:surescreening_general} and Condition (C4) hold, there exists a constant $ c_{11} > 0 $ such that:
	\begin{equation}
	P\left(\min _{j \in \mathcal{A}}\bar{\omega}_{j}^{(d)}>\max _{j \in \mathcal{I}}\bar{\omega}_{j}^{(d)}\right) \geq 1-12(2^d-1)pR \exp \left( -c_{11} N^{1-2\eta} \right).
	\end{equation}
\end{theorem}

From Theorems \ref{the:surescreening_general} and \ref{the:rank_general}, as $d$ increases, the error bounds tend to become increasingly loose. For $ d > 1 $, an analysis similar to Proposition \ref{pro:variance}, detailed in Proposition \ref{pro:general}, can be established. In addition, a larger $d$ will also result in a greater computational burden.

Comparing Theorems \ref{the:surescreening_general}, \ref{the:rank_general} with Theorems \ref{the:surescreening}, \ref{the:rank}, the general framework's flexibility in terms of the weights $\zeta_r$ and the power of $\omega_{j,r,d}$ may introduce biases during the estimation of the utilities.  Notably, when setting $d=1$ in Theorems \ref{the:surescreening_general} and \ref{the:rank_general}, the resulting order of the bound aligns with that of LR-FFS. However, the bounds provided in Theorems \ref{the:surescreening} and \ref{the:rank} are more precise.

\section{Numerical Studies}\label{sec:simulation}
\subsection{Simulations}
In this section, we investigate the numerical performance of the proposed LR-FFS procedure under possible label shifts. Example \ref{exa:example2} considers various feature distributions and heterogeneity settings. 
	
\begin{example}\label{exa:example2}
We generated $N$ random copies of $\left(\boldsymbol{X},Y\right)$ independently, where the categorical response $Y$ follows a distribution with $P\left( Y=r \right) ={\pi_r}^l$ on the $l$-th client, for $r = 1,\cdots, R$. 

For the $r$-th category, $p = 10,000$ features are generated by $$\boldsymbol{X}=\boldsymbol{\mu }_r+\boldsymbol{\varepsilon },$$
where $\boldsymbol{\mu}_r=\left(\mu_{r1},\cdots ,\mu_{rp} \right) ^T$ is a location parameter, and $\boldsymbol{\varepsilon} = \left(\varepsilon_1,\cdots,\varepsilon_p \right)^T$ is a random noise vector. A feature $X_j$ is considered irrelevant for classification when $\mu_{1j}=\cdots =\mu_{Rj}$.

\begin{itemize}
	\item[(a)] Set $n_l=100$ for $l=1,\cdots,30$. The distribution proportions $P\left( Y=r \right) ={\pi_r}^l$ across clients are determined by heterogeneous parameter $v$. The noise term $\boldsymbol{\varepsilon}$ independently follows a standard normal distribution $N(0,1)$. For $R=4,5,6,7$, $\mu_{1j}=0.28,0.30,0.32,0.34, 1\le j\le 8$ respectively and $\mu_{rj}=0$ elsewhere. The index set of relevant features is given by $\mathcal{A}=\{1,\cdots,8\}$.

    The proportion $\pi_r^l$ on the $l$-th client for each category $r$ is $\pi _r^l=\frac{\exp({\beta _r}^l)}{\sum{\exp({\beta _r}^l)}}$, where $\beta_r^l$ is a random number uniformly distributed on $(1,v)$. Increasing $v$ increases the degree of category heterogeneity. In this setting, we examine scenarios where $v$ varies from 1 (corresponding to IID label distribution) to 7 (exhibiting significant label shift).

	\item[(b)] Set $n_l=100$ for $l=1,\cdots,30$. The distribution proportions $P\left( Y=r \right) =\pi_r^l$ across clients follow a Dirichlet distribution with parameter $u$. The noise term $\boldsymbol{\varepsilon}$ independently follows a Student's $t$-distribution with 2 degrees of freedom. For $R=5,6,7$, $\mu_{11}=\cdots=\mu_{14}=\mu_{25}=\cdots=\mu_{28}=0.45,0.47,0.50$ respectively and $\mu_{rj}=0$ elsewhere. The index set of relevant features is $\mathcal{A}=\{1,\cdots,8\}$.
			
	\item[(c)] There are $16$ clients in total, and the clients are divided into 4 groups, with sample sizes of 100, 200, 300 and 400 in each group, respectively. The number of categories is $R=8$. We considered the case where some clients do not have data for certain categories. The number of missing categories for each client ranges from 0 to 4, while the remaining categories maintain the same relative proportion. The noise term $\boldsymbol{\varepsilon}$ independently follows a standard log-normal distribution (i.e., $\log(\varepsilon)\sim N(0,1)$). Set $\mu_{1j}=0.32, 1\le j \le 10$, $\mu_{2j}=0.08, 1\le j \le 10 $ and $\mu_{rj}=0$ elsewhere. The index set of relevant features is  $\mathcal{A}=\{1,\cdots,10\}$.

    \item[(d)] In this setting, we examine the effect of FDR control, employing the threshold selection process from Subsection \ref{subsec:FDR_control}. The heterogeneity ($u = 5$) and sample size settings are the same as in setting (a), with $R=5$ , $\mu_{1j}=0.4, 1 \le j \le 8$ and $\mu_{rj}=0$ elsewhere.
\end{itemize}
\end{example}

In the above setups: (a) considers normally
distributed $\boldsymbol{X}$ and is the most straightforward scenario for feature screening; (b) and (c) both investigate heavy-tailed distributions of $\boldsymbol{X}$, where (b) assumes that class distributions among different clients follow Dirichlet distributions, with increasing heterogeneity as $u$ decreases, whereas (c) considers the presence of missing class labels and varying sample sizes across clients; finally, (d) quantifies the effectiveness of our proposed FDR control mechanism under the threshold selection framework detailed in Subsection \ref{subsec:FDR_control}.

In each setup, we apply the LR-FFS procedure to screen irrelevant features distributively. For comparison, we also utilize existing classification-based utilities: CRU, PSIS, FKF, MV-SIS, and CAVS. These distributed algorithms for feature screening are based on \citet{li_distributed_2020}, and detailed algorithms can be found in Appendix \ref{sec:alg}. To simulate potential noise in the data, we randomly selected a total of $50$ samples from these clients and replaced all features with random numbers drawn from a uniform distribution ranging from 0 to 100.

To obtain a suitable threshold $\delta$ for Settings (a)-(c) while ensuring data privacy, we follow the strategy of \citet{zhu2011model, li_feature_2023}. Initially, we create a set of $q=1000$ auxiliary features $(Z_1, \ldots, Z_q)$ by permuting observed values of randomly selected features. Since the auxiliary features are unrelated to $Y$, we set the threshold $\delta = \max_{j=1,\cdots,q} \tilde{\omega}_{z,j}$, where $\tilde{\omega}_{z,j}$ is the OSA estimate of a screening utility between $Y$ and $Z_j$.

We evaluate screening accuracy using the successful screening rate (SSR), positive selection rate (PSR), and FDR over $T = 200$ repetitions:

\begin{align*}
    \textrm{SSR} = \frac{1}{T}\sum_{t=1}^T I(\mathcal{A} \subset \hat{\mathcal{A}}(t)), \textrm{PSR} = \frac{1}{T}\sum_{t=1}^T 
    \frac{\left|\mathcal{A} \cap \hat{\mathcal{A}}(t) \right|}{\left|\mathcal{A}  \right|},\textrm{FDR}= \frac{1}{T}\sum_{t=1}^T 
    \frac{\left| \hat{\mathcal{A}}(t) - \mathcal{A} \right|}{\left| \hat{\mathcal{A}}(t) \right|}
\end{align*}

where $\hat{\mathcal{A}}(t)$ denotes the index set of retained features in the $t$-th iteration. Additionally, we present the mean value of  $\left|\hat{\mathcal{A}}(t)\right|$ to indicate the number of retained features after screening (Size), along with the average of the largest rank of the relevant features in each simulation (wRank).

For each method, we also report the average computation time in seconds required by a local machine to perform distributed screening on a dataset. Table \ref{tab:table1} presents the results for all the performance measures across seven heterogeneity levels (shown in the second row), with $u=1$ indicating complete class homogeneity and $u=7$ reflecting extreme class heterogeneity. For clarity, the figures focus on SSR and wRank to aid comprehension. Complete simulation results are provided in the supplementary material.

\begin{table}[!htbp]
	\centering
	\caption{Case for $R=7$ in setting (a).}
 \begin{threeparttable}
	\resizebox{\textwidth}{!}{
    
	\begin{tabular}{cccccccccccccccccc}
		\toprule
		\multicolumn{9}{c}{Simulation without noise}   & \multicolumn{9}{c}{Simulation with noise} \\
		\midrule
		&  $v$     & 1     & 2     & 3     & 4     & 5     & 6     & 7     &       &    $v$     & 1     & 2     & 3     & 4     & 5     & 6     & 7 \\
        \midrule
		\multicolumn{1}{c}{\multirow{7}[1]{*}{SSR $\uparrow$}} & LR-FFS & \textbf{0.93} & \textbf{0.91} & \textbf{0.88} & \textbf{0.81} & \textbf{0.71} & 0.57  & 0.42  & \multicolumn{1}{c}{\multirow{7}[1]{*}{SSR$\uparrow$ \tnote{1} }} &LR-FFS & \textbf{0.90} & \textbf{0.85} & \textbf{0.83} & \textbf{0.70} & \textbf{0.58} & \textbf{0.46} & \textbf{0.30} \\
		& CRU   & 0.72  & 0.71  & 0.68  & 0.56  & 0.54  & 0.45  & 0.35  &       & CRU   & 0.65  & 0.66  & 0.65  & 0.49  & 0.39  & 0.35  & 0.28  \\
		& LR-FFS-PAIR & 0.73  & 0.66  & 0.45  & 0.30  & 0.07  & 0.03  & 0  &       & LR-FFS-PAIR & 0.68  & 0.61  & 0.45  & 0.19  & 0.08  & 0.01  & 0 \\
		& MV-SIS & 0     & 0     & 0     & 0     & 0     & 0     & 0     &       & MV-SIS & 0     & 0     & 0     & 0     & 0     & 0     & 0 \\
		& FKF   & 0     & 0     & 0     & 0     & 0     & 0     & 0     &       & FKF   & 0     & 0     & 0     & 0     & 0     & 0     & 0 \\
		& PSIS  & 0.84  & 0.83  & 0.82  & 0.79  & \textbf{0.71} & \textbf{0.67} & \textbf{0.62} &       & PSIS  & 0.01  & 0  & 0  & 0.01  & 0  & 0  & 0.01 \\
		& CAVS  & \textbf{0.93} & 0.90  & 0.85  & 0.73  & 0.60  & 0.41  & 0.24  &       & CAVS  & \textbf{0.90} & 0.84  & 0.81  & 0.64  & 0.49  & 0.32  & 0.18 \\
		\midrule
		\multicolumn{1}{c}{\multirow{7}[2]{*}{PSR$\uparrow$}} & LR-FFS & \textbf{0.99} & \textbf{0.99} & \textbf{0.98} & \textbf{0.97} & \textbf{0.95} & 0.92  & 0.87  & \multicolumn{1}{c}{\multirow{7}[2]{*}{PSR$\uparrow$}} & LR-FFS & \textbf{0.98} & \textbf{0.98} & \textbf{0.98} & \textbf{0.95} & \textbf{0.91} & \textbf{0.88} & \textbf{0.81} \\
		& CRU   & 0.95  & 0.95  & 0.95  & 0.90  & 0.88  & 0.80  & 0.76  &       & CRU   & 0.94  & 0.94  & 0.94  & 0.85  & 0.81  & 0.78  & 0.69 \\
		& LR-FFS-PAIR & 0.96  & 0.94  & 0.90  & 0.82  & 0.66  & 0.45  & 0.22  &       & LR-FFS-PAIR & 0.94  & 0.93  & 0.86  & 0.76  & 0.61  & 0.40  & 0.18 \\
		& MV-SIS & 0.08  & 0.01  & 0.01  & 0.01  & 0.01  & 0.01  & 0.01  &       & MV-SIS & 0.06  & 0.02  & 0  & 0.01  & 0.01  & 0.01  & 0 \\
		& FKF   & 0.05  & 0.02  & 0.01  & 0  & 0  & 0  & 0  &       & FKF   & 0.05  & 0.03  & 0.01  & 0  & 0  & 0  & 0 \\
		& PSIS  & 0.97  & 0.98  & 0.97  & 0.96  & 0.94  & \textbf{0.94} & \textbf{0.93} &       & PSIS  & 0.04  & 0.05  & 0.04  & 0.05  & 0.05  & 0.06  & 0.05 \\
		& CAVS  & \textbf{0.99} & 0.98  & 0.97  & 0.96  & 0.91  & 0.83  & 0.73  &       & CAVS  & \textbf{0.98} & \textbf{0.98} & 0.97  & 0.93  & 0.87  & 0.79  & 0.69 \\
		\midrule
		\multicolumn{1}{c}{\multirow{7}[2]{*}{FDR$\downarrow$\tnote{2}}} & LR-FFS & 0.44  & \textbf{0.42} & \textbf{0.44} & 0.45  & 0.43  & 0.43  & 0.45  & \multicolumn{1}{c}{\multirow{7}[2]{*}{FDR$\downarrow$}} &LR-FFS & 0.46  & \textbf{0.45} & \textbf{0.44} & \textbf{0.42} & \textbf{0.42} & \textbf{0.45} & \textbf{0.46}  \\
		& CRU   & 0.46  & 0.43  & 0.46  & 0.46  & 0.46  & 0.46  & 0.48  &       & CRU   & \textbf{0.44} & 0.46  & 0.45  & 0.46  & 0.47  & 0.47  & 0.53\\
		& LR-FFS-PAIR & 0.46  & 0.44  & 0.44  & 0.48  & 0.48  & 0.63  & 0.76  &       & LR-FFS-PAIR & 0.49  & 0.48  & 0.44  & 0.46  & 0.57  & 0.64  & 0.74 \\
		& MV-SIS & 0.81  & 0.89  & 0.91  & 0.86  & 0.90  & 0.94  & 0.88  &       & MV-SIS & 0.84  & 0.84  & 0.84  & 0.85  & 0.79  & 0.89  & 0.85 \\
		& FKF   & 0.89  & 0.89  & 0.90  & 0.92  & 0.90  & 0.94  & 0.92  &       & FKF   & 0.83  & 0.89  & 0.90  & 0.91  & 0.87  & 0.86  & 0.91 \\
		& PSIS  & \textbf{0.44} & 0.45  & 0.44  & 0.45  & \textbf{0.42} & \textbf{0.43} & \textbf{0.43} &       & PSIS  & 0.66  & 0.72  & 0.64  & 0.69  & 0.63  & 0.64  & 0.62 \\
		& CAVS  & 0.45  & 0.43  & 0.44  & \textbf{0.44} & 0.45  & 0.47  & 0.48  &       & CAVS  & 0.46  & 0.45  & 0.45  & 0.44  & 0.42  & 0.47  & 0.50 \\
		\midrule
		\multirow{7}[2]{*}{Size} & LR-FFS & 17.74 & 17.16 & 17.66 & 17.96 & 17.59 & 16.22 & 16.50 & \multirow{7}[2]{*}{Size} & LR-FFS & 18.60 & 18.58 & 18.29 & 16.19 & 16.65 & 17.14 & 16.48  \\
		& CRU   & 18.14 & 16.68 & 18.31 & 17.14 & 16.33 & 15.74 & 15.28 &       & CRU   & 18.09 & 18.56 & 17.24 & 16.54 & 17.61 & 15.64 & 15.85\\
		& LR-FFS-PAIR & 17.97 & 17.79 & 16.91 & 17.42 & 13.55 & 14.89 & 11.83 &       & LR-FFS-PAIR & 19.20 & 19.25 & 16.64 & 15.51 & 17.16 & 14.32 & 11.05 \\
		& MV-SIS & 10.96 & 10.70 & 10.67 & 7.89  & 9.32  & 10.21 & 9.22  &       & MV-SIS & 8.79  & 18.00 & 20.87 & 21.30 & 23.02 & 16.65 & 20.06 \\
		& FKF   & 10.65 & 11.58 & 10.42 & 10.11 & 9.95  & 10.38 & 10 &       & FKF   & 9.10  & 10.36 & 11.65 & 10.93 & 10.36 & 13.37 & 15.87 \\
		& PSIS  & 17.26 & 17.80 & 17.87 & 18.11 & 16.77 & 16.93 & 16.88 &       & PSIS  & 265.46 & 398.36 & 245.75 & 319.75 & 289.17 & 303.87 & 211.69 \\
		& CAVS  & 17.82 & 17.00 & 17.54 & 17.70 & 17.53 & 15.58 & 14.65 &       & CAVS  & 18.64 & 18.66 & 18.45 & 16.72 & 16.27 & 16.44 & 15.57 \\
		\midrule
		\multicolumn{1}{c}{\multirow{7}[2]{*}{wRank$\downarrow$}} & LR-FFS & 9.32  & \textbf{12.60} & \textbf{12.19} & 20.72 & 27.89 & 50.45 & 112.40 & \multicolumn{1}{c}{\multirow{7}[2]{*}{wRank$\downarrow$}} & LR-FFS & 13.01 & \textbf{13.14} & \textbf{17.29} & \textbf{41.40} & \textbf{80.51} & \textbf{71.10} & \textbf{155.81} \\
		& CRU   & 17.85 & 25.19 & 26.69 & 44.86 & 123.44 & 207.42 & 308.19 &       & CRU   & 37.03 & 33.60 & 58.29 & 118.79 & 256.09 & 221.66 & 419.92 \\
		& LR-FFS-PAIR & 18.44 & 27.73 & 38.50 & 95.90 & 250.81 & 614.37 & 3258.99 &       & LR-FFS-PAIR & 25.04 & 31.83 & 66.86 & 147.75 & 375.01 & 798.82 & 3786.20 \\
		& MV-SIS & 4120  & 7191  & 7851  & 8037  & 8236  & 8113  & 8225  &       & MV-SIS & 4830  & 7132  & 7685  & 8150  & 8203  & 8065  & 8262 \\
		& FKF   & 5155  & 5864  & 7200  & 8103  & 8405  & 8784  & 8641  &       & FKF   & 5378  & 6061  & 7179  & 8058  & 8660  & 8556  & 8538 \\
		& PSIS  & 13.04 & 17.02 & 13.69 & \textbf{15.96} & \textbf{14.35} & \textbf{21.60} & \textbf{26.55} &       & PSIS  & 8226.00 & 8183.00 & 8085.00 & 8179.00 & 7949.00 & 7819.00 & 7926.00 \\
		& CAVS  & \textbf{9.27} & 12.86 & 12.54 & 23.36 & 44.64 & 84.15 & 179.14 &       & CAVS  & \textbf{12.90} & 13.16 & 18.25 & 45.37 & 85.16 & 123.64 & 235.32 \\
		\midrule
		\multicolumn{1}{c}{\multirow{7}[2]{*}{Time$\downarrow$}} & LR-FFS & 0.71  & 0.72  & 0.72  & 0.71  & 0.71  & 0.69  & 0.69   & \multicolumn{1}{c}{\multirow{7}[2]{*}{Time$\downarrow$}} & LR-FFS & 0.74  & 0.75  & 0.73  & 0.72  & 0.71  & 0.70  & 0.69 \\
		& CRU   & 0.67  & 0.66  & 0.67  & 0.67  & 0.67  & 0.67  & 0.67  &       &CRU   & 0.69  & 0.69  & 0.67  & 0.68  & 0.68  & 0.68  & 0.68  \\
		& LR-FFS-PAIR & 1.88  & 1.90  & 1.91  & 1.91  & 1.88  & 1.86  & 1.86  &       & LR-FFS-PAIR & 1.94  & 1.95  & 1.91  & 1.90  & 1.86  & 1.86  & 1.85 \\
		& MV-SIS & 18.18 & 18.19 & 18.19 & 18.19 & 18.18 & 18.25 & 18.19 &       & MV-SIS & 18.30 & 18.30 & 18.22 & 18.21 & 18.23 & 18.25 & 18.20 \\
		& FKF   & 2.38  & 2.42  & 2.46  & 2.48  & 2.47  & 2.40  & 2.24  &       & FKF   & 2.46  & 2.48  & 2.44  & 2.43  & 2.44  & 2.38  & 2.21 \\
		& PSIS  & \textbf{0.29} & \textbf{0.29} & \textbf{0.29} & \textbf{0.29} & \textbf{0.29} & \textbf{0.30} & \textbf{0.30} &       & PSIS  & \textbf{0.30} & \textbf{0.30} & \textbf{0.29} & \textbf{0.29} & \textbf{0.29} & \textbf{0.30} & \textbf{0.30} \\
		& CAVS  & 0.77  & 0.76  & 0.78  & 0.79  & 0.79  & 0.77  & 0.76  &       & CAVS  & 0.79  & 0.78  & 0.79  & 0.79  & 0.78  & 0.77  & 0.75 \\
		\bottomrule
	\end{tabular}}
          \begin{tablenotes} 
        \footnotesize
        \item[1] In the results presented, an upward arrow indicates that a higher value is preferable.
        \item[2] A downward arrow signifies that a lower value is better.
      \end{tablenotes}
      
 \end{threeparttable}
 \label{tab:table1}
\end{table}

Table \ref{tab:table1} shows that both MV-SIS and FKF are not suitable for addressing the distributed screening problem, exhibiting poor performance and high computational costs. Under homogeneous settings ($v=1$, no noise), all five alternative methods perform comparably well, consistently ranking the eight truly important features within the top 20 candidates—a result that allows reliable selection using conservative thresholds. However, the introduction of noise reveals remarkable differences in robustness: while the model-based PSIS suffers rapid performance degradation, model-free methods demonstrate strong resilience to outliers and noise, maintaining stable screening accuracy with only minor declines.

As the degree of heterogeneity increases ($v>1$), all methods exhibit performance deterioration, though PSIS shows a slight advantage in noiseless settings for 
$v>5$. This advantage, however, disappears entirely under noisy conditions. Among model-free approaches, LR-FFS consistently outperforms its counterparts, maintaining superior feature ranking and successful screening rates. Crucially, as shown in Table \ref{tab:table_advantage}, LR-FFS achieves this robustness without introducing additional computational overhead to handle label shift. In contrast, LR-FFS-PAIR’s explicit handling of label shifts comes at the cost of reduced accuracy, rendering it less effective than methods that do not prioritize label-shift correction. Given these findings, we focus subsequent analyses on PSIS, CRU, and CAVS for comparative evaluation.

\begin{figure}[H]
	\centering
	\subfigure[$R=5$]
	{
		\begin{minipage}[b]{.3\linewidth}
			\centering
			\includegraphics[scale=0.195]{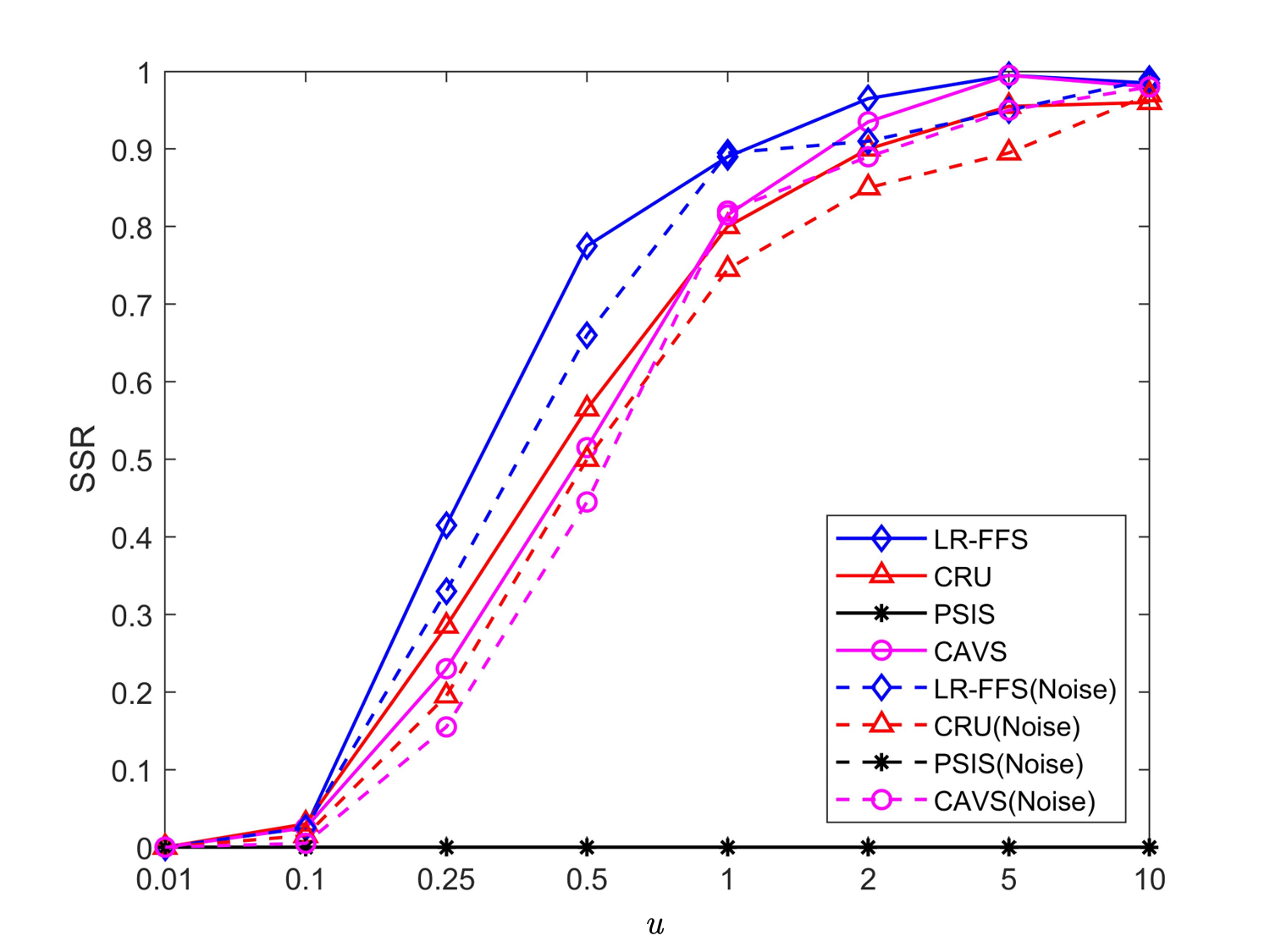} \\
			\includegraphics[scale=0.195]{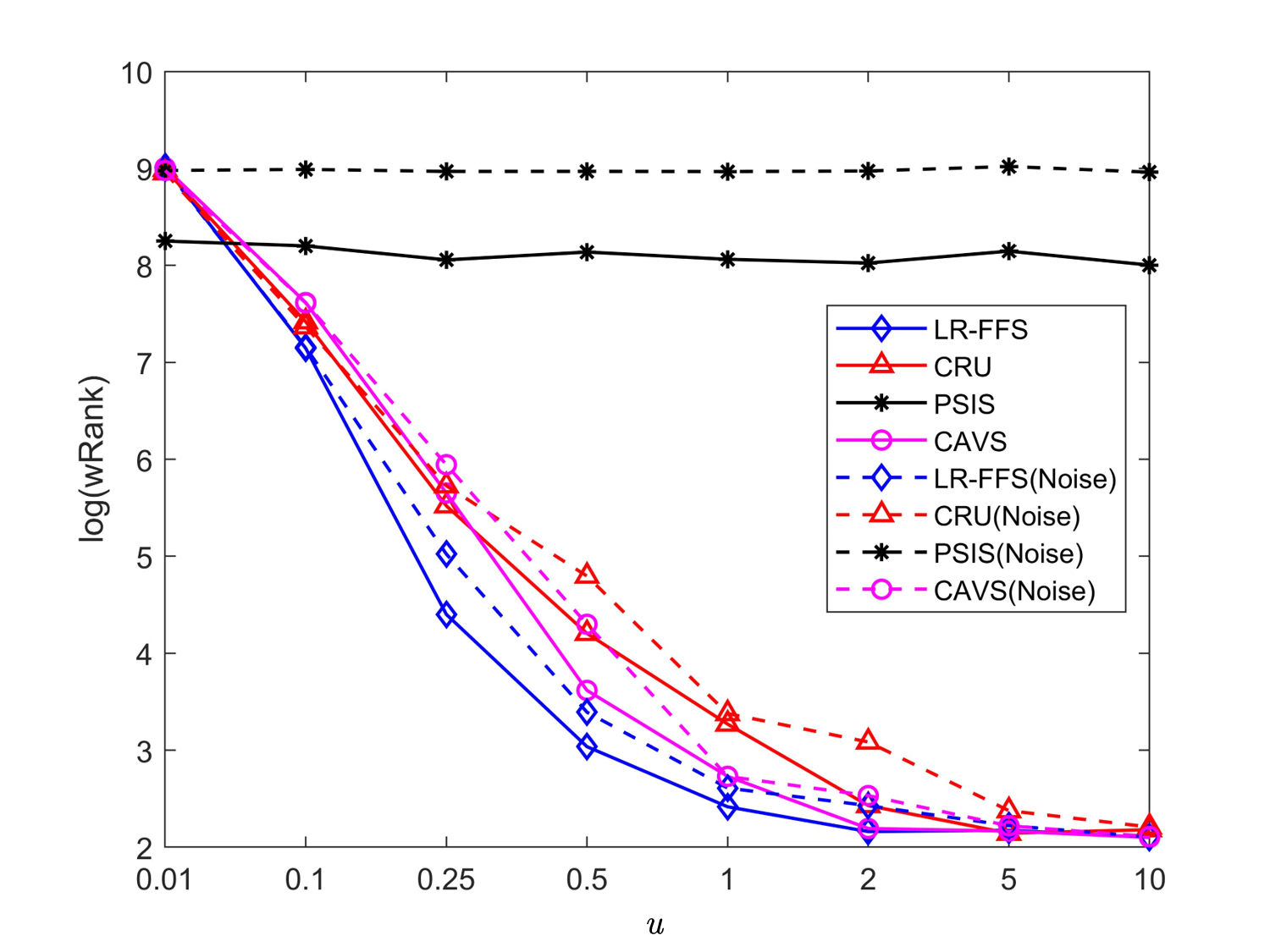}
		\end{minipage}
	}
	\subfigure[$R=6$]
	{
		\begin{minipage}[b]{.3\linewidth}
			\centering
			\includegraphics[scale=0.195]{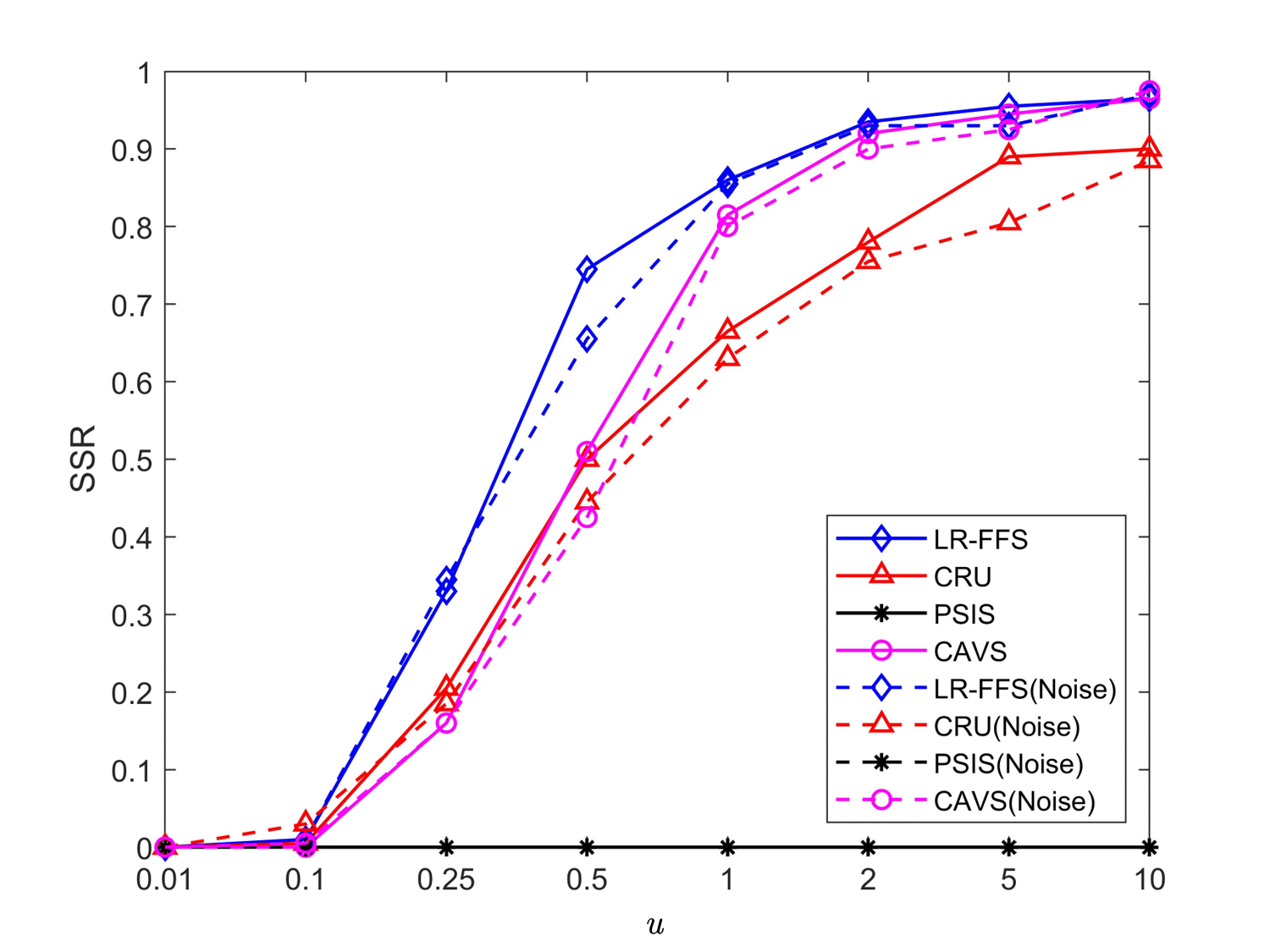} \\
			\includegraphics[scale=0.195]{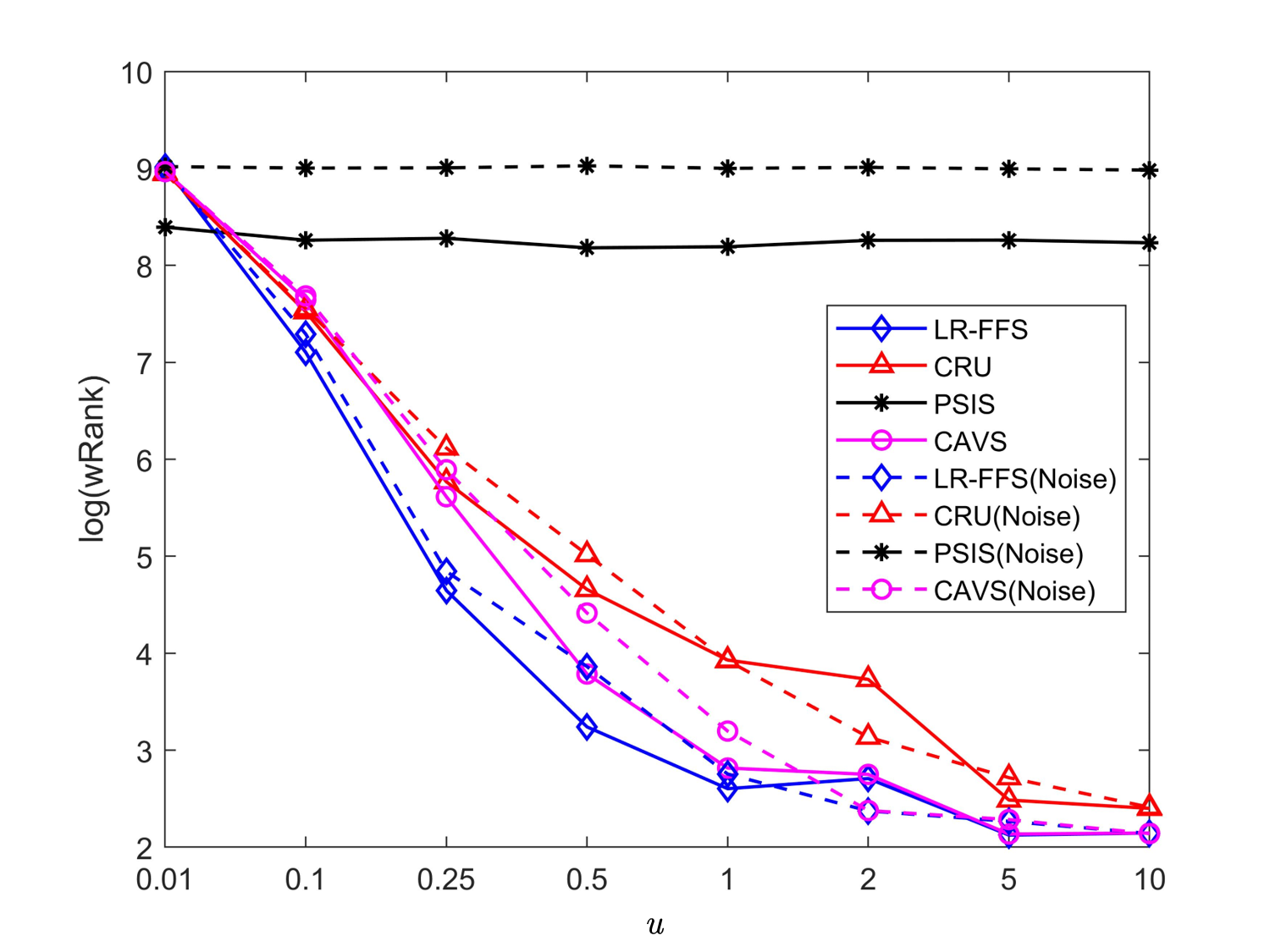}
		\end{minipage}
	}
	\subfigure[$R=7$]
	{
		\begin{minipage}[b]{.3\linewidth}
			\centering
			\includegraphics[scale=0.195]{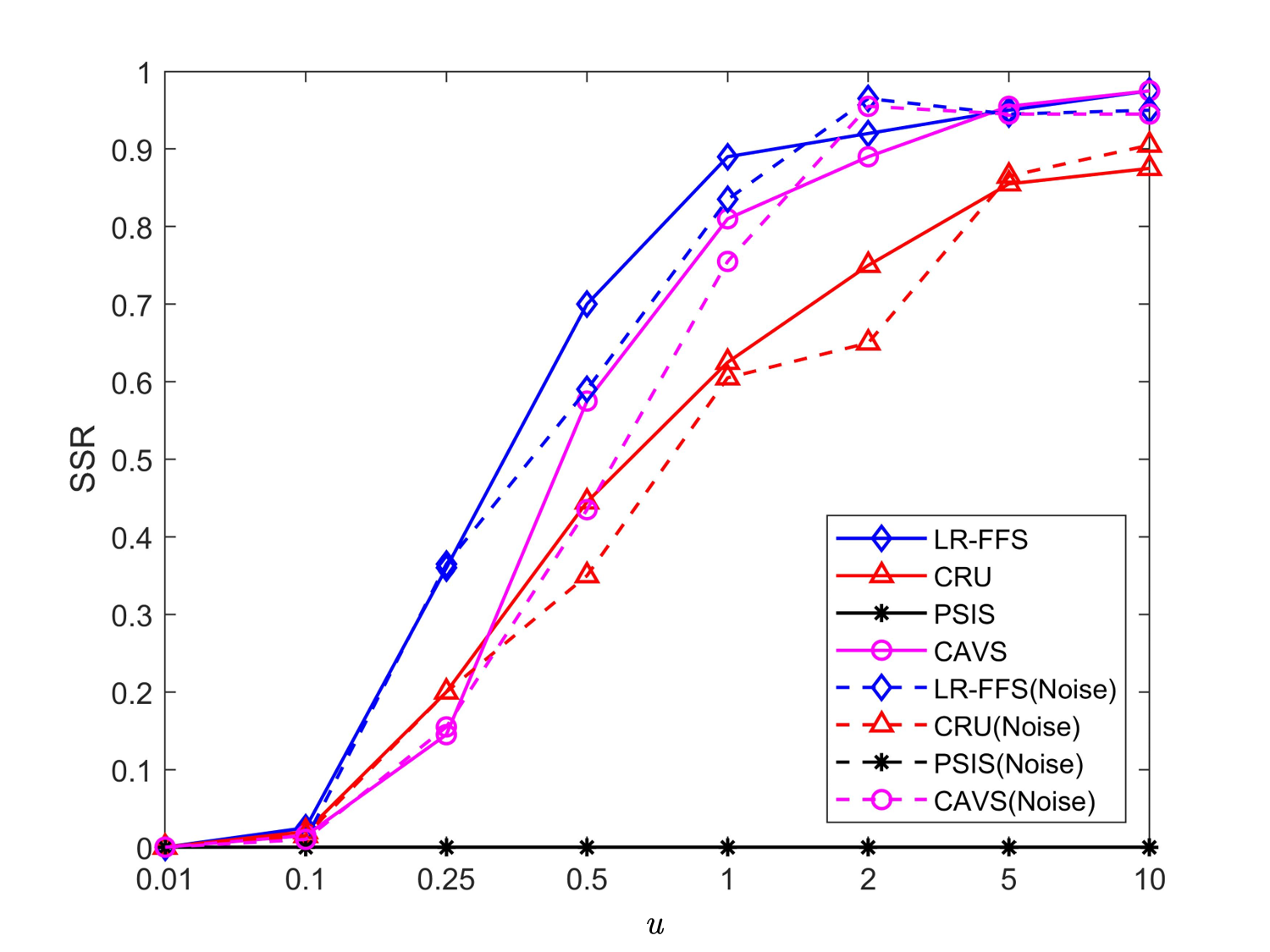} \\
			\includegraphics[scale=0.195]{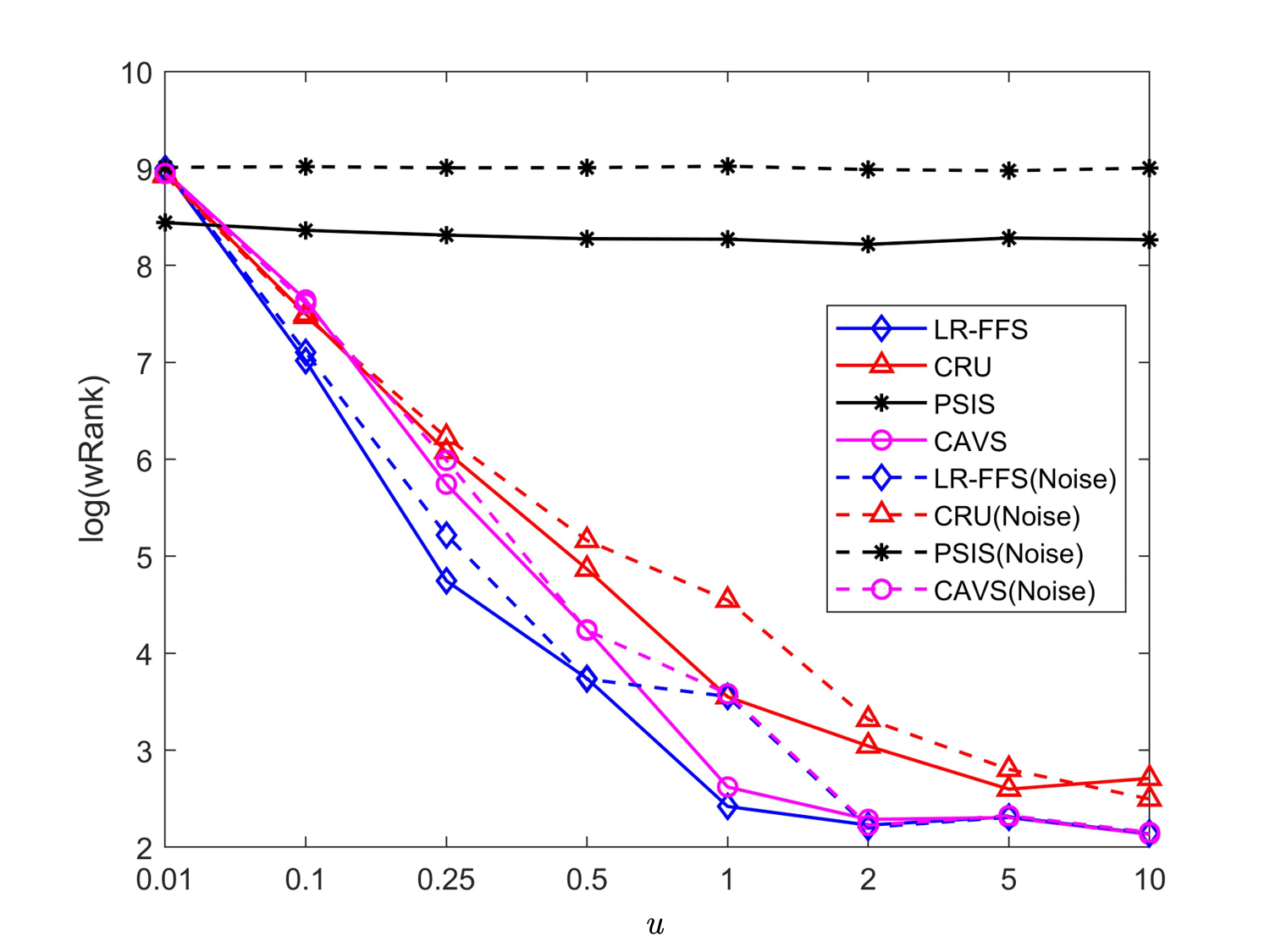}
		\end{minipage}
	}
	\caption{Simulation results for Setting (b) in Example \ref{exa:example2}, proportion of each category follows Dirichlet distribution among different clients. First row represents SSR and second row represents log(wRank).}	\label{fig:settingb}
\end{figure}

\begin{table}[!htbp]
	\centering
	\caption{Simulation results for Setting (c) in Example \ref{exa:example2}, with only partial category data on each client (parentheses indicate results of adding noise).}
	\resizebox{\textwidth}{!}{
		\begin{tabular}{ccccccc}
			\toprule
			\multicolumn{2}{c}{Number of missing categories} & 0     & 1     & 2     & 3     & 4 \\
			\midrule

			\multirow{6}[2]{*}{LR-FFS} & SSR$\uparrow$   & \textbf{1(1)} & \textbf{1(0.99)} & \textbf{0.97(0.97)} & \textbf{0.95(0.94)} & \textbf{0.82(0.77)} \\
			& PSR$\uparrow$   & \textbf{1(1)} & \textbf{1(1)} & \textbf{1(1)} & \textbf{0.99(0.99)} & \textbf{0.97(0.95)} \\
			& FDR$\downarrow$   & 0.39(0.38) & \textbf{0.41(0.40)} & 0.4(0.39) & 0.42(0.42) & \textbf{0.41(0.41)} \\
			& Size  & 19.79(19.23) & 20.77(20.27) & 20.55(20.66) & 21.21(21.03) & 19.93(20.28) \\
			& wRank$\downarrow$ & \textbf{10.05(10.04)} & \textbf{10.07(10.04)} & \textbf{10.35(10.92)} & \textbf{11.12(11.41)} & \textbf{18.1(30.11)} \\
			& Time$\downarrow$  & 1.91(1.76) & 1.8(1.75) & 1.78(1.71) & 1.75(1.71) & 1.68(1.61) \\
			\midrule
   			\multirow{6}[2]{*}{CRU} & SSR$\uparrow$   & 0.99(0.99) & 0.94(0.95) & 0.86(0.88) & 0.82(0.75) & 0.69(0.66) \\
			& PSR$\uparrow$   & 1(1)  & 0.99(0.99) & 0.97(0.98) & 0.95(0.93) & 0.88(0.87) \\
			& FDR$\downarrow$  & \textbf{0.35(0.40)} & 0.41(0.41) & 0.39(0.41) & \textbf{0.4(0.38)} & 0.44(0.44) \\
			& Size  & 18.48(20.75) & 20.87(20.52) & 19.35(21.14) & 18.94(18.19) & 19.71(18.82) \\
			& wRank$\downarrow$ & 10.65(10.25) & 11.46(14.25) & 30.22(20.92) & 27.98(58.71) & 132.08(196.3) \\
			& Time$\downarrow$  & 1.97(1.82) & 1.86(1.82) & 1.87(1.82) & 1.87(1.82) & 1.87(1.82) \\
			\midrule
			\multirow{6}[2]{*}{PSIS} & SSR$\uparrow$   & 0(0)  & 0(0)  & 0(0)  & 0(0)  & 0(0) \\
			& PSR$\uparrow$   & 0.08(0.03) & 0.06(0.05) & 0.05(0.03) & 0.05(0.04) & 0.03(0.04) \\
			& FDR$\downarrow$   & 0.86(0.66) & 0.85(0.66) & 0.85(0.69) & 0.89(0.64) & 0.88(0.72) \\
			& Size  & 10.4(158.01) & 10.83(273.02) & 10.89(158.94) & 11.16(271.42) & 8.7(273.25) \\
			& wRank$\downarrow$ & 4226.73(8424.4) & 4239.21(8475.08) & 4372.79(8309.14) & 4451.28(8450.63) & 4837.6(8391.79) \\
			& Time$\downarrow$  & \textbf{0.51(0.45)} & \textbf{0.48(0.45)} & \textbf{0.48(0.44)} & \textbf{0.48(0.44)} & \textbf{0.48(0.44)} \\
			\midrule
			\multirow{6}[2]{*}{CAVS} & SSR$\uparrow$   & 0.99(1) &\textbf{1(0.99)}  & 0.97(0.97) & 0.93(0.92) & 0.79(0.74) \\
			& PSR $\uparrow$  & 1(1)  & 1(1)  & 0.99(0.99) & 0.98(0.98) & 0.95(0.93) \\
			& FDR$\downarrow$   & 0.39(0.38) & 0.42(0.40) & \textbf{0.39(0.4)} & 0.42(0.42) & \textbf{0.41(0.41)} \\
			& Size  & 19.81(19.20) & 20.83(20.35) & 20.26(20.69) & 20.94(20.79) & 19.4(19.66) \\
			& wRank$\downarrow$ & 10.07(10.04) & 10.09(10.1) & 10.46(10.99) & 11.74(12.15) & 21.31(32.96) \\
			& Time$\downarrow$  & 2.42(2.27) & 2.33(2.28) & 2.17(2.13) & 2.13(2.08) & 2.04(1.99) \\
			\bottomrule
	\end{tabular}} \label{tab:table2}
\end{table}

In Setting (a), where outliers are absent, PSIS shows robustness against label shift and achieves effective screening. However, in Settings (b) and (c), where features exhibit heavy-tailed distributions or outliers, PSIS behaves similar to a random guess, significantly reducing its effectiveness. LR-FFS consistently delivers optimal performance across all scenarios, particularly excelling in settings with moderate client heterogeneity. CAVS, while performing suboptimally compared to LR-FFS, demonstrates the benefits of using the maximum as a special weight. For CRU, we observe that label shift severely impacts the screening results, reducing its accuracy and reliability.

Due to space constraints, the results for Setting (d) are presented in Tables \ref{result:FDR1} and \ref{result:FDR2} in the Appendix. Additionally, we explore the impact of different weight selections $\zeta_r$ on feature screening under the general framework \ref{equ:utility}, based on Setting (a) of Example  \ref{exa:example2}. Relevant details and results are provided in Example \ref{exa:example extend} in the Appendix. Among all weight selection methods, LR-FFS exhibits the optimal performance. Among the remaining weight choices, the weights used by CRU and MV-SIS, which favor categories with proportions close to 0.5 (i.e., categories with higher estimation efficiency), deliver the second-best results.

We further validate the proposed screening method in scenarios where features are correlated with each other, and the influence of features on the response variable is not constant.

\begin{example}\label{exa:example3}
    We conduct a simulation with $R$ classes and generate $ N = 3000 $ observations from a multinomial logistic model where $\log(P(Y = 1 \mid X)) \propto X\beta + \iota $. Here, $ \boldsymbol{\beta} = (\beta_1, \cdots, \beta_p)^T $ represents a vector of $ p = 8000$ regression coefficients, and $\iota$ is a constant. We deliberately assign zero values to most elements in $\boldsymbol{\beta}$, ensuring that only features with non-zero coefficients contribute to the response. We consider two setups as follows:
	
	\begin{itemize}
		\item[(e)]  $\mathbf{X} \sim N\left(\mathbf{0}, \boldsymbol{\Sigma}_1\right)$, where $\boldsymbol{\Sigma}_1$ is a $p \times p$ identity matrix. We set the index set of the relevant features $\mathcal{A}=\{1,2, \ldots, 8\}$, $\beta_j=(-1)^W\times 1$ for $j \in \mathcal{A}$ and $\beta_j=0$ for $j \notin \mathcal{A}$, where $W \sim \operatorname{Bernoulli}(0.5)$, $\iota=-0.25$. Additionally, we set $P(Y = 2|X)/1.2=P(Y = 3|X)=\cdots=P(Y = R|X)$ and substitute 30 samples with random noise, generated from a uniform distribution between 0 and 100. The setting for category heterogeneity follows Setting (b) in Example \ref{exa:example2}.
		
		\item[(f)] $\mathbf{X} \sim N\left(\mathbf{0}, \boldsymbol{\Sigma}_2\right)$, where $\boldsymbol{\Sigma}_2=\left[\sigma_{j, h}\right]_{p \times p}$ with $\sigma_{j, j}=1$, $\sigma_{j, h}=2 / 3$ for $|j-h|=1, \sigma_{j, h}=1 / 3$ for $|j-h|=2$, and $\sigma_{j, h}=0$ for $|j-h| \geq 3$. We set $\mathcal{A}=\{2,4,6,8,10,12\}$, $\beta_j=(-1)^W\times 1.5$ for $j \in \mathcal{A}$ and $\beta_j=0$ for $j \notin \mathcal{A}$, where $W \sim \operatorname{Bernoulli}(0.5)$ , $\iota=-0.2$. Similar to the previous setup, we adjust class probabilities such that $P(Y = 2|X)/0.8=P(Y = 3|X)=\cdots=P(Y = R|X)$ and introduce noise. The setting for category heterogeneity follows Setting (a) in Example \ref{exa:example2}.
	\end{itemize}
\end{example}

\begin{figure}[H]
	\centering
	\subfigure[$R=6$ for setting (e).]
	{
		\begin{minipage}[b]{.21\linewidth}
			\centering
			\includegraphics[scale=0.145]{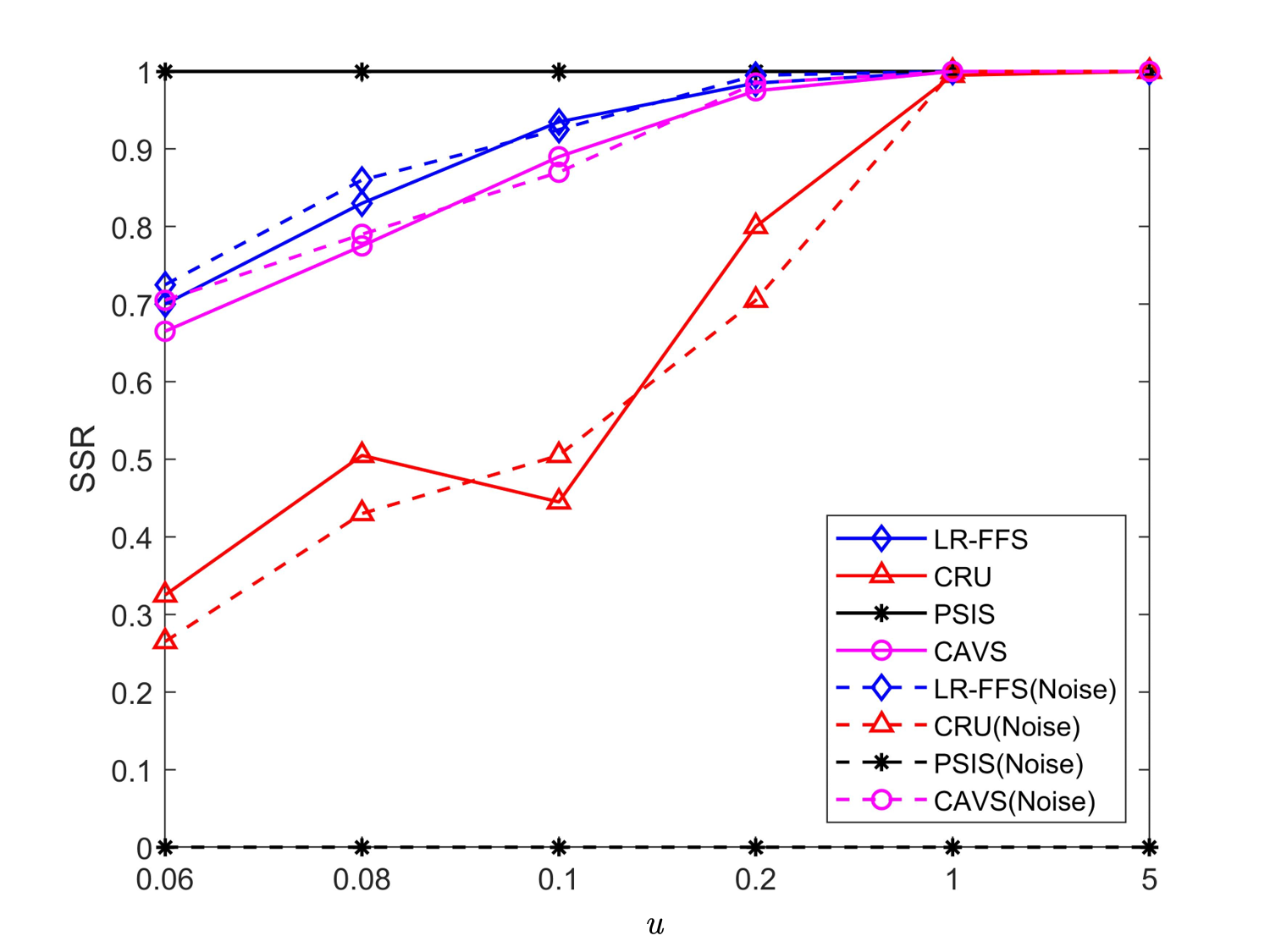} \\
			\includegraphics[scale=0.145]{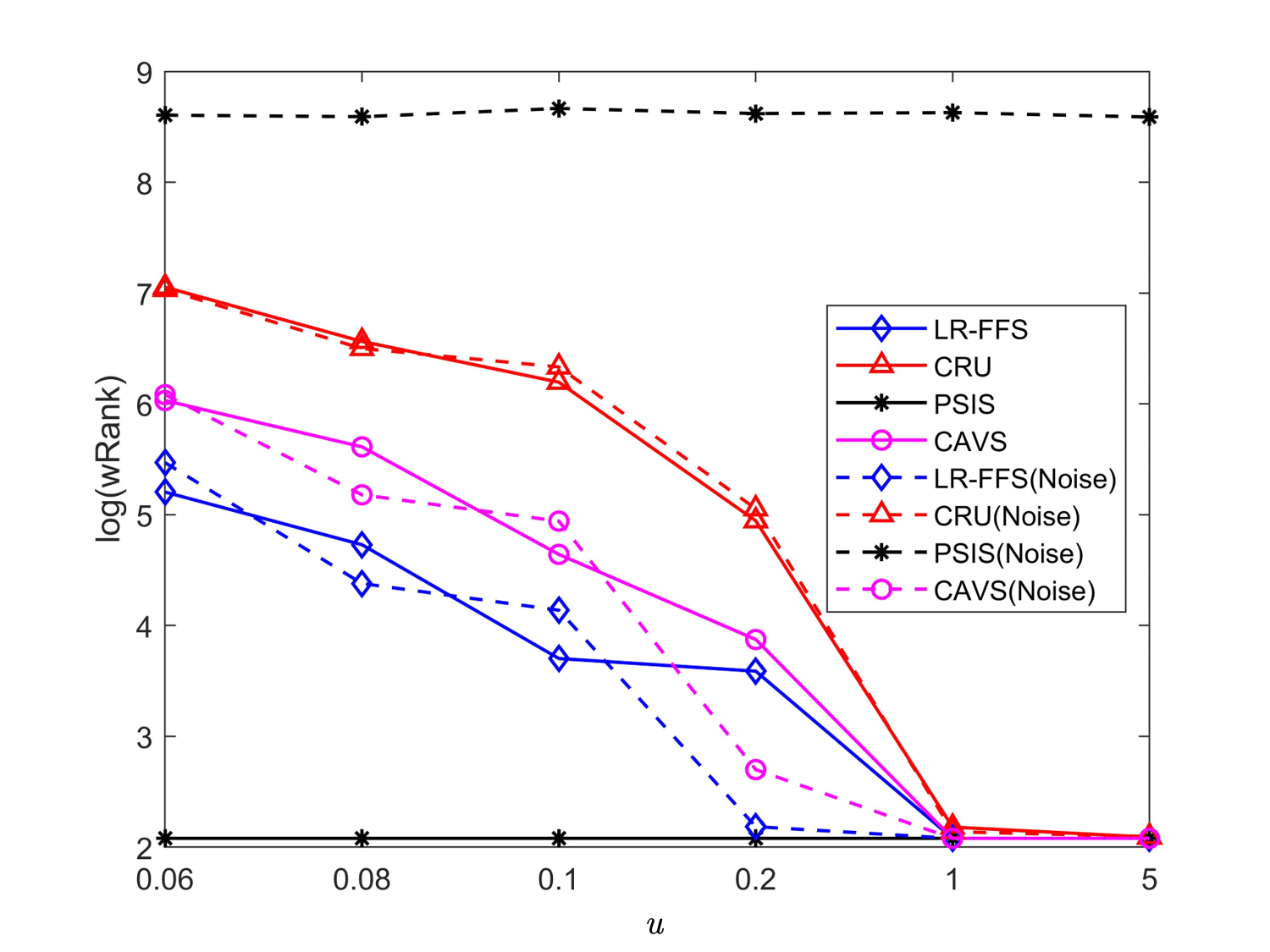}
		\end{minipage}
	}
	\subfigure[$R=7$ for setting (e).]
	{
		\begin{minipage}[b]{.21\linewidth}
			\centering
			\includegraphics[scale=0.145]{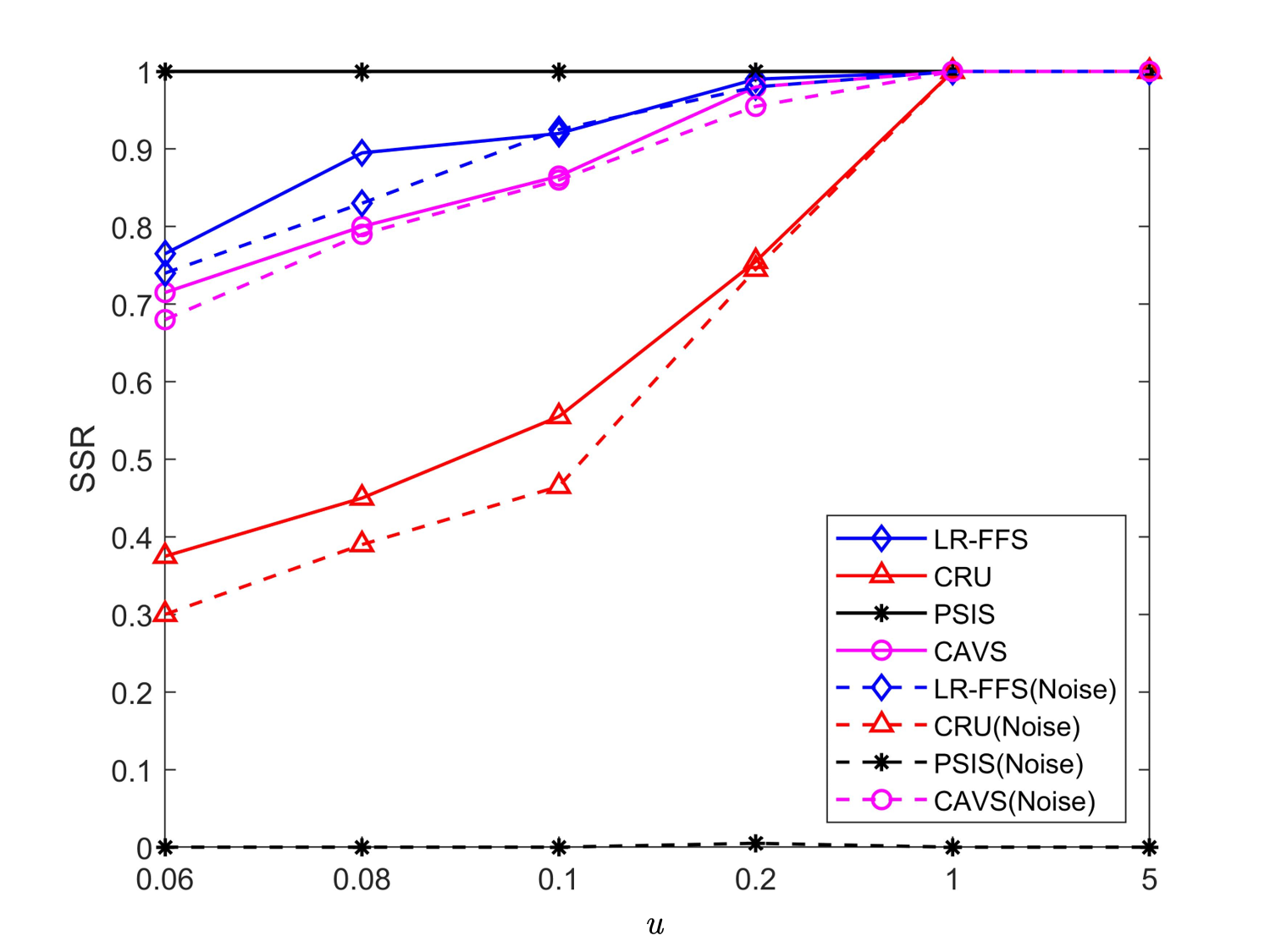} \\
			\includegraphics[scale=0.145]{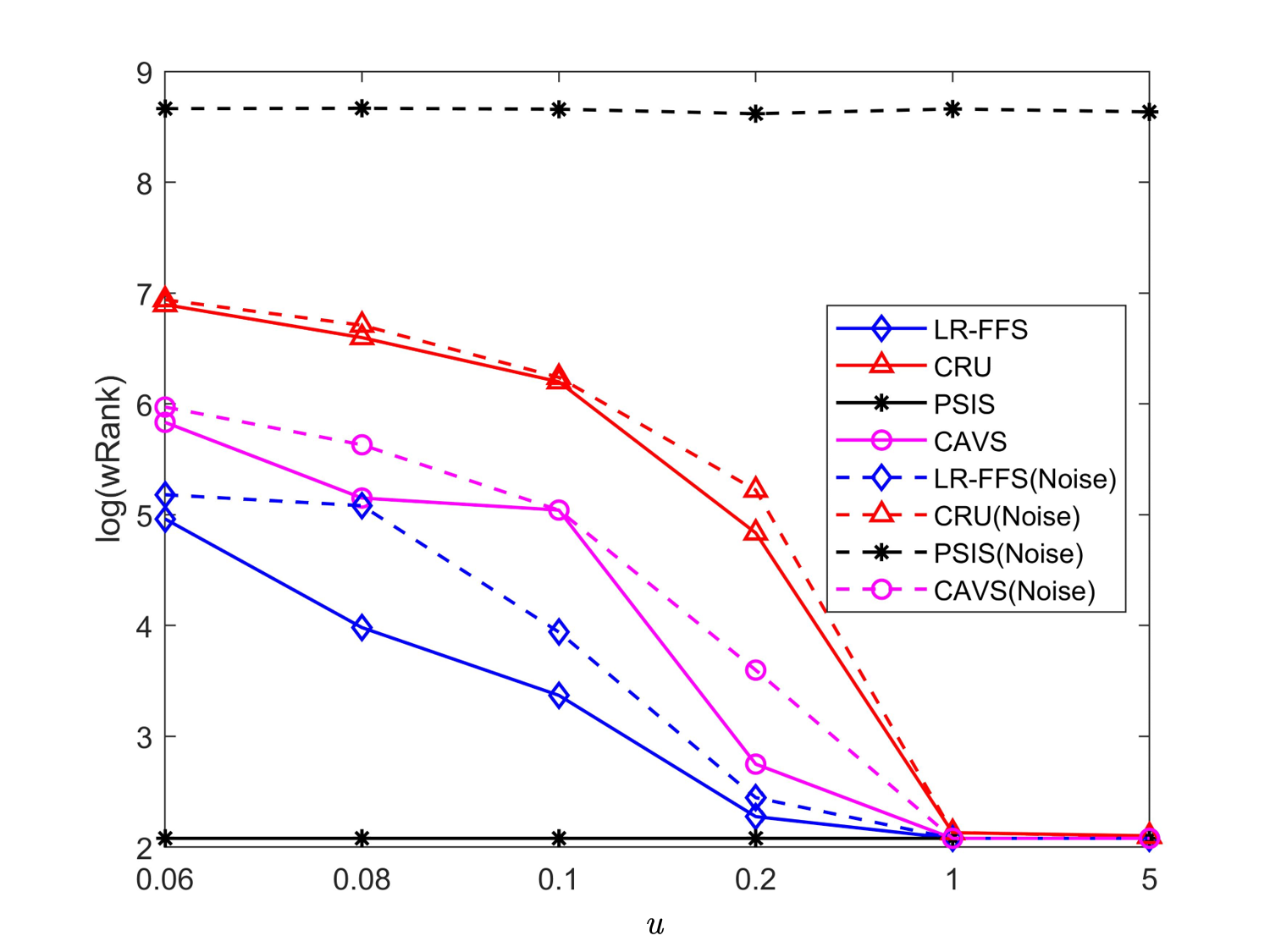}
		\end{minipage}
	}
	\subfigure[$R=6$ for setting (f).]
	{
	\begin{minipage}[b]{.21\linewidth}
		\centering
		\includegraphics[scale=0.145]{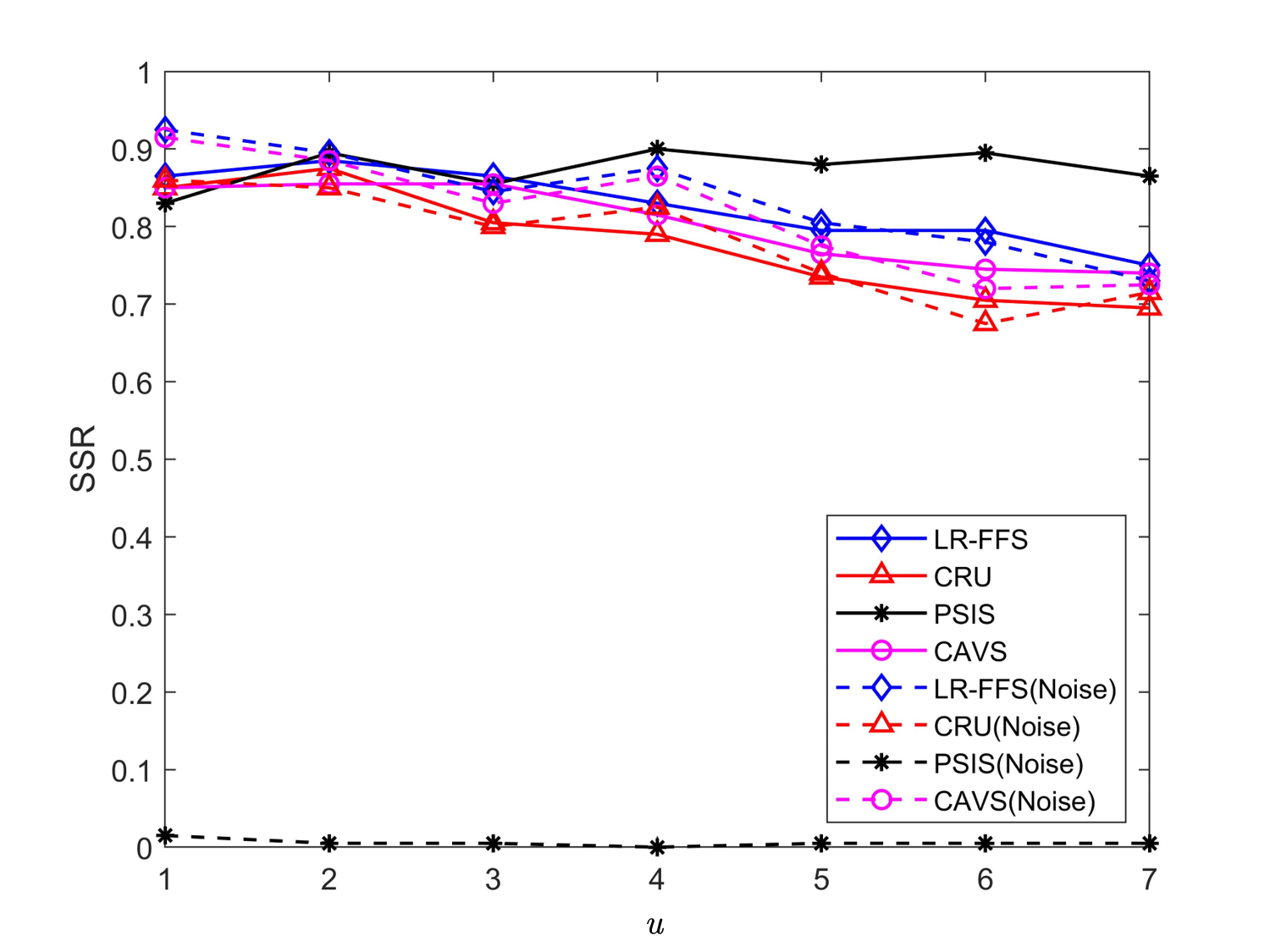} \\
		\includegraphics[scale=0.145]{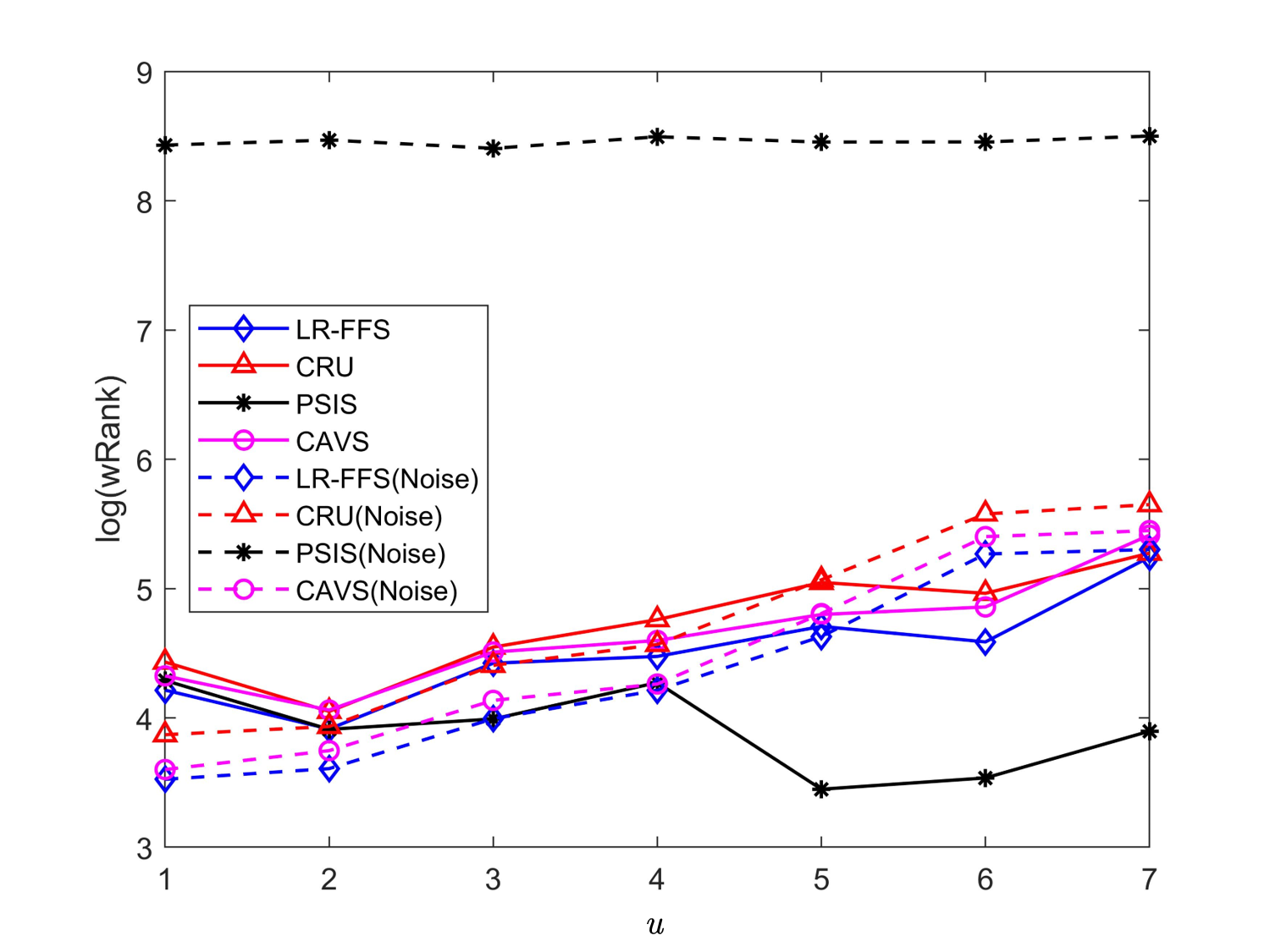}
	\end{minipage}
	}
	\subfigure[$R=7$ for setting (f).]
	{
		\begin{minipage}[b]{.21\linewidth}
			\centering
			\includegraphics[scale=0.145]{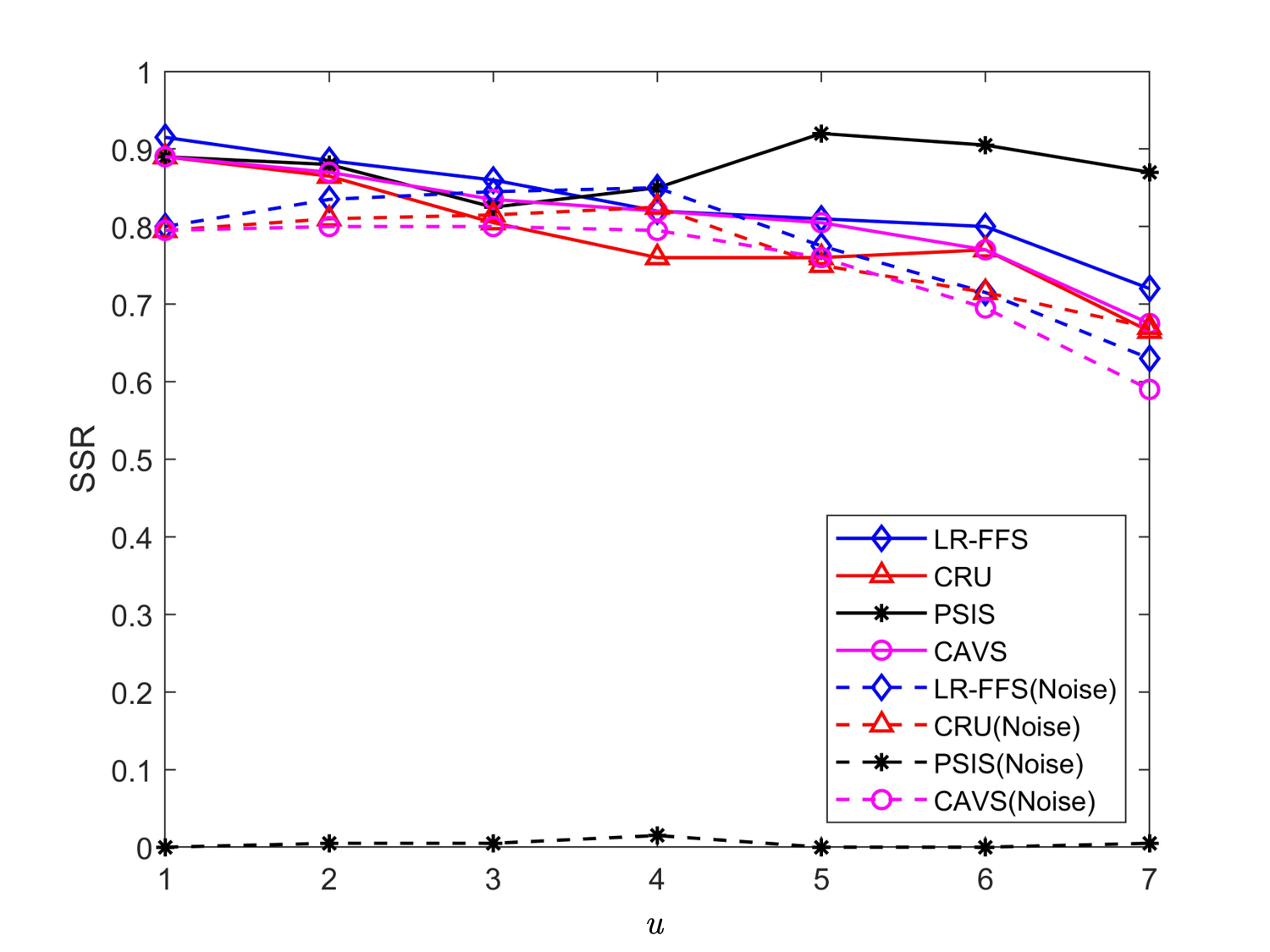} \\
			\includegraphics[scale=0.145]{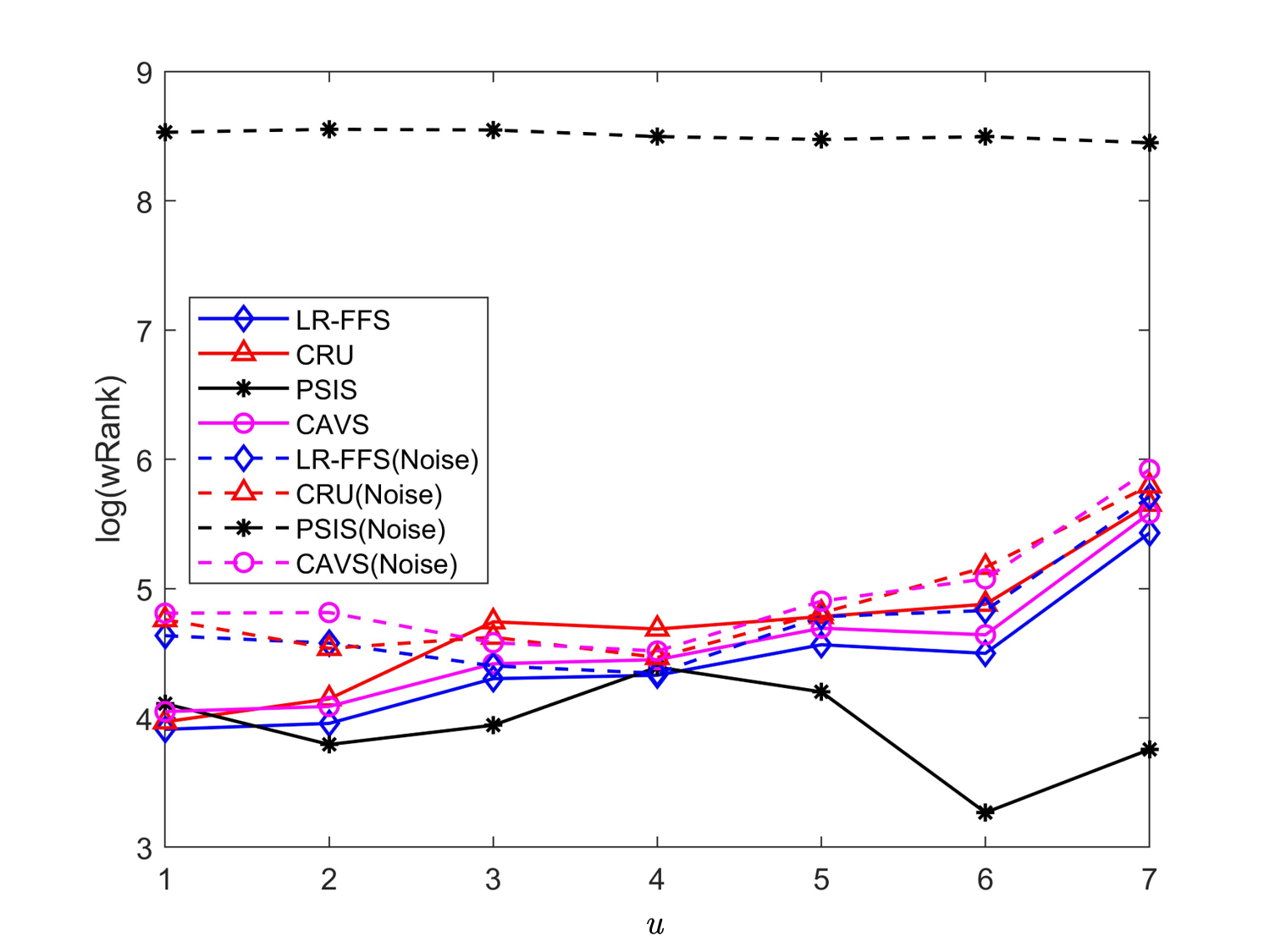}
		\end{minipage}
	}
	\caption{Simulation results for Settings (e) and (f) in Example \ref{exa:example3}, where first row represents SSR and second row represents wRank.}\label{fig:settinge}
\end{figure}

This classification problem is complex, involving multiple classes, discrepant class distributions, label shifts, and the presence of noise. Out of the $p=8000$ features, we retained 50, and the results are reported in Figure \ref{fig:settinge}.

As expected, when correlations exist among features, both configurations present greater challenges for accurate feature screening. Due to significant differences in overall class distributions, the impact of label shift on screening is relatively smaller compared to that in Example \ref{exa:example2}. Nevertheless, in all these challenging scenarios, LR-FFS continues to demonstrate superior accuracy over its competitors.

In Example \ref{exa:example4}, we assess the performance of LR-FFS under data contamination and parameter misalignment during data transmission. Specifically, we examine cases where the integrity of original client data is compromised, rather than direct tampering with parameters. Both scenarios can be considered as client attacks on the distributed system.

\begin{example}\label{exa:example4}
    We consider a total sample size of $N = 3000$, equally partitioned into 30 segments. The category heterogeneity among clients is configured to be the same as Setting (a) of Example \ref{exa:example2}, controlled by the parameter $ u$.

	\begin{itemize}
		\item[(g)] Set $R=8$, $u =6$ or $10$. The noise term $\boldsymbol{\varepsilon}$ independently follows an exponential distribution with a mean of 1. The location parameters are set to $\mu_{1j}=0.34, for 1\le j\le 8$. In addition, we assume that during the transmission process, a proportion of clients, denoted by $\phi$, misalign the parameters of different categories.
		
		\item[(h)] Set $R=7$,  $u =1$ or $6$. The noise term $\boldsymbol{\varepsilon}$ independently follows an exponential distribution with a mean of 2. The location parameters are set to $\mu_{1j}=0.50, 1\le j\le 8$. In addition, assume that a proportion $\phi$, of clients' data is contaminated, where the labels $Y$ are randomly shuffled.
		
		In both settings, we retain the top 50 most important features, and the index set of relevant features is $\mathcal{A}=\{1,\cdots,8\}$.
	\end{itemize}
\end{example}

We vary the proportion $\phi$ from 0 to 30\%. The simulation results show that even in the absence of label shift, LR-FFS exhibits advantages over competitors. Moreover, when label shifts are present,  LR-FFS's advantages become more pronounced. Detailed results for Settings (g) and (h) are provided in the Appendix (Figure \ref{fig:settingfg}) due to space limitations.

\subsection{Real Data Analysis}
We applied our proposed methodology to the Breast Invasive Carcinoma dataset, which includes comprehensive data from 981 patients across 38 institutions. This dataset contains information on mutated genes, patient demographics, and tumor typing and is part of the PanCancer Atlas initiative. It is accessible for download from the official website \href{https://gdc.cancer.gov/about-data/publications/pancanatlas}{pancanatlas}.

\begin{figure}[H]
	\centering
\includegraphics[width=0.8\textwidth]{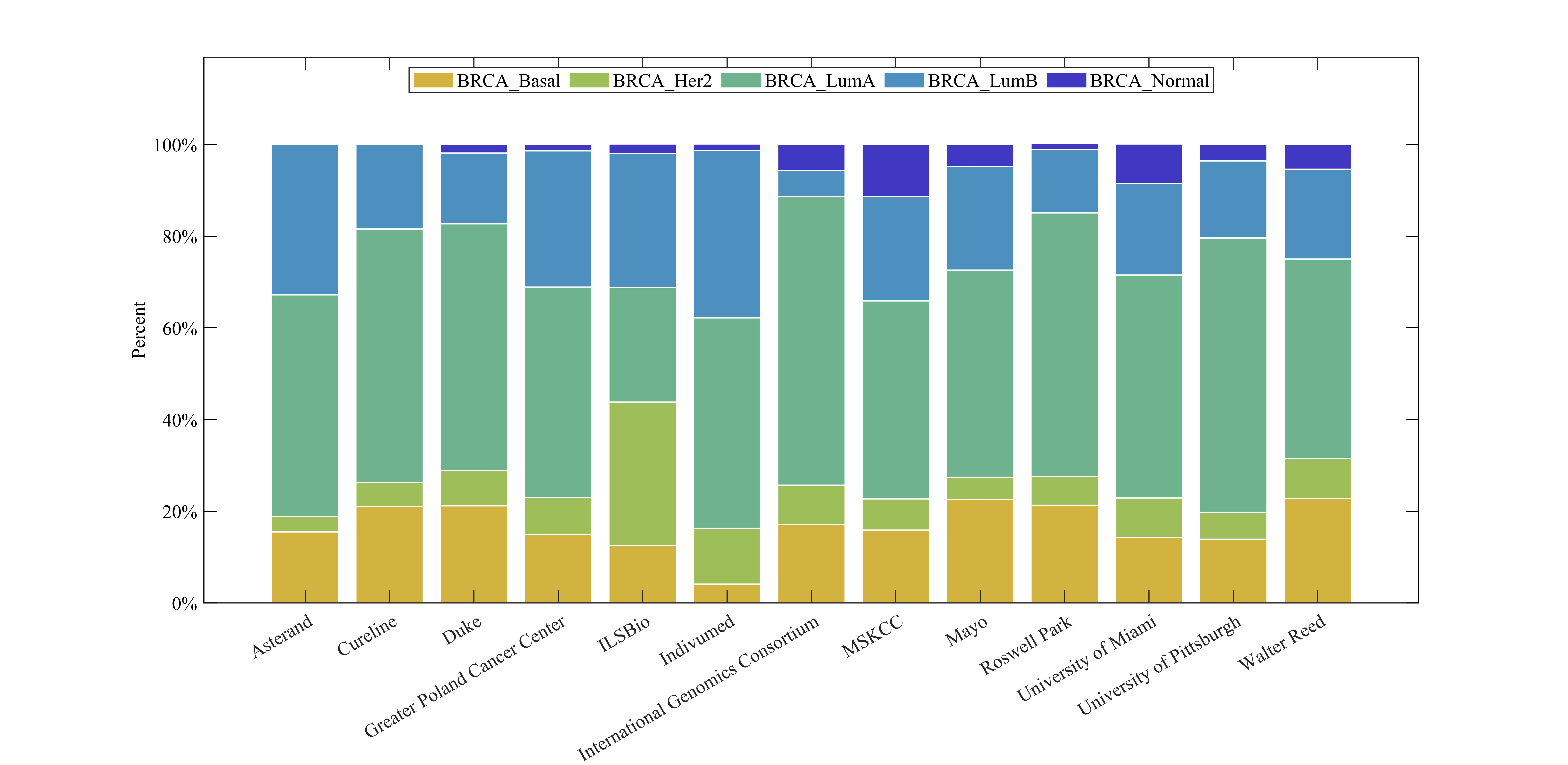}
	\caption{Proportions of Subtypes in different institutions (coefficient of contingency: 0.336, $p$-value of Pearson's chi-square test: $3.267e-06$), high coefficient of contingency and extremely low $p$-value both indicate that distribution of subtypes among different hospitals is non-IID.}
	\label{fig:tcga}
\end{figure}

Our primary objective was to develop a classifier for identifying breast cancer gene subtypes, approached as a 5-class classification task. Despite the dataset's mRNA expression data comprising 20,531 features, the limited number of available samples poses a significant challenge for accurate discrimination. Additionally, each institution's contributions to subtype proportions exhibit heterogeneity due to variations in collection time and space, as shown in Figure \ref{fig:tcga}. Ethical and privacy concerns often prevent institutions from sharing raw data, necessitating the use of federated feature screening in this medical context to ensure data privacy and compliance while leveraging the full breadth of the dataset across multiple institutions.

To train our classifier, institutions with a minimum sample size of 32 were designated as clients within the training set, while others were reserved for testing. Consequently, our training set comprised 13 clients with 829 samples, and the test set consisted of 152 samples. Detailed sample size data for each client can be found in the Appendix (Table \ref{tab:number of inst}).

In addition to presenting results from the original dataset, we conducted experiments involving noise contamination and attacks. In the noise test, we replaced all features of 30 samples in the training set with random numbers drawn from a uniform distribution ranging from 0 to 30. For the attack test, we randomly shuffled the labels of one client's samples in the training set. Each test was repeated 50 times. To ensure robustness, we reported the number of features among the top 100 in utility values that appeared more than 45 times.

\begin{figure}[H]
	\centering
	\subfigure[Original]
	{
		\begin{minipage}[b]{.3\linewidth}
			\centering
			\includegraphics[scale=0.03]{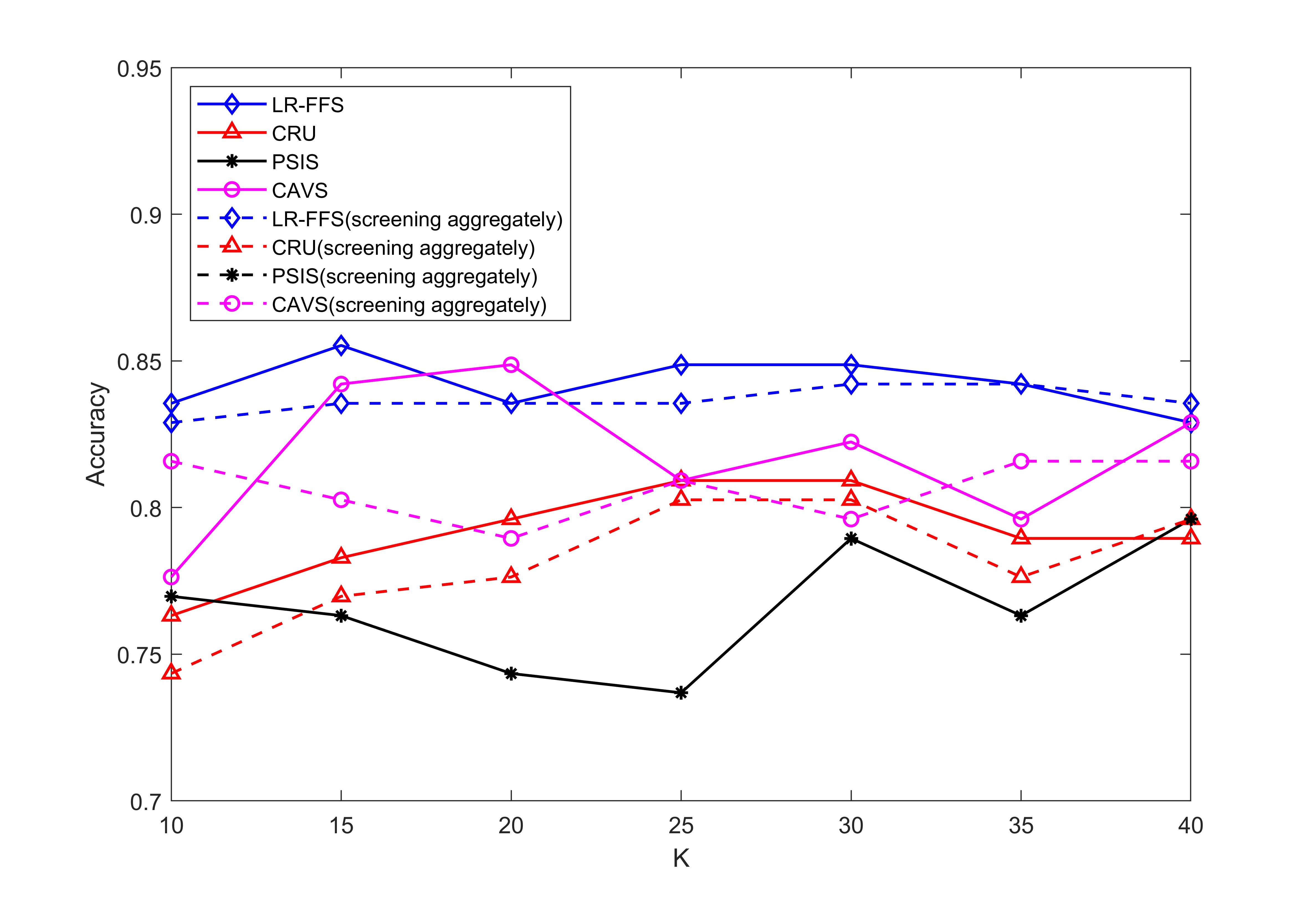}
		\end{minipage}
	}
	\subfigure[Introduce noises]
	{
		\begin{minipage}[b]{.3\linewidth}
			\centering
			\includegraphics[scale=0.03]{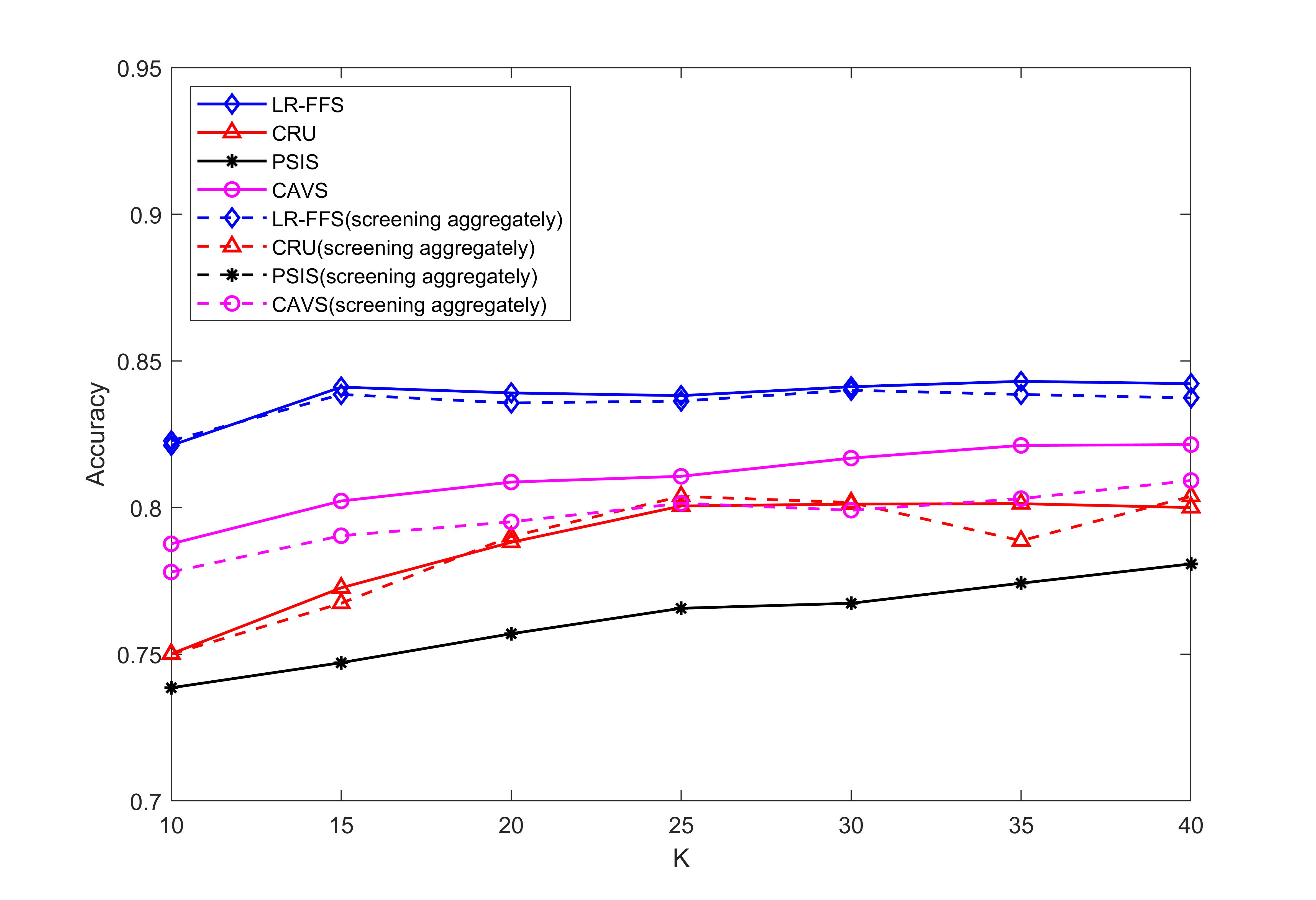}
		\end{minipage}
	}
	\subfigure[Introduce attacks]
	{
		\begin{minipage}[b]{.3\linewidth}
			\centering
			\includegraphics[scale=0.03]{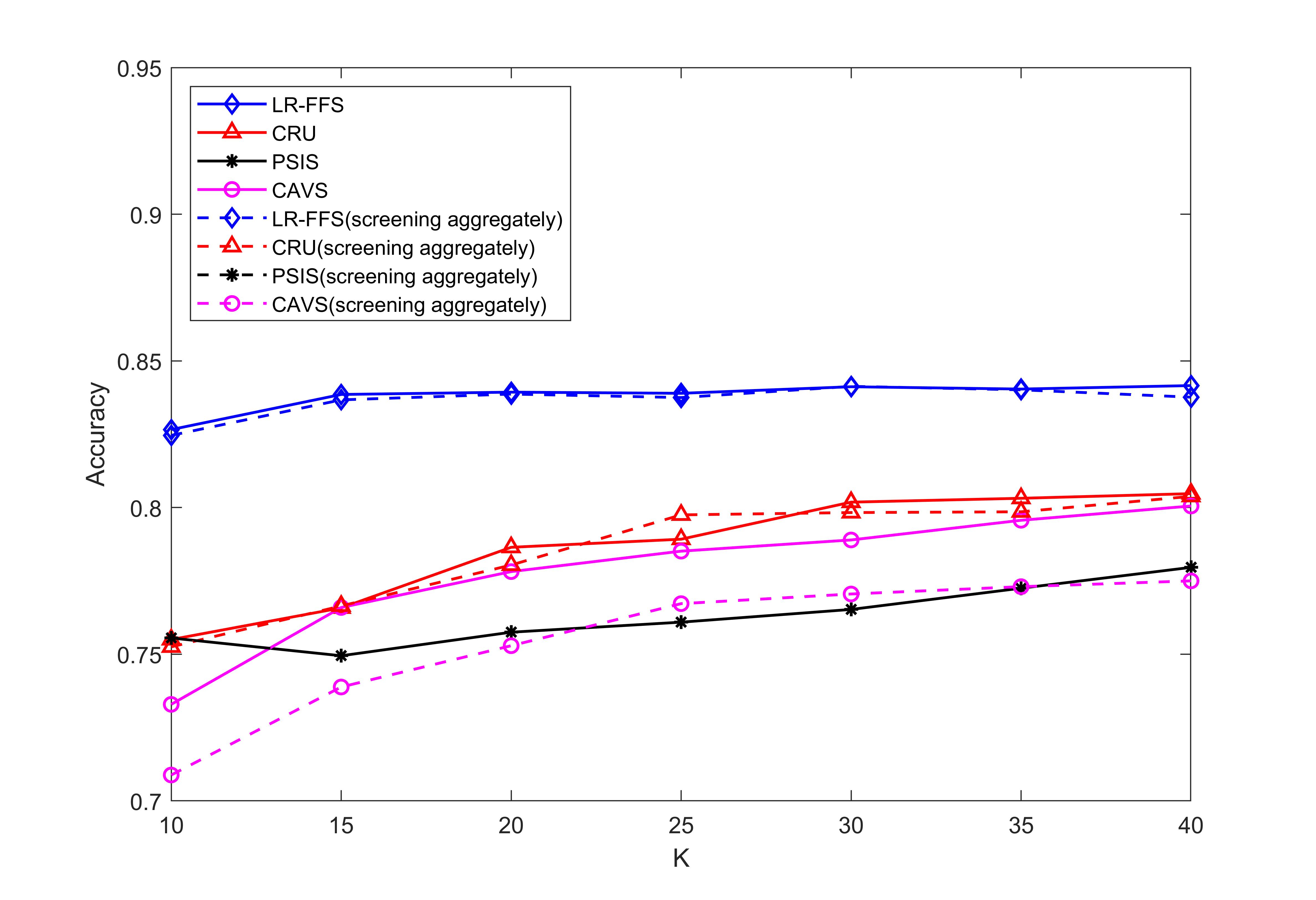}
		\end{minipage}
	}
	\caption{Classification accuracy for different screening methods in TCGA example by KNN.}\label{fig:real_result}
\end{figure}

We applied the LR-FFS method along with CRU, PSIS, and CAVS for feature screening, retaining $K$ key features. A $K$-nearest neighbors (KNN) classifier with 40 neighbors was trained using the selected features. We reported the average accuracy on the test set for both distributed estimation and estimation with aggregated data across 50 repeated experiments. The results are presented in Figure \ref{fig:real_result} and Table \ref{fig:real_robust}.

\begin{table}[!htbp]
	\centering
	\caption{In 50 repeated experiments, number of features ranking within top 100 in terms of importance exceeded 45 instances.}
	\begin{tabular}{ccccc}
		\toprule
		& LR-FFS   & CRU & PSIS  & CAVS \\
		\midrule
		Noise & 66& 76    & 3  & 42 \\
		Attack & 65& 77    & 35  & 15 \\
		\bottomrule
	\end{tabular}%
	\label{fig:real_robust}%
\end{table}%

Due to inherent outliers and noise in medical data, PSIS did not perform well. Even with additional noise and attacks introduced, LR-FFS consistently maintained superior feature screening effectiveness and exhibited stability in feature screening. Comparing the results of aggregated data versus federated screening, PSIS consistently maintained uniformity, while the other three methods showed some differences, with LR-FFS demonstrating the least variation.

\section{Conclusion and Discussions}\label{sec:conclusion}
In this study, we introduced a novel feature screening method, LR-FFS, and proposed its federated estimation procedure. This approach effectively addressed the challenges posed by heterogeneity resulting from label shifts without incurring additional computational burden, making it advantageous even in non-distributed or IID settings. The LR-FFS method efficiently quantified the relevance of features to the categorical response, ensuring stability and effectiveness in feature screening, even in scenarios with noise and outliers. The federated feature screening procedure demonstrated high computational efficiency and privacy protection, maintaining screening effectiveness comparable to centralized data processing. Our experiments and theoretical analysis confirmed that LR-FFS performed well across client environments with varying degrees of class distribution disparities and differing client sample sizes, including severe cases involving missing categorical data.

We precisely identified the sources of impact on utility estimation in a distributed context due to class distribution heterogeneity. We extended the distributed procedure to a more generalized framework, allowing various existing methods within this new framework to alleviate the impact of label shifts and achieve excellent screening properties.

This study focused on distribution heterogeneity in the context of label shifts. Future research could consider distribution heterogeneity caused by covariate shifts or minor model shifts. A promising direction would be designing a personalized federated feature screening method that iteratively identifies and retains important features in data segments. Finally, while we conducted experimental simulations involving node attacks and noted that robust aggregation methods could enhance attack resistance, reducing accuracy loss in this process remains an interesting topic for future research.

While our current privacy framework effectively prevents clients from sharing raw data, stronger privacy guarantees—such as those achievable through differential privacy mechanisms—could be explored in future work. However, integrating such methods into our proposed framework introduces significant technical challenges, including the careful balancing of privacy budgets with model utility, which lies beyond the scope of this study. We identify this as an important direction for future research to further enhance privacy preservation in distributed feature screening.

% Acknowledgements and Disclosure of Funding should go at the end, before appendices and references

\acks{Xingxiang Li’s work was supported by NSFC grant (12401394), Postdoctoral Fellowship of CPSF (GZB20240611), and China Postdoctoral Science Foundation (2024M752549). Sun’s work was supported by National Nature Science Foundation of China (12171479) and the MOE Project of Key Research Institute of Humanities and Social Sciences (NO. 22JJD110001). Wang's work was supported by National Natural Science Foundation of China (NO. 12201627). Xu's work was supported by Major Project of Pengcheng Laboratory under grant PCL2024AS103. This work was supported by Public Computing Cloud, Renmin University of China. No potential conflict of interest was reported by the author(s).}


\newpage

\appendix
In the supplementary material, we provide additional simulation results and proofs for the main paper's propositions and theorems. All notations and formula labels refer to the main text.

\section{Proofs of theorems and lemmas}
\begin{proof}[Proof of proposition \ref{pro:equal}]
	The derivation process is straightforward, notice that
    \begin{align*}
    & \mathbb{E}_{Y=y_{r}}\left(\left(F_{Y \neq y_{r}}(X)-F_{Y=y_{r}}(X)\right)^{d}\right)=\int\left(F_{Y \neq y_{r}}(X)-F_{Y=y_{r}}(X)\right)^{d} d F_{Y=y_{r}}(X) \\
    & =\int\left(F_{Y \neq y_{r}}(X)-F_{Y=y_{r}}(X)\right)^{d} d\left(F_{Y=y_{r}}(X)-F_{Y \neq y_{r}}(X)\right) \\
    &+ \int\left(F_{Y \neq y_{r}}(X)-F_{Y=y_{r}}(X)\right)^{d} d F_{Y \neq y_{r}}(X) 
    =\mathbb{E}_{Y \neq y_{r}}\left(\left(F_{Y \neq y_{r}}(X)-F_{Y=y_{r}}(X)\right)^{d}\right).
    \end{align*}
    
The last equation holds for 
\begin{align*}
    \int\left(F_{Y \neq y_{r}}(X)-F_{Y=y_{r}}(X)\right)^{d} &d\left(F_{Y=y_{r}}(X)-F_{Y \neq y_{r}}(X)\right)\\
    &=-\frac{1}{d+1}\left(F_{Y \neq y_{r}}(X)-F_{Y=y_{r}}(X)\right)^{d+1}\mid_{-\infty}^{\infty}=0.
\end{align*}
\end{proof}

\begin{proof}[Proof of proposition \ref{pro:consistent case}]
	Let's assume that $Z_1$ follows the distribution function of $x$ under the condition $Y=y_r$,  $ Z_1 \sim F_{Y=y_r}(x) $, then it's evident that $Z_1$ is independent of the proportion of $Y = y_r$.
	
	Additionally, assume $ Z_2 \sim F_{Y\ne y_r}(x) $, then
	\begin{align*}
	F_{Y\ne y_r}(x)=\sum_{y\ne y_r}{\frac{P\left( Y=y \right)}{\sum_{y\ne y_r}{P\left( Y=y \right)}} F_{Y=y}(x)} =\sum_{y\ne y_r}\eta_y{F_{Y=y}(x)},
	\end{align*}
	where $\eta_y$ represents the relative proportion of $Y=y$ relative to the proportion of $Y\ne y_r$, hence $ Z_2$ is unaffected by the proportion of $Y= y_r$.
	
	Therefore, from \ref{equ:compare},
	\begin{equation}
	\mathbb{E} _{Y=y_r}\left( F_{Y\ne y_r}\left( X_j \right) \right) =P\left( X_{j,i}<X_{j,k}|Y_i \ne y_r,Y_k= y_r \right) =P\left( Z_2<Z_1 \right), 
	\end{equation}
	is not influenced by $P(Y=y_r)$.
\end{proof}

Before presenting the proofs of Proposition \ref{pro:omega_bound} and \ref{pro:variance}, as well as theorems, we need to introduce some technical lemmas.

\begin{lemma}\label{lem:hoeffding}
(Hoeffding's inequality) Let $X_{1}, \ldots, X_{N}$ be independent random variables. Assume that $P\left(X_{i} \in\left[a_{i}, b_{i}\right]\right)=1$ for $1 \leq i \leq N$, where $a_{i}$ and $b_{i}$ are constants. Let $\bar{X}=N^{-1} \sum_{i=1}^{N} X_{i}$. Then the following inequality holds

$$P(|\bar{X}-\mathbb{E}(\bar{X})| \geq \varepsilon) \leq 2 \exp \left(-\frac{2 N^{2} \varepsilon ^{2}}{\sum_{i=1}^{N}\left(b_{i}-a_{i}\right)^{2}}\right),$$

where $\varepsilon$ is a positive constant and $\mathbb{E}(\bar{X})$ is the expected value of $\bar{X}$.
\end{lemma}

\begin{lemma} \label{lemma:hoeffding2}
     (Hoeffding's lemma) Let $X$ be a bounded random variable with $X \in[a, b]$. Then
    
    $$\mathbb{E}(\exp \{s(X-\mathbb{E}(X))\}) \leq \exp \left(\frac{s^{2}(b-a)^{2}}{8}\right) \text { for any } s>0.$$
\end{lemma}

\begin{lemma} \label{lem:bound}
For any $\varepsilon>0$ and $j \in \{1,\cdots,p\}$, we have
\begin{align}
    P\left(\left|\bar{\theta}_{r}-\theta_{r}^* \right| \geq \varepsilon\right) \leq 2 \exp \left( -\sum_{l=1}^m \lfloor \frac{n_l}{2} \rfloor \varepsilon ^{2}\right) \label{lem:bound1} \\
    P\left(\left|\bar{U}_{j,r}-{U}_{j,r}^* \right| \geq \varepsilon\right) \leq 2 \exp \left( -\sum_{l=1}^m \lfloor \frac{n_l}{2} \rfloor \varepsilon ^{2}\right)  \label{lem:bound2}
\end{align}
\end{lemma}

Lemma \ref{lemma:hoeffding2} and \ref{lem:bound} are widely applied, and their proofs can be found in most textbooks. Therefore, we will skip their detailed proofs here.

\begin{proof}[Proof of Lemma \ref{lem:bound}]
The proof is similar to the steps outlined in \citet{li_feature_2023}, we first prove the first conclusion and bound the term $\left|\bar{\theta}_{r}-\theta_{r}^* \right|$.

Let $\tilde{\theta}_{r}^l\left(Z_{i_{1}}, Z_{i_{2}}\right)=\left[I\left( 
Y_{i_1}=y_r \right) I\left(Y_{i_2}\ne y_r \right)+I\left( 
Y_{i_2}=y_r \right) I\left(Y_{i_1}\ne y_r \right)\right] / 2$ be an unbiased and symmetric estimator (kernel) of $\theta_{r}^l$ with the minimal 2 i.i.d copies of $Z_{j}=\left\{X_{j},Y\right\}$. Recall that $\mathcal{S}_{l}=\left\{l_{1}, \ldots, l_{n_{l}}\right\}$ denotes the index set of $\{ \mathbf{X},Y\}$ copies based on $\mathcal{D}_{l}$, on which we can construct $h_{l}=\left\lfloor n_{l} / 2\right\rfloor$ independent $\tilde{\theta}_{r}^l $s. Then, we define an averaged estimator based on these independent $\tilde{\theta}_{r}$ by
$$V_{r}^l\left(Z_{l_{1}}, \ldots, Z_{l_{n_{l}}}\right)=\frac{1}{h_{l}} \sum_{u=1}^{h_{l}} \tilde{\theta}_{r}^l\left(Z_{l_{2(u-1)+1}}, Z_{l_{2 u}}\right)$$

Based on $V_{r}^l\left(Z_{l_{1}}, \ldots, Z_{l_{n_{l}}}\right), \hat{\theta}_{r}^{l}$ can be further expressed by

\begin{equation}
\hat{\theta}_{r}^{l}=\frac{1}{n_{l}!} \sum_{\left\{i_{1}, \ldots, i_{n_{l}}\right\} \in \Omega_{l}} V_{r}^l \left(Z_{l_{i_{1}}}, \ldots, Z_{l_{i_{l}}}\right) \label{equ:thetaV}
\end{equation}

where $\Omega_{l}=\left\{1, \ldots, n_{l}\right\}$ and the summation is over all $\left\{Z_{l_{i_{1}}}, \ldots, Z_{l_{i_{l}}}\right\}$ permutations from $\mathcal{D}_{l}$. 

Consequently,
\begin{equation}
\bar{\theta}_{r}=\frac{\sum_{l=1}^{m} h_{l} \hat{\theta}_{r}^{l}}{\sum_{l=1}^{m} h_{l}}=\frac{1}{\sum_{l=1}^{m} h_{l}}
 \sum_{l=1}^{m} \sum_{\left\{i_{1}, \ldots, i_{n_{l}}\right\} \in \Omega_{l}} \frac{h_{l}}{n_{l}!} V_{r}^l \left(Z_{l_{i_{1}}}, \ldots, Z_{l_{i_{l}}}\right)
\end{equation}

By Markov's inequality, we have
\begin{align*}
P\left(\bar{\theta}_{r}-\theta_{r}^*  \geq \varepsilon\right) & =P\left(\exp \left\{\nu\left( \bar{\theta}_{r}-\theta_{r}^* \right)\right\} \geq \exp \{\nu \varepsilon\}\right) \\
& \leq \exp \{-\nu \varepsilon\} \exp \left\{-\nu \theta_{r}^*\right\} \mathbb{E}\left[\exp \left\{\nu \bar{\theta}_{r}\right\}\right]
\end{align*}

for any $\varepsilon>0$ and $\nu>0$. Since $\exp (\cdot)$ is convex, Jensen's inequality implies that
\begin{align*}
\mathbb{E}\left[\exp \left\{\nu \bar{\theta}_{r}\right\}\right] & =\mathbb{E}\left[\exp \left\{ \frac{\nu}{\sum_{l=1}^{m} h_{l}} \sum_{l=1}^{m}\sum_{\left\{i_{1}, \ldots, i_{n_{l}}\right\} \in \Omega_{l}} \frac{h_{l}}{n_{l}!} V_{r}^l \left(Z_{l_{i_{1}}}, \ldots, Z_{l_{i_{l}}}\right)   \right\}\right] \\
& = \prod_{l=1}^m \mathbb{E}\left[\exp \left\{ \frac{\tau}{n_{l}!} \sum_{\left\{i_{1}, \ldots, i_{n_{l}}\right\} \in \Omega_{l}} h_{l} V_{r}^l \left(Z_{l_{i_{1}}}, \ldots, Z_{l_{i_{l}}}\right)   \right\}  \right]\\
&\le \prod_{l=1}^m \left\{\frac{1}{n_{l}!}  \sum_{\left\{i_{1}, \ldots, i_{n_{l}}\right\} \in \Omega_{l}} \mathbb{E}\left[\exp \left\{ \tau  h_{l} V_{r}^l \left(Z_{l_{i_{1}}}, \ldots, Z_{l_{i_{l}}}\right)   \right\}  \right]\right\}\\
&\le \prod_{l=1}^m \mathbb{E}\left[\exp \left\{ \tau  h_{l} V_{r}^l \left(Z_{l_{i_{1}}}, \ldots, Z_{l_{i_{l}}}\right)   \right\}  \right]=\prod_{l=1}^m \mathbb{E}^{h_l}\left[\exp \left\{\tau \tilde{\theta}_{r}^l \right\}\right]
\end{align*}

where $\tau=\nu /\left(\sum_{l=1}^{m} h_{l}\right)$.

Besides, from $\sum_{l=1}^m{h_l }{\pi_r}^l\left( 1-{\pi_r}^l \right) =\sum_{l=1}^m{h_l}{\pi_r}^*\left( 1-{\pi_r}^* \right) =\sum_{l=1}^m{h_l}\theta_{r}^*$, We can deduce that 
$$\exp \left\{-\nu \theta_{r}^*\right\} = \exp \left\{-\tau \sum_{l=1}^{m} h_{l}\theta_{r}^*\right\}=\exp \left\{-\tau \sum_{l=1}^{m} h_{l}{\pi_r}^l(1-{\pi_r}^l)\right\}=\prod_{l=1}^m \exp^{h_l} \left\{-\tau \theta_r^l\right\}$$

Then,
\begin{align}
    P\left(\bar{\theta}_{r}-\theta_{r}^*  \geq \varepsilon\right) &\leq \exp \{-\nu \varepsilon\} \exp \left\{-\nu \theta_{r}^*\right\} \mathbb{E}\left[\exp \left\{\nu \bar{\theta}_{r}\right\}\right] \nonumber \\
   & \le \prod_{l=1}^m \left[\exp \{-\tau \varepsilon\} \exp \left\{-\tau \theta_r^l\right\}\exp \left\{\tau\tilde{\theta}_{r}^l \right\} \right]^{h_{l}}. \label{equ:permu}
\end{align}

Since $\tilde{\theta}_{r}^l \in[0,1]$ and $\mathbb{E}(\tilde{\theta}_{r}^l)={\theta}_{r}^l$, by using Lemma \ref{lemma:hoeffding2}, the factor $ \exp \left\{-\tau \theta_r^l\right\}\exp \left\{\tau \tilde{\theta}_{r}^l \right\}$  can be bounded by

$$ \exp \left\{-\tau \theta_r^l\right\}\exp \left\{\tau \tilde{\theta}_{r}^l \right\} \leq \exp \left(\tau^{2} / 8\right).$$

Thus, $\exp \{-\tau \varepsilon\} \exp \left\{-\tau \theta_r^l\right\}\exp \left\{\tau \tilde{\theta}_{r}^l \right\}$ can be further bounded by 
\begin{equation}
    \exp \{-\tau \varepsilon\} \exp \left(\tau^{2} / 8\right)\leq \exp \left(-2 \varepsilon ^{2}\right), \label{equ:e5}
\end{equation}
where the last inequality is based on the fact that $\tau^{2} / 8-\tau \varepsilon$ is a quadratic function achieving its minimum at $\tau=4 \varepsilon$. 

Combining \ref{equ:permu} and \ref{equ:e5}, we have
$$
P\left(\bar{\theta}_{r}-\theta_{r}^*  \geq \varepsilon\right)  \leq \exp \left(-2 \sum_{l=1}^m h_l \varepsilon ^2\right) .
$$

Similarly, we can show that $P\left(\bar{\theta}_{r}-\theta_{r}^*  \leq-\varepsilon\right) \leq \exp \left(-2 \sum_{l=1}^m h_l \varepsilon ^2\right)$. Therefore, we obtain
$$
P\left(\left|\bar{\theta}_{r}-\theta_{r}^* \right| \geq \varepsilon\right) \leq 2 \exp \left(-2 \sum_{l=1}^m\left\lfloor n_l / 2\right\rfloor \varepsilon ^2\right) .
$$

Repeating the above steps, we can easily get inequality \ref{lem:bound2}, the proof of Lemma \ref{lem:bound} is completed.
\end{proof}

In fact, we did not specifically focus on the exact values of $h_l$ during the proof process. Even when $h_l = n_l $ or $\frac{n_l(n_l-1)}{n_l+1}$, we can still provide corresponding bounds. Particularly, when $ h_l = \frac{n_l(n_l-1)}{n_l+1}$, it is equivalent to weighting the estimates of $ \omega_{j,r} $ from different clients with $\lambda_{l,r} = \frac{12|A_r^l||B_r^l|}{n_l+1}$ in Algorithm \ref{alg:procedure}. According to the classical result of the Mann-Whitney test, when samples of $Y = y_r $ and $Y \neq y_r$ come from the same distribution, the variance of $\omega_{j,r} $ is $\frac{n_l+1}{12|A_r^l||B_r^l|} $. Using a weight of $\lambda_{l,r} = \frac{12|A_r^l||B_r^l|}{n_l+1}$ achieves the ``minimum unbiased variance combination'' of aggregated results, further improving estimation accuracy. Details regarding the choice of weights are provided in the Appendix (\ref{exa:example extend}). To demonstrate that our method's superior performance effectively identifies the source of label shift effects rather than merely adjusting weights, we continue to use $ h_l = \lfloor n_l / 2 \rfloor $ in the main text.

Now we turn to analyze the estimation properties of $\bar{\gamma}_{j,r}$. 

\begin{lemma} \label{lem:boundgamma}
   Suppose condition (C1) hold. For any $\varepsilon \in (0,1/2)$ and $j=1, \cdots, p$, there exists a positive constant $c_{11}$ such that
    \begin{equation*}
    P\left(\left|\bar{\gamma}_{j, r}-\gamma_{j, r}\right| \geq \varepsilon\right) \leq 6 \exp \left( -c_{11} \sum_{l=1}^m \lfloor \frac{n_l}{2}\rfloor   (\frac{\varepsilon}{R^2} )^{2}\right)
    \end{equation*}
\end{lemma}

\begin{proof}
	From condition (C1), we can derive: $\pi_r^*(1-\pi_r^*)=\vartheta_r \pi_r(1-\pi_r)\ge \frac{b_3/b_1 b_2}{R^2} \triangleq \frac{b_4}{R^2}$
	
\begin{align*}
    P\left(\left|\bar{\gamma}_{j, r}-\gamma_{j, r}\right| \geq \varepsilon\right)&=P\left(\left|\frac{\bar{U}_{j,r}}{\bar{\theta}_{r}}-\frac{{U}_{j,r}^*}{{\theta}_{r}^*}\right| \geq \varepsilon\right)\\
    &=P\left(\left|\frac{\bar{U}_{j,r}}{\bar{\theta}_{r}}-\frac{{U}_{j,r}^*}{{\theta}_{r}^*}\right| \geq \varepsilon, \bar{\theta}_{r}\le \frac{b_4}{2R^2} \right)+P\left(\left|\frac{\bar{U}_{j,r}}{\bar{\theta}_{r}}-\frac{{U}_{j,r}^*}{{\theta}_{r}^*}\right| \geq \varepsilon, \bar{\theta}_{r}>\frac{b_4}{2R^2} \right)\\
    &\le P\left( \bar{\theta}_{r}\le \frac{b_4}{2R^2} \right)+P\left(\left|\frac{\bar{U}_{j,r}-{U}_{j,r}^*}{\bar{\theta}_{r}}-{U}_{j,r}^*\frac{\bar{{\theta}}_{r}-{\theta}_{r}^*}{{\theta}_{r}^*\bar{{\theta}}_{r}}\right| \geq \varepsilon, \bar{\theta}_{r}>\frac{b_4}{2R^2} \right)\\
    & =: I_{1}+I_{2}.
\end{align*}

We first consider $I_1$,
\begin{align*}
    I_1 = P\left(\frac{b_4}{2R^2}  \ge \bar{\theta}_{r}\right)=P\left({\theta}_{r}^*- \bar{\theta}_{r}\ge {\theta}_{r}-\frac{b_4}{2R^2} \right)&\le P\left( \left| \theta_{r}^*- \bar{\theta}_{r} \right|\ge \frac{b_4}{2R^2}\right)\\
    &\le P\left( \left| \theta_{r}^*- \bar{\theta}_{r} \right|\ge \frac{b_4 \varepsilon}{4R^2} \right)
    \end{align*}

We next consider $I_{2}$, 

\begin{align*}
    I_2 &= P\left(\left|\frac{\bar{U}_{j,r}-{U}_{j,r}^*}{\bar{\theta}_{r}}-{U}_{j,r}^*\frac{\bar{{\theta}}_{r}-{\theta}_{r}^*}{{\theta}_{r}\bar{{\theta}}_{r}}\right| \geq \varepsilon, \bar{\theta}_{r}>\frac{b_4}{2 R^2} \right)\\
    &\le P\left(\left|(\bar{U}_{j,r}-{U}_{j,r}^*)-{U}_{j,r}^*\frac{\bar{{\theta}}_{r}-{\theta}_{r}^*}{{\theta}_{r}}\right| \geq \frac{b_4}{2 R^2} \varepsilon \right)\\
    &\le P\left(\left|\bar{U}_{j,r}-{U}_{j,r}^*\right| \geq \frac{b_4}{4 R^2} \varepsilon \right)+P\left(\frac{{U}_{j,r}^*}{{\theta}_{r}^*} \left|\bar{{\theta}}_{r}-{\theta}_{r}^*\right| \geq \frac{b_4}{4 R^2} \varepsilon \right)\\
    & \le  P\left(\left|\bar{U}_{j,r}-{U}_{j,r}^*\right| \geq \frac{b_4}{4 R^2} \varepsilon \right)+P\left(\left|\bar{{\theta}}_{r}-{\theta}_{r}^*\right| \geq \frac{b_4}{4 R^2} \varepsilon \right)
\end{align*}
where we use the property $\frac{{U}_{j,r}^*}{{\theta}_{r}^*} \le 1$

From Lemma \ref{lem:bound},
\begin{align*}
    P\left(\left|\bar{\gamma}_{j, r}-\gamma_{j, r}\right| \geq \varepsilon\right)&\le I_{1}+I_{2}\\
    &\le   P\left(\left|\bar{U}_{j,r}-{U}_{j,r}^*\right| \geq \frac{b_4}{4 R^2} \varepsilon \right)+2P\left(\left|\bar{{\theta}}_{r}-{\theta}_{r}^*\right| \geq \frac{b_4}{4 R^2} \varepsilon \right)\\
    &\le 6 \exp \left( -\sum_{l=1}^m \lfloor \frac{n_l}{2} \rfloor (\frac{b_4}{4 R^2} \varepsilon)^{2}\right)=6 \exp \left( -c_{11} \sum_{l=1}^m \lfloor \frac{n_l}{2}\rfloor   (\frac{\varepsilon}{R^2} )^{2}\right)
\end{align*}
\end{proof}

\begin{lemma}\label{lem:boundomega}
    For any $\varepsilon \in (0,1/2)$ and $j=1, \cdots, p$, there exists a positive constant $c_{11}$ defined in Lemma \ref{lem:boundgamma} such that
    \begin{equation}
    P\left(\left|\bar{\omega}_{j, r}-\omega_{j, r}\right| \geq \varepsilon\right) \leq 6 \exp \left( -c_{11} \sum_{l=1}^m \lfloor \frac{n_l}{2}\rfloor   (\frac{\varepsilon}{R^2} )^{2}\right)
    \end{equation}
\end{lemma}

\begin{proof}[Proof of Lemma \ref{lem:boundomega}]

Notice that $$\left|\bar{\omega}_{j, r}-\omega_{j, r}\right|
=\left|\left|\bar{\gamma}_{j, r}-\frac{1}{2}\right|-\left|\gamma_{j, r}-\frac{1}{2}\right|\right|\le \left|\bar{\gamma}_{j, r}-\gamma_{j, r}\right|$$

From \ref{equ:permu},
\begin{align*}
    P\left(\left|\bar{\omega}_{j, r}-\omega_{j, r}\right| \geq \varepsilon\right) \leq P\left(\left|\bar{\gamma}_{j, r}-\gamma_{j, r}\right| \geq \varepsilon\right) \le  6 \exp \left( -c_{11} \sum_{l=1}^m \lfloor \frac{n_l}{2}\rfloor   (\frac{\varepsilon}{R^2} )^{2}\right)
\end{align*}    
    Then, we complete the proof of Lemma \ref{lem:boundomega}.
\end{proof}

\begin{proof}[Proof of Proposition \ref{pro:omega_bound}]
    From Lemma \ref{lem:boundomega}, by setting $\varepsilon=c_1 N^{-\kappa}$ for $0 \leq \kappa<1 / 2$, we have
    \begin{align*}
    P\left(\max _{1 \leq j \leq p}\left|\bar{\omega}_{j, r}-\omega_{j, r}\right| \geq c_1 N^{-\kappa}\right)&\leq p P\left(\left|\bar{\omega}_{j, r}-\omega_{j, r}\right| \geq c_1 N^{-\kappa}\right) \\
& \leq 6p \exp \left( -c_2 N^{1-2\kappa-4\xi } \right)
    \end{align*}
    
     We have completed the proof of Proposition \ref{pro:omega_bound}.
\end{proof}

\begin{proof}[Proof of Proposition \ref{pro:variance}]
    Drawing from the proof technique presented in Proposition 4 of \citet{li_feature_2023}, it is straightforward to establish the orders of variance of $\bar{\theta}_{r}$ and $\bar{U}_{j,r}$.

    Notice that 
    $\left|\bar{\omega}_{j,r}-{\omega}_{j,r}\right|\le \left|\bar{\gamma}_{j,r}-{\gamma}_{j,r}\right|$ and 
    
    \begin{align*}
        \left(\bar{\gamma}_{j,r}-{\gamma}_{j,r}\right)^2&=\left(\frac{\bar{U}_{j,r}}{\bar{\theta}_{r}}-\frac{{U}_{j,r}^*}{{\theta}_{r}^*}\right)^2=\left(\frac{\bar{U}_{j,r}}{\bar{\theta}_{r}}-\frac{\bar{U}_{j,r}}{{\theta}_{r}^*}+\frac{\bar{U}_{j,r}}{{\theta}_{r}^*}-\frac{{U}_{j,r}^*}{{\theta}_{r}^*}\right)^2\\
        &\le 2\left(\frac{\bar{U}_{j,r}}{\bar{\theta}_{r}}-\frac{\bar{U}_{j,r}}{{\theta}_{r}^*}\right)^2 +2 \left(\frac{\bar{U}_{j,r}}{{\theta}_{r}^*}-\frac{{U}_{j,r}^*}{{\theta}_{r}^*}\right)^2\\
        &= 2\left(\frac{\bar{U}_{j,r}}{\bar{\theta}_{r}}\frac{{\theta}_{r}^*-\bar{\theta}_{r}}{{\theta}_{r}^*}\right)^2 +2 \left(\frac{\bar{U}_{j,r}-{U}_{j,r}^*}{{\theta}_{r}^*}\right)^2
    \end{align*}

    Then 
    \begin{align*}
        \mathbb{E}(\bar{\omega}_{j,r}-{\omega}_{j,r})^2
    &\le 2\mathbb{E}\left(\frac{\bar{U}_{j,r}}{\bar{\theta}_{r}}\frac{{\theta}_{r}^*-\bar{\theta}_{r}}{{\theta}_{r}^*}\right)^2+2\mathbb{E}\left(\frac{\bar{U}_{j,r}-{U}_{j,r}^*}{{\theta}_{r}^*}\right)^2\\
    &\le 2\frac{\mathrm{var}(\bar{\theta}_{r})}{{\theta}_{r}^{*2}}+2\frac{\mathrm{var}(\bar{U}_{r})}{{\theta}_{r}^{*2}}\\
    & =  O(N^{4\xi-1})
    \end{align*}
\end{proof}

\begin{proof}[Proof of Theorem \ref{the:surescreening} and \ref{the:rank}]
    Following the lines of the proofs of Proposition \ref{pro:omega_bound},
    \begin{align*}
    &P\left( \underset{1\le j\le p}{\max}|\bar{\omega}_j-\omega _j|\ge c_3 N^{-\kappa} \right) \le pP\left( |\bar{\omega}_j-\omega _j|\ge c_3 N^{-\kappa} \right)\\
    &\le pP\left( \underset{1\le r\le R}{\max}|\bar{\omega}_{j,r}-\omega _{j,r}|\ge c_3 N^{-\kappa} \right) \le pRP\left( |\bar{\omega}_{j,r}-\omega _{j,r}|\ge c_3 N^{-\kappa} \right)  \\
    &\le 6pR \exp \left( -c_4 N^{1-2\kappa-4\xi } \right).
    \end{align*}
    
    Additionally, under conditions (C2) and (C3), we have 
    $$P\left( \underset{1\le j\le p}{\max}|\bar{\omega}_j-\omega _j|\ge c_3 N^{-\kappa} \right)\leq  6p \exp \left( -c_4 N^{1-2\kappa-4\xi } +\xi \log(N)\right)$$

    If $\mathcal{A} \not \subset \hat{\mathcal{A}}$, then there must exist $k \in \mathcal{A}$ such that $\bar{\omega}_{j}<c N^{-\kappa}$.
    
    Furthermore, when $\max _{j \in \mathcal{A}}\left|\bar{\omega}_{j}-\omega_{j}\right| \leq c N^{-\kappa}$ and condition (C4) holds,
    \begin{equation*}
    \min _{j \in \mathcal{A}}\bar{\omega}_{j} \geq \min _{j \in \mathcal{A}}\left(\omega_{j}-\left|\bar{\omega}_{j}-\omega_{j}\right|\right) \geq \min _{j \in \mathcal{A}}\omega_{j}-\max _{j \in \mathcal{A}}\left|\bar{\omega}_{j}-\omega_{j}\right| \geq c N^{-\kappa}.
    \end{equation*}
    
    Therefore,
    \begin{equation}
    P\left(\mathcal{A} \subset \hat{\mathcal{A}}\right) \geq P\left(\max _{ j \in \mathcal{A}}\left|\bar{\omega}_{j}-\omega_{j}\right| \leq c N^{-\kappa}\right) \geq 1-6 s \exp \left( -c_5 N^{1-2\kappa-4\xi } \right).
    \end{equation}
    for some constant $c_5>0$. We have completed the proof of Theorem \ref{pro:omega_bound}.
\end{proof}

\begin{proof}[Proof of Theorem \ref{the:rank}]
    Define $\Lambda=\min _{j \in \mathcal{A}}\omega_{j}-\max _{j \in \mathcal{I}}\omega_{j}$. From condition (C4) and Lemma \ref{lem:boundomega}.
    \begin{align*}
    & P\left(\min _{j \in \mathcal{A}}\bar{\omega}_{j} \leq \max _{j \in \mathcal{I}}\bar{\omega}_{j}\right)=P\left(\min _{j \in \mathcal{A}}\bar{\omega}_{j}-\min _{j \in \mathcal{A}}\omega_{j}+\Lambda \leq \max _{j \in \mathcal{I}}\bar{\omega}_{j}-\max _{j \in \mathcal{I}}\omega_{j}\right) \\
    & = P\left(\left[\max _{j \in \mathcal{I}}\bar{\omega}_{j}-\max _{j \in \mathcal{I}}\omega_{j}\right]-\left[\min _{j \in \mathcal{A}}\bar{\omega}_{j}-\min _{j \in \mathcal{A}}\omega_{j}\right] \geq \Lambda\right) \\
    & \leq P\left(\max _{j \in \mathcal{I}}\left|\bar{\omega}_{j}-\omega_{j}\right|+\min _{j \in \mathcal{A}}\left|\bar{\omega}_{j}-\omega_{j}\right| \geq \Lambda\right) \\
    & \leq P\left(2 \max _{1 \leq j \leq p}\left|\bar{\omega}_{j}-\omega_{j}\right| \geq \Lambda\right) \leq pR P\left(\left|\bar{\omega}_{j}-\omega_{j}\right| \geq \frac{\Lambda}{2}\right) \\
    &\leq pR P\left(\left|\bar{\omega}_{j}-\omega_{j}\right| \geq c N^{-\eta}\right)  \leq  6 pR \exp \left( -c_6 N^{1-2\eta-4\xi } \right)
    \end{align*}

    Then,
    \begin{equation}
    P\left(\min _{j \in \mathcal{A}}\bar{\omega}_{j}>\max _{j \in \mathcal{I}}\bar{\omega}_{j}\right) \geq 1-6 pR \exp \left( -c_6 N^{1-2\eta-4\xi } \right)
    \end{equation}
    holds for some constant $c_6>0$. We have completed the proof of Theorem \ref{the:rank}.	
\end{proof}

\begin{proof}[Proof of Theorem \ref{the:FDRcontrol}]
 When $\max_j\left|\bar{\omega}_{j}-\omega_{j}\right| \geq c N^{-\kappa}$ and condition (C4) holds, the number of $\left\{j:\bar{\omega}_{j} \geq c N^{-\kappa}\right\}$ can not exceed the number of $\left\{j:\omega_{j}\ \geq c N^{-\kappa}/2\right\}$, which is bounded by $\left(c / 2\right)^{-1} N^\kappa \sum_{j=1}^p\omega_{j}$. Therefore,
    \begin{align*}
P\left\{\left|\hat{\mathcal{A}}\right| \leq\left(c / 2\right)^{-1} N^\kappa \sum_{j=1}^p \omega_{j}\right\} &\ge  P\left\{\max_{j}\left|\bar{\omega}_{j}-\omega_{j}\right|
\ge c N^{-\kappa} \right\} \\ 
&\geq 1-6pR \exp \left( -c_{7} N^{1-2\kappa-4\xi } \right)
\end{align*}
    holds for some constant $c_{7}>0$. We have completed the proof of Theorem \ref{the:FDRcontrol}.	
\end{proof}

\begin{proof}[Proof of Theorem \ref{the:FDRalpha}]
We have already demonstrated that the estimates of $\omega_{j}$ and $\omega_{j}^\prime$ have the same efficiency as the non-distributed estimates, thus ensuring that $\phi_j$ also maintains the corresponding estimation properties. The remaining proof can be completed using the same arguments as those in the proof of \citet{tong2023model}.
\end{proof}

The proof of the properties of the general framework will be shown in section \ref{sec:general}.

\section{Proof of feature screening properties under the general framework}\label{sec:general}
\subsection{Estimation process within the general framework}\label{subsec:general_procedure}
Next, we present the estimation procedure under the general framework, detailing how to estimate $\zeta_r$ and $\omega_{j,r,d}$ when $d > 1$.

$$\omega_j^{(d)}=\sum_{r=1}^R{\zeta_r {\omega_{j,r,d}^k}},$$
where  $\omega_{j,r,d}=\left| \mathbb{E}_{Y=y_r}\left( \left( F_{Y\ne y_r}\left( X_j \right) -F_{Y=y_r}\left( X_j \right) \right) ^d \right) \right| $,  $k$  is the exponent, $\zeta_r=g(\pi_r)$ is a function of $\pi_r$. 
 
Referring back to Section \ref{subsec:general}, only when $d = 1$ can $ \mathbb{E}_{Y=y_r} \left( F_{Y \ne y_r} (X_j) \right) $ and $ \mathbb{E}_{Y=y_r} \left( F_{Y = y_r} (X_j) \right) $ be separated, avoiding the introduction of interaction terms and reducing estimation complexity. In the general estimation process, different clients need to collaboratively estimate $\omega_{j,r}$ as shown in Algorithm \ref{alg:procedure}. When $ d > 1 $, estimating $\omega_{j,r,d}$ becomes relatively more complex. Regardless of the value of $ d $, they also need to estimate $ \pi_r $ and $\zeta_r$ jointly. Estimating  $ \pi_r $ and $\zeta_r$ is straightforward and involves the following steps:

\begin{enumerate}
	\item for $r=1,\cdots,R$, on the $l$-th client, $\pi_{r} $ can be estimated by $\hat{\pi}_{r}^l=\frac{\sum I\left(Y_{i}^l=y_{r}\right)}{n_{l}}$.
	
	\item When the estimates on each client are transmitted to the central computer, ${\pi}_{r}$  can be expressed as $\bar{\pi}_{r}=\frac{\sum_l n_l\hat{\pi}_{r}^l}{\sum n_l}$ and $\zeta_r$ can be estimated through $\bar{\eta}_r = g(\bar{\pi}_r)$.

	\item In the central computer, the estimation of $\omega_{j}$ is conducted using $\bar{\omega}_{j} = \sum_r \bar{\zeta}_r\bar{\omega}_{j,r,d}$. The remaining steps are similar to Algorithm  \ref{alg:procedure}.
\end{enumerate}

Next, we analyze how to estimate $\omega_{j,r,d} $ when $d > 1$.  Using the binomial expansion, $$\left( F_{Y \ne y_r}(X_j) - F_{Y=y_r}(X_j) \right)^d = \sum_{d_1=0}^{d} \binom{d}{d_1}(-1)^{d-d_1} F_{Y \ne y_r}(X_j)^{d_1} F_{Y=y_r}(X_j)^{d-d_1}, $$ from which we need to estimate $ \gamma_{j,r,d,d_1} = \mathbb{E}_{Y=y_r} \left[ F_{Y \ne y_r}(X_j)^{d_1} F_{Y=y_r}(X_j)^{d-d_1} \right] $ for $d_1 = 1, \cdots, d$. When $d_1=0$, we derive $\mathbb{E}_{Y=y_r} \left[ F_{Y \ne y_r}(X_j)^{d_1} F_{Y=y_r}(X_j)^{d-d_1} \right] =\frac{1}{d+1}$

Similar to section \ref{subsec:federated feature screening}, to estimate $ \mathbb{E}_{Y=y_r} \left[ F_{Y \ne y_r}(X_j)^{d_1} F_{Y=y_r}(X_j)^{d-d_1} \right] $, defined as $\gamma_{j,r,d,d_1}$, we can decompose it into two components and estimate them separately: $\gamma_{j,r,d,d_1} = \frac{U_{j,r,d,d_1}}{\theta_{r,d,d_1}}$, where
\begin{align*}
\theta_{r,d,d_1} &= \mathbb{E}\left[\mathbb{E}\left(I(Y_{i_1}\ne y_r)\right)^{d_1}\mathbb{E}\left(I(Y_{i_2}=y_r)\right)^{d-d_1}I(Y_k=y_r)\right],\\
U_{j,r,d,d_1} &= \mathbb{E}\left[\mathbb{E}\left[I\left( X_{j,i_1}<X_{j,k} \right) I(Y_{i_1}\ne y_r)\right]^{d_1}\mathbb{E}\left[I\left( X_{j,i_2}<X_{j,k} \right) I(Y_{i_2}=y_r)\right]^{d-d_1}I(Y_k= y_r)\right].
\end{align*}

To estimate $\theta_{r,d,d_1}$ and $U_{j,r,d,d_1}$, we construct a U-statistic involving a $(d+1)$-variate kernel related to $d_1$, denoted as $\tilde{\theta}_{r,d,d_1}(Z_{i_1,j}, \cdots, Z_{i_{d+1},j})$ and $\tilde{U}_{j,r,d,d_1}(Z_{i_1,j}, \cdots, Z_{i_{d+1},j})$. We summarize the algorithm to estimate $\theta_{r,d,d_1}$ and $U_{j,r,d,d_1}$ in the following steps:

\begin{enumerate}
\item For any $r=1,\cdots,R$, on the $l$-th client, for $d_1=1,\cdots,d$,$\theta_{r,d,d_1}$ and $U_{j,r,d,d_1}$ are estimated using local U-statistics:
\begin{align}
\hat{U}_{j,r,d,d_1}^l &= \binom{n_l}{d+1}^{-1} \sum_{\{i_1, \cdots, i_{d+1}\} \in \mathcal{S}_l} \tilde{U}_{j,r,d,d_1}(Z_{i_1,j}, \cdots, Z_{i_{d+1},j}),\\
\hat{\theta}_{r,d,d_1}^l &= \binom{n_l}{d+1}^{-1} \sum_{\{i_1, \cdots, i_{d+1}\} \in \mathcal{S}_l} \tilde{\theta}_{r,d,d_1}(Z_{i_1,j}, \cdots, Z_{i_{d+1},j}),
\end{align}
where the summation is over all combinations of $\{Z_{i_1,j}, \ldots, Z_{i_{d+1},j}\}$ chosen from $\mathcal{D}_l$, and $\mathcal{S}_l$ denotes the index set of observations in $\mathcal{D}_l$.

\item Each client sends the parameters $\{\hat{U}_{j,r,d,d_1}^l, \hat{\theta}_{r,d,d_1}^l\}_{d_1=1}^d$ and $n_l$ to the central server. The central server aggregates the parameters as follows:
\begin{equation*}
\bar{U}_{j,r,d,d_1} = \frac{\sum_l^m h_l \hat{U}_{j,r,d,d_1}^l}{\sum_l^m h_l}, \bar{\theta}_{r,d,d_1} = \frac{\sum_l^m h_l \hat{\theta}_{r,d,d_1}^l}{\sum_l^m h_l}, \bar{\gamma}_{j,r,d,d_1} = \frac{\bar{U}_{j,r,d,d_1}}{\bar{\theta}_{r,d,d_1}},
\end{equation*}
where $h_l = \lfloor \frac{n_l}{d+1}\rfloor$.

\item The central server calculates the final $\bar{\omega}_{j,r,d}$ using:
$$\bar{\omega}_{j,r,d}= \left|\sum_{d_1=1}^{d} \binom{d}{d_1}(-1)^{d-d_1} \bar{\gamma}_{j,r,d,d_1}+(-1)^{d}\frac{1}{d+1}\right|.$$

The remaining steps are similar to the previous content and will not be repeated here.
\end{enumerate}

We illustrate our kernel with a simple example: when $d = 2 $ and $d_1 = 0$,  the kernel for estimating $U_{j,r,d,d_1}$ and $\theta_{r,d,d_1}$ are
\begin{align*}
\tilde{U}_{j,r,d,d_1}(Z_{i_1,j}, Z_{i_2,j}, Z_{i_3,j})& = I(X_{j,i_1} < X_{j,i_3}) I(Y_{i_1} \ne y_r) I(X_{j,i_2} < X_{j,i_3}) I(Y_{i_2} \ne y_r) I(Y_{i_3} = y_r),\\
\tilde{\theta}_{r,d,d_1}(Z_{i_1,j}, Z_{i_2,j}, Z_{i_3,j})& = I(Y_{i_1} \ne y_r)I(Y_{i_2} \ne y_r) I(Y_{i_3} = y_r).
\end{align*}

when $d = 1 $ and $d_1 = 1$,  the kernel for estimating $U_{j,r,d,d_1}$ and $\theta_{r,d,d_1}$ are
\begin{align*}
\tilde{U}_{j,r,d,d_1}(Z_{i_1,j}, Z_{i_2,j})& = I(X_{j,i_1} < X_{j,i_2}) I(Y_{i_1} \ne y_r) I(Y_{i_2} = y_r),\\
\tilde{\theta}_{r,d,d_1}(Z_{i_1,j}, Z_{i_2,j})& = I(Y_{i_1} \ne y_r)I(Y_{i_2} = y_r).
\end{align*}

After introducing the estimation under the general framework, we will proceed with the proof of the related screening properties.

\subsection{Theoretical analysis for the general framework}
Before presenting the proof of the related theorems and some technical lemmas, we need to introduce a bridge parameter similar to equation \ref{equ:bridge}. For any $ d $ and $ d_1$, based on the proportion relationship among clients, there exists $ \pi_{r,d,d_1}$ satisfying
\begin{equation}
\sum_{l=1}^m h_l \left( \pi_{r,l} \right)^{d-d_1+1} \left( 1 - \pi_{r,l} \right)^{d_1} = \sum_{l=1}^m h_l \left( \pi_{r,d,d_1} \right)^{d-d_1+1} \left( 1 - \pi_{r,d,d_1} \right)^{d_1}, \label{equ:general_hl}
\end{equation}
where $\pi_{r,l}$ denotes the proportion of category $Y=y_r $ on the $ l $-th client, $h_l$ is related to sample size and we adopt $\left\lfloor \frac{n_l}{d+1} \right\rfloor$ here. Similarly, we define $ \theta_{r,d,d_1}^* $ and $U_{j,r,d,d_1}^*$ as follows:
\begin{equation*}
\theta_{r,d,d_1}^* = \pi_{r,d,d_1}^{d-d_1+1} \left( 1 - \pi_{r,d,d_1} \right)^{d_1}, U_{j,r,d,d_1}^* = \gamma_{j,r,d,d_1} \theta_{r,d,d_1}^*.
\end{equation*}

Theorems \ref{the:surescreening_general} and \ref{the:rank_general} demonstrate that under the general framework, the utilities still satisfy the classic Sure screening property and Ranking consistency property, thereby endowing them with feature screening capabilities even in the presence of label shift. Proposition \ref{pro:general} indicates that the maximum variance of the parameter estimates increases as both $d$ and $m$ increase. Before proving these theorems, we need to provide proofs for several lemmas that differ from those in the main text. Lemmas \ref{lem:lemmaG3} and \ref{lem:lemmaG4}, similar to Lemmas \ref{lem:bound} and \ref{lem:boundgamma}, provide bounds for estimating $\bar{\gamma}_{j,r,d,d_1}$. Lemma \ref{lem:lemmaG5} provides the bound for $\bar{\omega}_{j, r,d}$. Lemma \ref{lem:pi} proves the estimation properties of $\pi_r$, while Lemma \ref{lem:C_boundg} establishes the estimation properties of $\zeta_r$. When  $d = 1$, we can directly derive the bound for LR-FFS.

\begin{proposition}\label{pro:general}
    Similar to proposition \ref{pro:variance}, the variances of $\bar{\theta}_{r,d,d_1}$ and $\bar{U}_{j,r,d,d_1}$ can be expanded as
        \begin{align*}
        \max_{r} \mathrm{var}(\bar{\theta}_{r,d,d_1}) &= O(\frac{1}{N})+O(\frac{m}{N^2})+\cdots+O(\frac{m^d}{N^{d-1}}),\\
        \max_{j,r} \mathrm{var}(\bar{U}_{j,r,d,d_1}) &= O(\frac{1}{N})+O(\frac{m}{N^2})+\cdots+O(\frac{m^d}{N^{d-1}}).
    \end{align*}
    Moreover, under the condition(C1), (C3) and $m = O(N)$, the mean squared error of $\bar{\omega}_{j,r,d}$ has the
following uniform order
$$\max_{j,r}\mathrm{MSE}(\bar{\omega}_{j,r,d}) = \mathbb{E}(\bar{\omega}_{j,r,d}-{\omega}_{j,r,d})^2=O(N^{4\xi-1}).$$    
\end{proposition}

\begin{lemma} \label{lem:lemmaG3}
	For any $\varepsilon>0$ and $j \in \{1,\cdots,p\}$, we have
	\begin{align}
	P\left(\left|\bar{\theta}_{r,d,d_1} -\theta_{r,d,d_1}^* \right| \geq \varepsilon\right) \leq 2 \exp \left( -\sum_{l=1}^m \lfloor \frac{n_l}{d+1} \rfloor \varepsilon ^{2}\right)\\
	P\left(\left|\bar{U}_{j,r,d,d_1}-U_{j,r,d,d_1}^*\right| \geq \varepsilon\right) \leq 2 \exp \left( -\sum_{l=1}^m \lfloor \frac{n_l}{d+1} \rfloor \varepsilon ^{2}\right)
	\end{align}
\end{lemma}

\begin{lemma} \label{lem:lemmaG4}
	Suppose condition (C1) hold. For any $\varepsilon \in (0,1/2)$ and $j=1, \cdots, p$, there exists a positive constant $t_1$ such that
	\begin{equation*}
	P\left(\left|\bar{\gamma}_{j,r,d,d_1}-{\gamma}_{j,r,d,d_1}\right| \geq \varepsilon\right) \leq 6 \exp \left( -t_1 N  (\frac{\varepsilon}{R^2} )^{2}\right).
	\end{equation*}
	
	When $R$ is fixed, it can be derived that there exists a positive constant $t_2$ such that
	\begin{equation*}
	P\left(\left|\bar{\gamma}_{j,r,d,d_1}-{\gamma}_{j,r,d,d_1}\right| \geq \varepsilon\right) \leq 6 \exp \left( -t_2 N \varepsilon^{2}\right).
	\end{equation*}
\end{lemma}

\begin{lemma}\label{lem:lemmaG5}
	For any $\varepsilon \in (0,1/2)$ and $j=1, \cdots, p$, there exists a positive constant $t_3$ such that
	\begin{equation}
	P\left(\left|\bar{\omega}_{j, r,d}-\omega_{j, r,d}\right| \geq \varepsilon\right) \leq 6 (2^d-1)\exp \left( -t_3 N   (\frac{\varepsilon}{R^2} )^{2}\right).
	\end{equation}
	
	When $R$ is fixed, it can be derived that there exists a positive constant $t_4$ such that
	\begin{equation*}
	P\left(\left|\bar{\omega}_{j, r,d}-\omega_{j, r,d}\right| \geq \varepsilon\right) \leq 6 (2^d-1)\exp \left( -t_4 N   \varepsilon^{2}\right).
	\end{equation*}
\end{lemma}

\begin{proof}[Proof of Lemma \ref{lem:lemmaG3} and \ref{lem:lemmaG4}]
	The proof is similar to the proof of lemma \ref{lem:bound}, we first prove the first conclusion and bound the term $\left|\bar{\theta}_{r,d,d_1} -\theta_{r,d,d_1}^* \right|$. 
	
	Let $\tilde{\theta}_{r,d,d_1}^l(Z_{i_1,j}, \cdots, Z_{i_{d+1},j})$ be a basis unbiased estimator of $\theta_{r,d,d_1}^l$ with degree $d+1$. 
	
	 Recall that $\mathcal{S}_{l}=\left\{l_{1}, \ldots, l_{n_{l}}\right\}$ denotes the index set of $\{Y, \mathbf{X}\}$ copies based on $\mathcal{D}_{l}$, on which we can construct $v_l = \left\lfloor \frac{n_l}{d+1} \right\rfloor$ independent $\tilde{\theta}_{r,d,d_1}^l$s. Then, we can similarly define an averaged estimator based on these independent $\tilde{\theta}_{r,d,d_1}^l$ by
	\begin{align}
	&V_{r,d,d_1}^l\left(Z_{l_{1}}, \ldots, Z_{l_{n_{l}}}\right)=\frac{1}{v_{l}} \sum_{u=1}^{v_{l}} \tilde{\theta}_{r,d,d_1}^l\left(Z_{l_{(d+1)(u-1)+1}}, Z_{l_{(d+1) u}}\right),\\
	&\hat{\theta}_{r,d,d_1}^l=\frac{1}{n_{l}!} \sum_{\left\{i_{1}, \ldots, i_{n_{l}}\right\} \in \Omega_{l}} V_{r,d,d_1}^l \left(Z_{l_{i_{1}}}, \ldots, Z_{l_{i_{l}}}\right)
	\end{align}
	
	where $\Omega_{l}=\left\{1, \ldots, n_{l}\right\}$ and the summation is over all $\left\{Z_{l_{i_{1}}}, \ldots, Z_{l_{i_{l}}}\right\}$ permutations from $\mathcal{D}_{l}$. 
	
	Consequently,
	\begin{equation}
	\bar{\theta}_{r,d,d_1}=\frac{\sum_l^m h_l \hat{\theta}_{r,d,d_1}^l}{\sum_l^m h_l}=\frac{1}{\sum_{l=1}^{m} h_{l}}
	\sum_{l=1}^{m} \sum_{\left\{i_{1}, \ldots, i_{n_{l}}\right\} \in \Omega_{l}} \frac{h_{l}}{n_{l}!} V_{r,d,d_1}^l \left(Z_{l_{i_{1}}}, \ldots, Z_{l_{i_{l}}}\right)
	\end{equation}
	
	Combining with the definition of $\theta_{r,d,d_1}^*$ from Equation \ref{equ:general_hl}, through Markov's and Jensen's inequalities, we obtain
	\begin{align*}
	P\left(\bar{\theta}_{r,d,d_1} -\theta_{r,d,d_1}^*  \geq \varepsilon\right) & =P\left(\exp \left\{\nu\left( \bar{\theta}_{r,d,d_1} -\theta_{r,d,d_1}^* \right)\right\} \geq \exp \{\nu \varepsilon\}\right) \\
	& \leq \exp \{-\nu \varepsilon\} \exp \left\{-\nu \theta_{r,d,d_1}^* \right\} \mathbb{E}\left[\exp \left\{\nu \bar{\theta}_{r,d,d_1}\right\}\right]\\
	&\leq \prod_{l=1}^m \left[\exp \{-\tau \varepsilon\} \exp \left\{-\tau \theta_{r,d,d_1}^l\right\}\exp \left\{\tau\tilde{\theta}_{r,d,d_1}^l \right\} \right]^{h_l},
	\end{align*}
	where $\tau=\nu /\left(\sum_{l=1}^{m} h_l\right)$. 
	
	The remaining proof is similar to Lemma \ref{lem:bound} and \ref{lem:boundgamma}, we will not elaborate further. So, we have completed the proof of Lemma \ref{lem:lemmaG3} and  \ref{lem:lemmaG4}.
\end{proof}

Note that in our proof, we utilized the fundamental facts that $\tilde{\theta}_{r,d,d_1}^l \in [0,1]$ and $\mathbb{E}(\tilde{\theta}_{r,d,d_1}^l) = \theta_{r,d,d_1}^l$. In fact, by applying Hölder's inequality, we can obtain $\tilde{\theta}_{r,d,d_1}^l \in \left[0, \frac{d_1^{d_1} (d-d_1)^{d-d_1}}{d^d}\right]$. Based on this fact, we can bound $\tilde{\theta}_{r,d,d_1}^l$ more tightly relative to $d_1$. However, this refinement does not affect the order of the bound. For simplicity, we opted to use a uniform bound in our analysis.

\begin{proof}[Proof of Lemma \ref{lem:lemmaG5}]
	Notice that  $\sum_{d_1=1}^{d} \binom{d}{d_1} =2^d-1$ and	
	\begin{align*}
	\left|\bar{\omega}_{j,r,d}-{\omega}_{j,r,d}\right|
	&=\left|\left|\sum_{d_1=1}^{d} \binom{d}{d_1}(-1)^{d_1} \bar{\gamma}_{j,r,d,d_1}+\frac{1}{d+1}\right|-\left|\sum_{d_1=1}^{d} \binom{d}{d_1}(-1)^{d_1} \bar{\gamma}_{j,r,d,d_1}+\frac{1}{d+1}\right|\right|\\
	&\le \sum_{d_1=1}^{d} \binom{d}{d_1}  \left|\bar{\gamma}_{j,r,d,d_1}-{\gamma}_{j,r,d,d_1}\right|.
	\end{align*}
	
	Then
	\begin{align*}
	&P\left(\left|\bar{\omega}_{j,r,d}-{\omega}_{j,r,d}\right| \geq \varepsilon\right) \leq P\left(\sum_{d_1=1}^{d} \binom{d}{d_1}  \left|\bar{\gamma}_{j,r,d,d_1}-{\gamma}_{j,r,d,d_1}\right| \geq \varepsilon\right) \\
	&\le  \sum_{d_1=1}^{d} \binom{d}{d_1}  P\left( \left|\bar{\gamma}_{j,r,d,d_1}-{\gamma}_{j,r,d,d_1}\right| \geq \frac{\varepsilon}{2^d-1}\right) \le 6 (2^d-1)\exp \left( -t_3 N   (\frac{\varepsilon}{R^2} )^{2}\right).
	\end{align*} 
	
 So, we have completed the proof of Lemma \ref{lem:lemmaG5}.
\end{proof}

Lemmas \ref{lem:pi} to \ref{lem:C_boundg} adopt a straightforward-to-general approach to provide an analysis of the bound for $g(\bar{\pi_r}) = \bar{\zeta}_r$.

\begin{lemma}\label{lem:pi}
	For any $\varepsilon >0$ and $r=1, \cdots, R$, we have 
	\begin{equation}
	P\left( |\bar{\pi}_r-\pi _r|\ge \varepsilon \right) \le 2\exp \left( -2N\varepsilon ^2 \right)
	\end{equation}
\end{lemma}

\begin{lemma}\label{lem:pi2}
	Suppose condition (C1) hold. For any $\varepsilon >0$ and $r=1, \cdots, R$, there exists a positive constant $t_5$ such that
	\begin{align*}
	P\left( \bar{\pi}_r<b_1/2R \right) &\le 2\exp \left( -2t_5 N \right)\\
	P\left( \bar{\pi}_r>1-b_2/2R \right) &\le 2\exp \left( -2t_5 N \right) 
	\end{align*}
\end{lemma}

\begin{lemma}\label{lem:C_boundg}
	Suppose condition (C1) holds. For any continuous function $g(x)$ where $ x \in (0,1) $ , there exists a positive constant $t_6$ such that
	\begin{equation}
	P\left( \left|g\left( \bar{\pi}_r \right) -g\left( \pi _r \right) \right|\ge \varepsilon\right)  \le 6\exp \left( -2t_6	N\varepsilon ^2 \right), r=1, \cdots, R
	\end{equation}
	for any $\varepsilon \in (0,1)$
\end{lemma}

\begin{proof}[Proof of Lemma \ref{lem:pi} and \ref{lem:pi2}]
	Lemma \ref{lem:pi} can be directly derived from Lemma \ref{lem:hoeffding}.
	
	Given Condition (C1), it is noted that $\bar{\pi}_r \le  |\bar{\pi}_r-\pi _r|+\pi _r\le  |\bar{\pi}_r-\pi _r|+b_1/R$,
	
	then,
	$$P\left( \bar{\pi}_r<b_1/2R \right) \le P\left( |\bar{\pi}_r-\pi _r|>b_1/2R \right) \le 2\exp \left( -2N{b_1}^2/4R^2 \right) .$$
	
	Similarly,
		$$P\left( \bar{\pi}_r>1-b_2/2R \right)\le P\left( |\bar{\pi}_r-\pi _r|>b_2/2R \right)\le 2\exp \left( -2N{b_2}^2/4R^2 \right).$$
		
	Therefore, there exists a positive constant $t_5$ such that 
	$$P\left( \bar{\pi}_r<b_1/2R \right) \le 2\exp \left( -2t_5N \right), P\left( \bar{\pi}_r>1-b_2/2R \right) \le 2\exp \left( -2t_5N \right).$$
	
	We have completed the proofs of Lemma \ref{lem:pi} and \ref{lem:pi2}.
\end{proof}

\begin{proof}[Proof of Lemma \ref{lem:C_boundg}]
In the closed interval $[b_1/2R, 1-b_2/2R]$, for any continuous function $g(x)$, there exists a constant $L_g$ such that for any $x_1, x_2 \in [b_1/2R, 1-b_2/2R]$, we have $|g(x_1) - g(x_2)| \leq L_g |x_1 - x_2|$.

From Lemma \ref{lem:pi2},
\begin{align*}
P\left( \left|g\left( \bar{\pi}_r \right) -g\left( \pi _r \right) \right| \ge \varepsilon \right) &\le P\left( \left|g\left( \bar{\pi}_r \right) -g\left( \pi _r \right) \right|\ge \varepsilon ,b_1/2R\le \bar{\pi}_r\le 1-b_2/2R \right)\\ 
&+P\left( \bar{\pi}_r<b_1/2R \right) +P\left( \bar{\pi}_r>1-b_2/2R \right)\\
&\le P\left( L_g \left|\bar{\pi}_r-\pi _r\right|\ge \varepsilon \right) +P\left( \bar{\pi}_r<b_1/2R \right) +P\left( \bar{\pi}_r>1-b_2/2R \right) \\
&\le 2\exp \left( -2/L_g^2N\varepsilon ^2 \right) +4\exp \left( -2t_5 N \right) \le 6\exp \left( -2t_6N\varepsilon ^2 \right) 
\end{align*}
	
We have completed the proofs of Lemma \ref{lem:C_boundg}.
\end{proof}

\begin{lemma}\label{lem:C_boundomegad}
	For any $\varepsilon >0,k>1$ and $j=1, \cdots, p$, there exists a positive constant $t_7$  such that
	\begin{equation*}
	P\left(\left|\bar{\omega}_{j, r,d}^k-\omega_{j, r,d}^k\right| \geq \varepsilon\right) \leq 6(2^d-1) \exp \left( -2t_7 N   \varepsilon^2\right)
	\end{equation*}
\end{lemma}

\begin{proof}[Proof of Lemma \ref{lem:C_boundomegad}]
	According to Lagrange's Mean Value Theorem, there exists $ \tilde{\omega}_{j, r,d} \in (\bar{\omega}_{j, r,d}\land {\omega}_{j, r,d} , \bar{\omega}_{j, r,d}\lor {\omega}_{j, r,d}) $ such that $ \bar{\omega}_{j, r,d}^k - {\omega}_{j, r,d}^k =(\bar{\omega}_{j, r,d}-{\omega}_{j, r,d}) k \tilde{\omega}_{j, r,d}^{k-1}$. Then  $$\left|\bar{\omega}_{j, r,d}^k - {\omega}_{j, r,d}^k \right|\le k \left|\bar{\omega}_{j, r,d}-{\omega}_{j, r,d}\right|. $$
	
	From Lemma \ref{lem:boundomega},
	\begin{equation*}
	P\left(\left|\bar{\omega}_{j, r,d}^k-\omega_{j, r,d}^k\right| \geq \varepsilon \right) \le P\left(k \left|\bar{\omega}_{j, r,d}-{\omega}_{j, r,d}\right| \geq \varepsilon\right) \leq 6(2^d-1) \exp \left( -2t_7 N   \varepsilon^2\right).
	\end{equation*}
	We have completed the proofs of Lemma  \ref{lem:C_boundomegad}.
\end{proof}

\begin{lemma}\label{lem:C_boundomega}
	For any $\varepsilon \in (0,1),k>1$ and $j=1, \cdots, p$, there exists a positive constant $t_8$  such that
	\begin{equation*}
	P\left(\left|\bar{\omega}_{j}^{(d)}-\omega_{j}^{(d)}\right| \geq \varepsilon\right) \leq 12(2^d-1)R\exp \left( -2t_8 N\varepsilon ^2 \right),
	\end{equation*}
	where $\omega _j^{(d)}=\sum_{r=1}^R{\zeta_r \omega_{j,r,d}^k}$.
\end{lemma}

\begin{proof}[Proof of Lemma \ref{lem:C_boundomega}]
From Lemma \ref{lem:C_boundg} and \ref{lem:C_boundomegad},	
\begin{align*}
P\left( |\sum_{r=1}^R{\bar{\eta}_r\bar{\omega}_{j,r,d}^k}-\sum_{r=1}^R{\eta _r\omega_{j,r,d}^k}|\ge \varepsilon \right)& \le \sum_{r=1}^R{P\left( |\bar{\eta}_r\bar{\omega}_{j,r,d}^k-\eta _r\omega _{j,r,d}^k|\ge \varepsilon /R \right)}\\
&\le \sum_{r=1}^R{P\left( |\bar{\eta}_r\bar{\omega}_{j,r,d}^k-\bar{\eta}_r\omega _{j,r,d}^k+\bar{\eta}_r\omega _{j,r,d}^k-\eta _r\omega _{j,r,d}^k|\ge \varepsilon /R \right)}\\
&\le \sum_{r=1}^R{P\left( \bar{\eta}_r|\bar{\omega}_{j,r,d}^k-\omega _{j,r,d}^k|\ge \varepsilon /2R \right)}+\sum_{r=1}^R{P\left( \omega _{j,r,d}^k|\bar{\eta}_r-\eta _r|\ge \varepsilon /2R \right)}\\
&\le \sum_{r=1}^R{P\left( |\bar{\omega}_{j,r,d}^k-\omega _{j,r,d}^k|\ge \varepsilon /2R \right)}+\sum_{r=1}^R{P\left( |\bar{\eta}_r-\eta _r|\ge \varepsilon /2R \right)}\\
&\le 12(2^d-1)R\exp \left( -2t_8 N\varepsilon ^2 \right) \\
\end{align*}

We have completed the proofs of Lemma \ref{lem:C_boundomega}.
\end{proof}

\begin{proof}[Proof of Theorem \ref{the:surescreening_general}, \ref{the:rank_general} and Proposition \ref{pro:general}]
	Following the lines of the proofs of Theorem \ref{pro:omega_bound} and \ref{the:surescreening} by setting $\varepsilon =  c_8 N^{-\kappa}$,
	\begin{equation*}
	P\left( \underset{1\le j\le p}{\max}|\bar{\omega}_j^{(d)}-\omega _j^{(d)}|\ge c_3 N^{-\kappa} \right) \le pP\left( |\bar{\omega}_j^{(d)}-\omega _j^{(d)}|\ge c_8 N^{-\kappa} \right)\le 12(2^d-1)pR\exp \left( -c_9 N^{1-2\kappa}\right)
	\end{equation*}
	
	If $\mathcal{A} \not \subset \hat{\mathcal{A}}$, then there must exist $k \in \mathcal{A}$ such that $\bar{\omega}_{j}<c N^{-\kappa}$.
	
	Furthermore, when $\max _{j \in \mathcal{A}}\left|\bar{\omega}_{j}^{(d)}-\omega_{j}^{(d)}\right| \leq c N^{-\kappa}$ and condition (C4) holds,
	\begin{equation*}
	\min _{j \in \mathcal{A}}\bar{\omega}_{j}^{(d)} \geq \min _{j \in \mathcal{A}}\left(\omega_{j}^{(d)}-\left|\bar{\omega}_{j}^{(d)}-\omega_{j}^{(d)}\right|\right) \geq \min _{j \in \mathcal{A}}\omega_{j}^{(d)}-\max _{j \in \mathcal{A}}\left|\bar{\omega}_{j}^{(d)}-\omega_{j}^{(d)}\right| \geq c N^{-\kappa}.
	\end{equation*}
	
	Therefore,
	\begin{equation}
	P\left(\mathcal{A} \subset \hat{\mathcal{A}}\right) \geq P\left(\max _{ j \in \mathcal{A}}\left|\bar{\omega}_{j}^{(d)}-\omega_{j}^{(d)}\right| \leq c N^{-\kappa}\right) \geq 1-12(2^d-1)sR\exp \left( -c_{10} N^{1-2\kappa}\right).
	\end{equation}
	for some constant $c_{10}>0$. 
	
	The proof of Theorem \ref{the:rank_general} is similar to the proof of Theorem \ref{the:rank}, we will skip it here. The proof of Proposition \ref{pro:general} can be directly derived from Proposition 1 in \citet{li_distributed_2020}. Hence, we also omit the detailed proof here. Then, we completed the proof of Theorem \ref{the:surescreening_general}, \ref{the:rank_general} and Proposition \ref{pro:general}.
\end{proof}

\section{Proofs of Proposition \ref{pro:special case}}\label{sec:conversion}
\subsection{Conditional Rank Utility}
Conditional Rank Utility (CRU) is constructed based on the ratio of the mean conditional rank to the mean unconditional rank of a feature
\begin{equation}
\omega_{j}=\sum_{r=1}^{R}\left(\mathbb{E}\left(F_{j}\left(X_{j}\right) I\left(Y=y_{r}\right)\right)-\frac{P\left(Y=y_{r}\right)}{2}\right)^{2}.
\end{equation}

First, we should focus on the decomposition of cumulative distribution function:
\begin{align*}
&F_{j}\left(X_{j}\right)=\mathbb{E}\left(I\left(x \leq X_{j}\right)\right)\\
&=\mathbb{E}\left(I\left(X \leq X_{j}\right) \mid Y=y_{r}\right) P\left(Y=y_{r}\right)+\mathbb{E}\left(I\left(X \leq X_{j}\right) \mid Y \neq y_{r}\right) P\left(Y \neq y_{r}\right) \\
& =F_{Y=y_{r}}\left(X_{j}\right) P\left(Y=y_{r}\right)+F_{Y \neq y_{r}}\left(X_{j}\right) P\left(Y \neq y_{r}\right).
\end{align*}

$\mathbb{E}\left(F_{j}\left(X_{j}\right) I\left(Y=y_{r}\right)\right)$ can be further expressed as 
\begin{align*}
&\mathbb{E}\left(F_{j}\left(X_{j}\right) I\left(Y=y_{r}\right)\right)=\mathbb{E}_{Y}\left(\mathbb{E}\left(F_{j}\left(X_{j}\right) I\left(Y=y_{r}\right)\right) \mid Y=y\right)\\
&=\mathbb{E}\left(F_{j}\left(X_{j}\right) I\left(Y=y_{r}\right) \mid Y=y_{r}\right) P\left(Y=y_{r}\right) \\
& =\mathbb{E}\left(F_{Y=y_{r}}\left(X_{j}\right) P\left(Y=y_{r}\right)+F_{Y \neq y_{r}}\left(X_{j}\right) P\left(Y \neq y_{r}\right) \mid Y=y_{r}\right) P\left(Y=y_{r}\right) \\
& =\mathbb{E}_{Y=y_{r}}\left(F_{Y=y_{r}}\left(X_{j}\right)\right) P\left(Y=y_{r}\right)^{2}+\mathbb{E}_{Y=y_{r}}\left(F_{Y \neq y_{r}}\left(X_{j}\right)\right) P\left(Y \neq y_{r}\right) P\left(Y=y_{r}\right) \\
& =\frac{1}{2} P\left(Y=y_{r}\right)^{2}+\mathbb{E}_{Y=y_{r}}\left(F_{Y \neq y_{r}}\left(X_{j}\right)\right) P\left(Y \neq y_{r}\right) P\left(Y=y_{r}\right).
\end{align*}

Based on the above two expressions, the CRU can be expressed as
\begin{equation}
\omega_{j}=\sum_{r=1}^{R}\left[P\left(Y=y_{r}\right)\left(1-P\left(Y=y_{r}\right)\right)\right]^{2} \omega_{j, r}^{2}.
\end{equation}

\subsection{Category-Adaptive Variable Screening}
Category-Adaptive Variable Screening is another model-free approach, defined by
\begin{equation}
\tau_{j, r}=\mathbb{E}\left(F\left(X_{j}\right) \mid Y_{j}=y_{r}\right)-\frac{1}{2}.
\end{equation}
As a special case, $\tau_{j}=\max _{r \in\{1, \cdots, R\}}\left|\tau_{j, r}\right|$ can be used to measure the dependence between $X_j$ and the categorical response $Y$.

Similar to the decomposition of the cumulative distribution function,
$\tau_{j, r}=\mathbb{E}\left(F\left(X_{j}\right) \mid Y_{j}=y_{r}\right)-\frac{1}{2}=\left(1-P\left(Y=y_{r}\right)\right) \omega_{j, r}$,
and the utility is 
\begin{equation*}
\tau_{p,r} =P\left(Y\ne y_{r}\right) \omega_{j, r} \text{ or }  \tau_{p}=\max _{r \in\{1, \cdots, R\}}\left[P\left(Y\ne y_{r}\right) \omega_{j, r}\right].
\end{equation*}

\subsection{Model-Free Feature Screening}
\citet{cui_model-free_2015} consider the marginal utility 
\begin{equation}
\mathbb{E}\left(\operatorname{Var}_{Y}(F(X \mid Y))\right)=\sum_{r=1}^{R} P\left(Y=y_{r}\right) \int\left[F_{j}\left(x \mid Y=y_{r}\right)-F_{j}(x)\right]^{2} d F_{j}(x).
\end{equation}
the utility can be expressed as $\sum_{r=1}^{R}\left(\frac{\theta_{j, r, 1}}{P\left(Y=y_{r}\right)}-2 \theta_{j, r, 2}+P\left(Y=y_{r}\right) \theta_{j, r, 3}\right)$

where 
\begin{align}\label{equ:MV}
&\theta_{j, r, 1}=\mathbb{E}_{X^{\prime}}\left[\mathbb{E}_{X_{j}, Y}\left(I\left(X_{j} \leq X_{j}^{\prime}, Y=y_{r}\right)\right)^{2}\right]\\
&\theta_{j, r, 2}=\mathbb{E}_{X_{j}^{\prime}}\left[\mathbb{E}_{X_{j}, Y}\left(I\left(X_{j} \leq X_{j}^{\prime}, Y=y_{r}\right)\right) \mathbb{E}_{X_{j}}\left(I\left(X_{j} \leq X_{j}^{\prime}\right)\right)\right]\\
&\theta_{j, r, 3}=\mathbb{E}_{X_{j}^{\prime}}\left[\mathbb{E}_{X_{j}}\left(I\left(X_{j} \leq X_{j}^{\prime}\right)\right)^{2}\right]=\mathbb{E}_{X_{j}^{\prime}}\left[F\left(X_{j}^{\prime}\right)^{2}\right].
\end{align}

We first focus on the transformation of $\theta_{j, r, 1}$. Continuing the same ideas as the CRU
\begin{align*}
\mathbb{E}_{X_{j}, Y}\left(I\left(X_{j} \leq X_{j}^{\prime}, Y=y_{r}\right)\right)&=\mathbb{E}_{X_{j}, Y}\left(I\left(X_{j} \leq X_{j}^{\prime}\right) \mid Y=y_{r}\right) P\left(Y=y_{r}\right)\\
&=F_{Y=y_{r}}\left(X_{j}^{\prime}\right) P\left(Y=y_{r}\right).
\end{align*}
Substituting the above results into the component
\begin{align*}
\theta_{j, r, 1}&=\mathbb{E}\left[F_{Y=y_{r}}\left(X_{j}^{\prime}\right)^{2} P\left(Y=y_{r}\right)^{2}\right]=\mathbb{E}_{Y}\left(\mathbb{E}\left[F_{Y=r}\left(X_{j}^{\prime}\right)^{2} P\left(Y=y_{r}\right)^{2}\right] \mid Y=y_j\right) \\
& =P\left(Y=y_{r}\right) \mathbb{E}_{Y=y_{r}}\left[F_{Y=y_{r}}\left(X_{j}^{\prime}\right)^{2} P\left(Y=y_{r}\right)^{2}\right]\\
&+P\left(Y \neq y_{r}\right) \mathbb{E}_{Y \neq y_{r}}\left[F_{Y=y_{r}}\left(X_{j}^{\prime}\right)^{2} P\left(Y=y_{r}\right)^{2}\right] \\
& =P\left(Y=y_{r}\right)^{2}\left\lbrace  P\left(Y=y_{r}\right)\mathbb{E}_{Y=y_{r}}\left[F_{Y=y_{r}}(X)^{2}\right]+P\left(Y \neq y_{r}\right)  \mathbb{E}_{Y \neq y_{r}}\left[F_{Y=y_{r}}(X)^{2}\right] \right\rbrace  .
\end{align*}

Next, we turn our attention to $\theta_{j, r, 2}$.

\begin{align*}
\mathbb{E}_{X_{j}}\left(I\left(X_{j} \leq X_{j}^{\prime}\right)\right)&=\mathbb{E}_{Y}\left(\mathbb{E}\left(I\left(X_{j} \leq X_{j}^{\prime}\right)\right) \mid Y=y_r\right)\\
&=P\left(Y=y_{r}\right) F_{Y=y_{r}}\left(X_{j}^{\prime}\right)+P\left(Y \neq y_{r}\right) F_{Y \neq y_{r}}\left(X_{j}^{\prime}\right). 
\end{align*}

Define $Q(X)=: \mathbb{E}_{X_{j}, Y}\left(I\left(X_{j} \leq X, Y=y_{r}\right)\right) \mathbb{E}_{X_{j}}\left(I\left(X_{j} \leq X\right)\right).$

Then $Q(x)$ and $\theta_{j, r, 2}$ can be expressed as
\begin{align*}
Q(x)=P\left(Y=y_{r}\right)^{2} F_{Y=y_{r}}(X)^{2}+P\left(Y=y_{r}\right) P\left(Y \neq y_{r}\right) F_{Y \neq y_{r}}(X) F_{Y=y_{r}}(X),
\end{align*}
\begin{align*}
\theta_{j, r, 2}&=\mathbb{E}_{X}[Q(x)]=P\left(Y=y_{r}\right) \mathbb{E}_{Y=y_{r}}[Q(X)]+P\left(Y \neq y_{r}\right) \mathbb{E}_{Y \neq y_{r}}[Q(X)] \\
& =P\left(Y=y_{r}\right)^{3} \mathbb{E}_{Y=y_{r}}\left[F_{Y=y_{r}}(X)^{2}\right]+P\left(Y=y_{r}\right)^{2} P\left(Y \neq y_{r}\right) \mathbb{E}_{Y \neq y_{r}}\left[F_{Y=y_{r}}(X)^{2}\right]\\
&+P\left(Y=y_{r}\right)^{2} P\left(Y \neq y_{r}\right) \mathbb{E}_{Y=y_{r}}\left[F_{Y \neq y_{r}}(X) F_{Y=y_{r}}(X)\right] \\
&+P\left(Y=y_{r}\right) P\left(Y \neq y_{r}\right)^{2} \mathbb{E}_{Y \neq y_{r}}\left[F_{Y \neq y_{r}}(X) F_{Y=y_{r}}(X)\right] \\
&=P(Y=r)^{3}\mathbb{E}_{Y=r}\left[F_{Y=r}(X)^{2}\right]+ \frac{P(Y=r)^{2} P(Y \neq r)}{2} \mathbb{E}_{Y \neq r}\left[F_{Y=r}(X)^{2}\right] \\
& +\frac{P(Y=r) P(Y \neq r)}{2}-\frac{P(Y=r) P(Y \neq r)^{2}}{2} \mathbb{E}_{Y=r}\left[F_{Y \neq r}(X)^{2}\right].
\end{align*}

Besides, notice that

\begin{align*}
\mathbb{E}_{Y \neq y_{r}}\left[F_{Y=y_{r}}(X)^{2}\right]&=\int F_{Y=y_{r}}(X)^{2} d F_{Y \neq y_{r}}(X)=1-2 \mathbb{E}_{Y=y_{r}}\left[F_{Y \neq y_{r}}(X) F_{Y=y_{r}}(X)\right],\\
\mathbb{E}_{Y=y_{r}}\left[F_{Y \neq y_{r}}(X)^{2}\right]&=1-2 \mathbb{E}_{Y \neq y_{r}}\left[F_{Y \neq y_{r}}(X) F_{Y=y_{r}}(X)\right].
\end{align*}
With the converted expressions of $\theta_{j, r, 1}$ and $\theta_{j, r, 2}$, for each category $r$,

\begin{align*}
& \frac{\theta_{j, r, 1}}{P\left(Y=y_{r}\right)}-2 \theta_{j, r, 2}+P\left(Y=y_{r}\right) \theta_{j, r, 3} \\
& =\left[P\left(Y=y_{r}\right)^{2}-2 P\left(Y=y_{r}\right)^{3}+P\left(Y=y_{r}\right)\right] \mathbb{E}_{Y=y_{r}}\left[F_{Y=y_{r}}(X)^{2}\right] \\
& +P\left(Y \neq y_{r}\right)^{2} P\left(Y=y_{r}\right)\left[\mathbb{E}_{Y \neq y_{r}}\left[F_{Y=y_{r}}(X)^{2}\right]+\mathbb{E}_{Y=y_{r}}\left[F_{Y \neq y_{r}}(X)^{2}\right]\right]\\
&-P\left(Y=y_{r}\right) P\left(Y \neq y_{r}\right) \\
& =P\left(Y \neq y_{r}\right)^{2} P\left(Y=y_{r}\right) \omega_{j, r,2}.
\end{align*}
So the MV-SIS utility can be transformed into the newly proposed form:
\begin{equation}
M V=\sum_{r=1}^{R} P\left(Y \neq y_{r}\right)^{2} P\left(Y=y_{r}\right){\omega_{j, r,2}}
\end{equation}

Based on the above decomposition, we can estimate $\omega_{j, r}$ and $\zeta_r$ separately to obtain similar label shift robust estimates.

\section{Additional results from the main text}
\begin{table}[!htbp]
	\centering
	\caption{Sample size for each institution.}
	\resizebox{\textwidth}{!}{
	\begin{tabular}{cc|cc}
		\toprule
		Institution & Number of sample & Institution & Number of sample \\
		\midrule
		Asterand & 58    & MSKCC & 44 \\
		Cureline & 38    & Mayo  & 62 \\
		Duke  & 52    & Roswell Park & 80 \\
		Greater Poland Cancer Center & 74    & University of Miami & 35 \\
		ILSBio & 48    & University of Pittsburgh & 137 \\
		Indivumed & 74    & Walter Reed & 92 \\
		International Genomics Consortium & 35    &       &  \\
		\bottomrule
	\end{tabular}}
	\label{tab:number of inst}
\end{table}

\subsection{Complete simulation results}
\begin{figure}[H]
	\centering
	\subfigure[$R=4$]
	{
		\begin{minipage}[b]{.21\linewidth}
			\centering
			\includegraphics[scale=0.145]{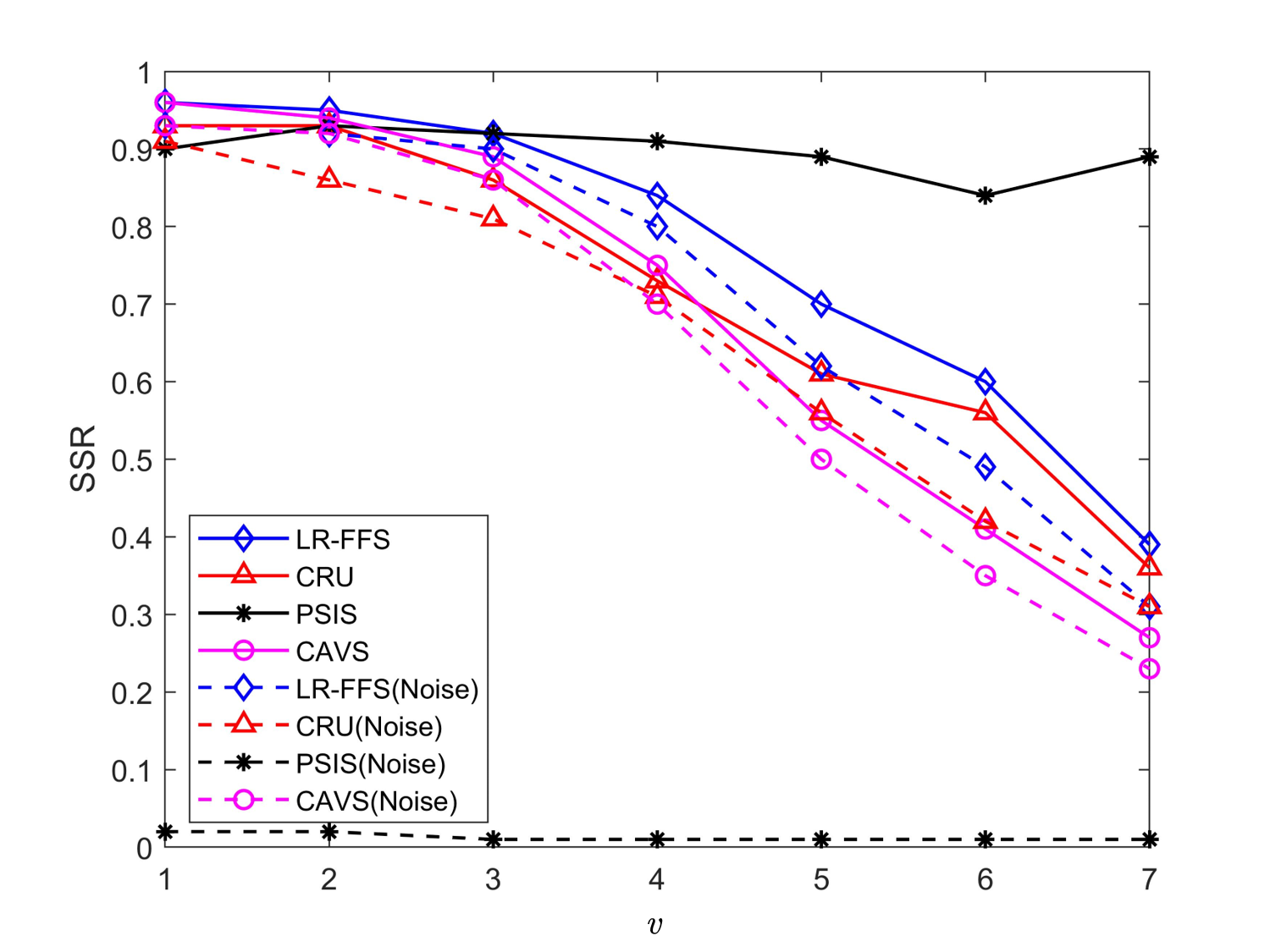} \\
			\includegraphics[scale=0.145]{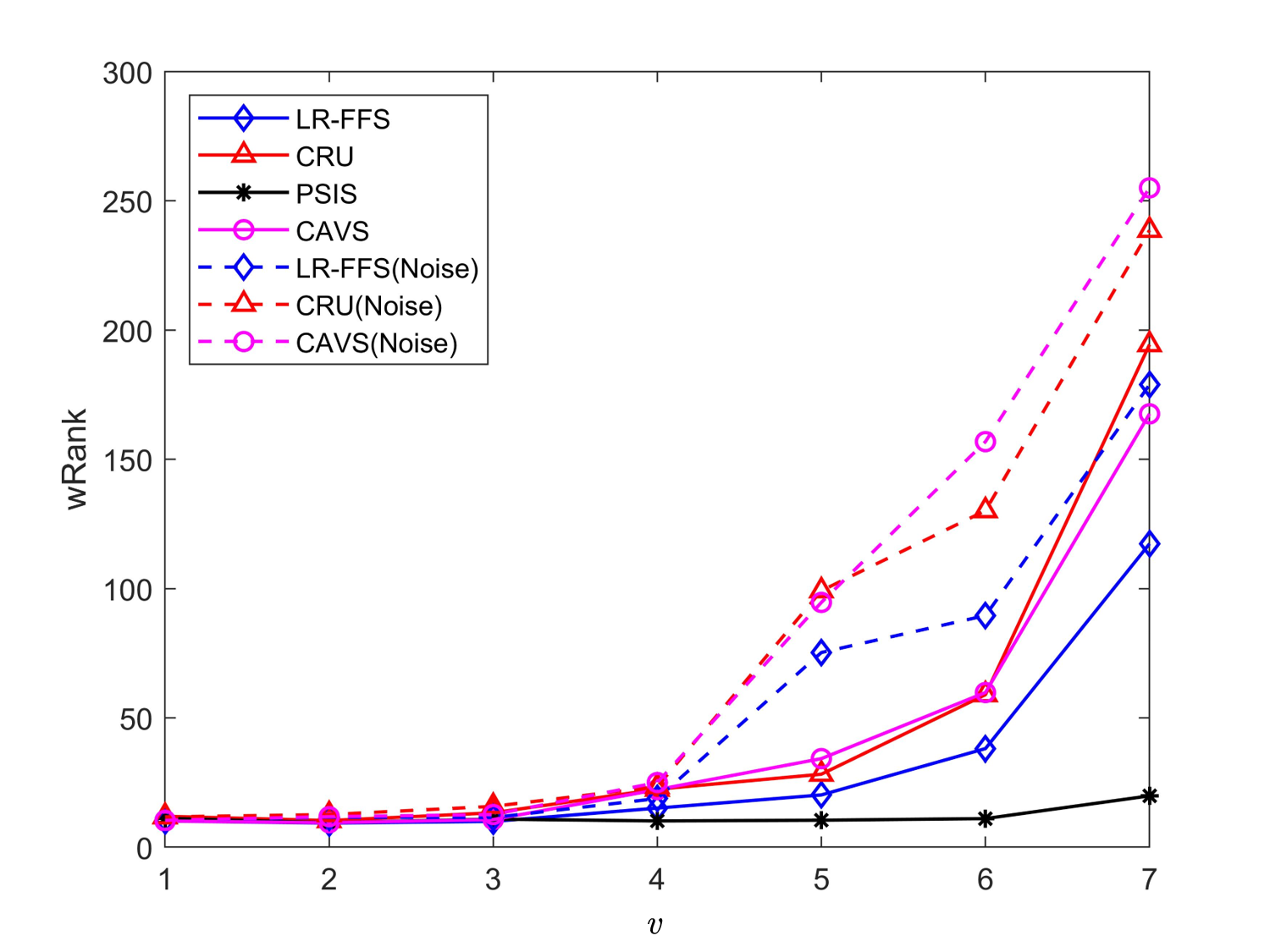}
		\end{minipage}
	}
	\subfigure[$R=5$]
	{
		\begin{minipage}[b]{.21\linewidth}
			\centering
			\includegraphics[scale=0.145]{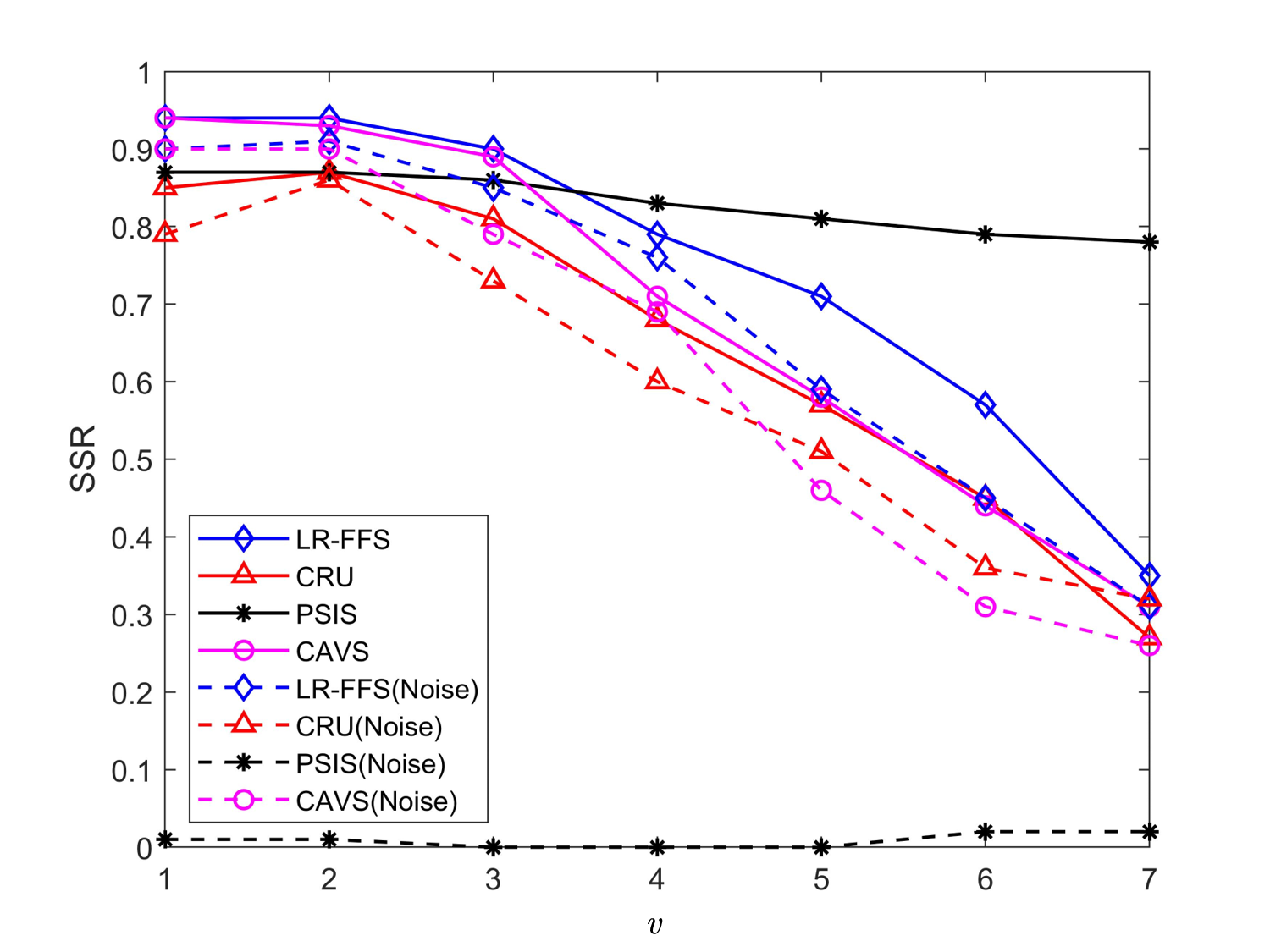} \\
			\includegraphics[scale=0.145]{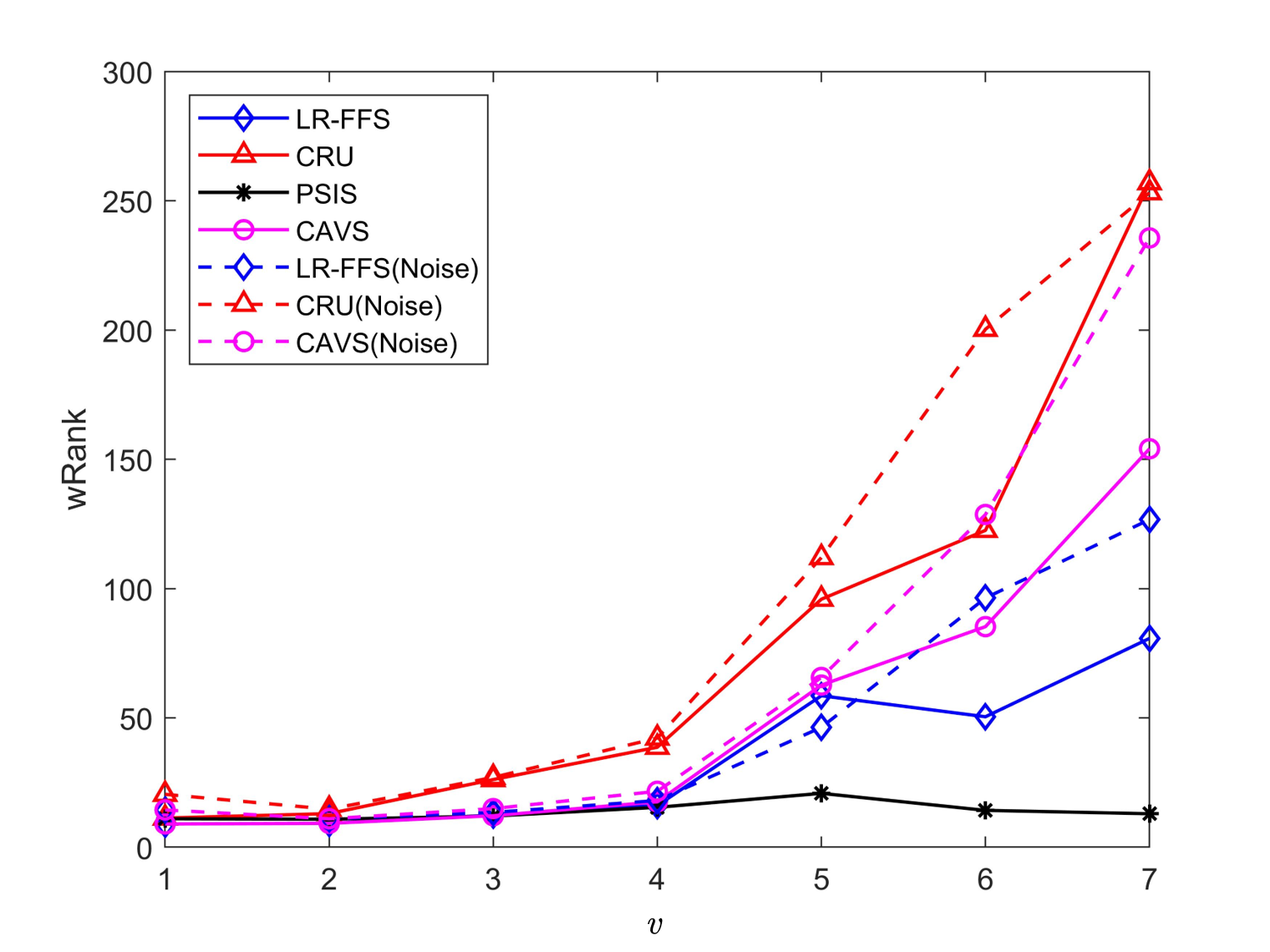}
		\end{minipage}
	}
	\subfigure[$R=6$]
	{
		\begin{minipage}[b]{.21\linewidth}
			\centering
			\includegraphics[scale=0.145]{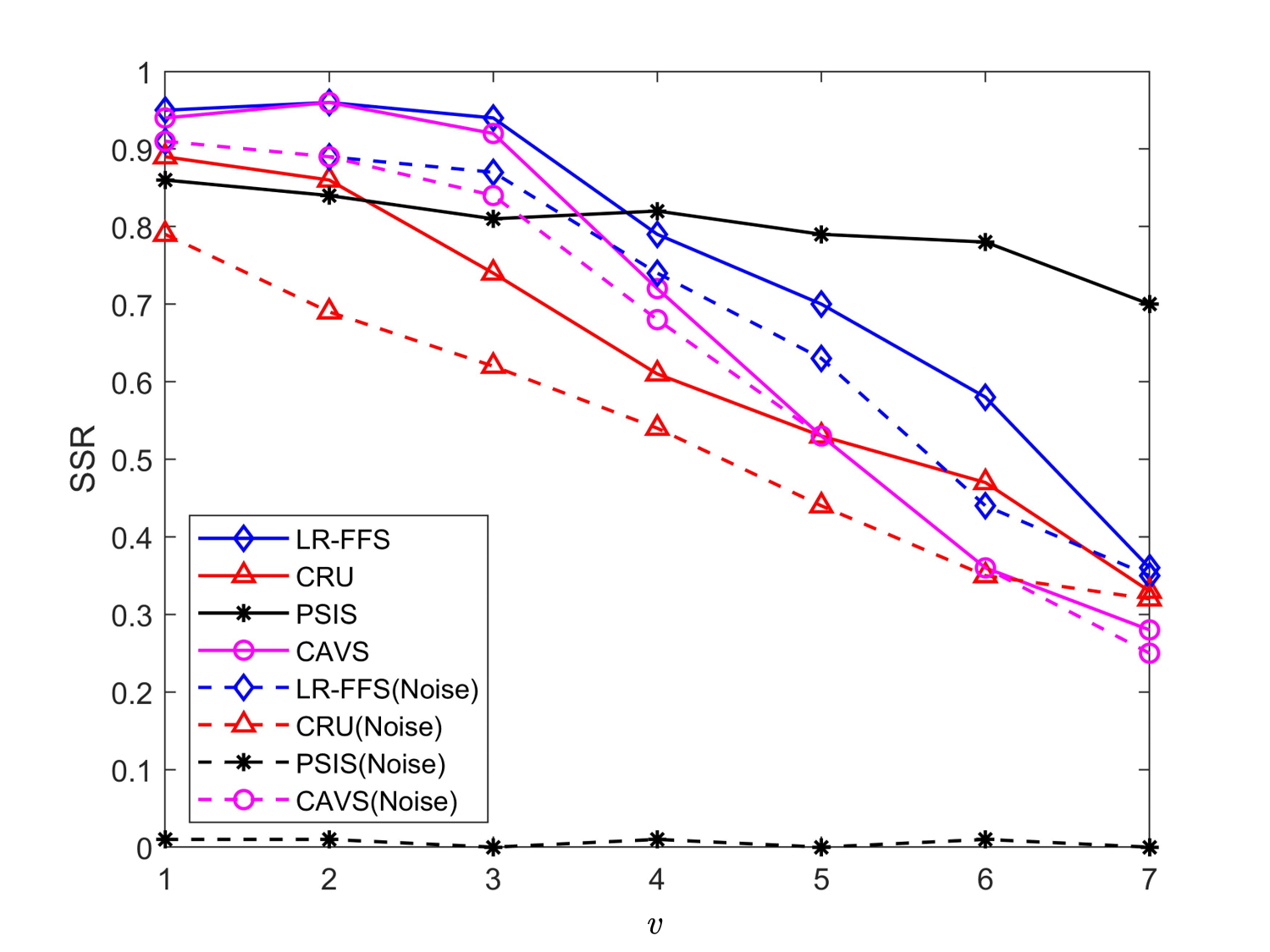} \\
			\includegraphics[scale=0.145]{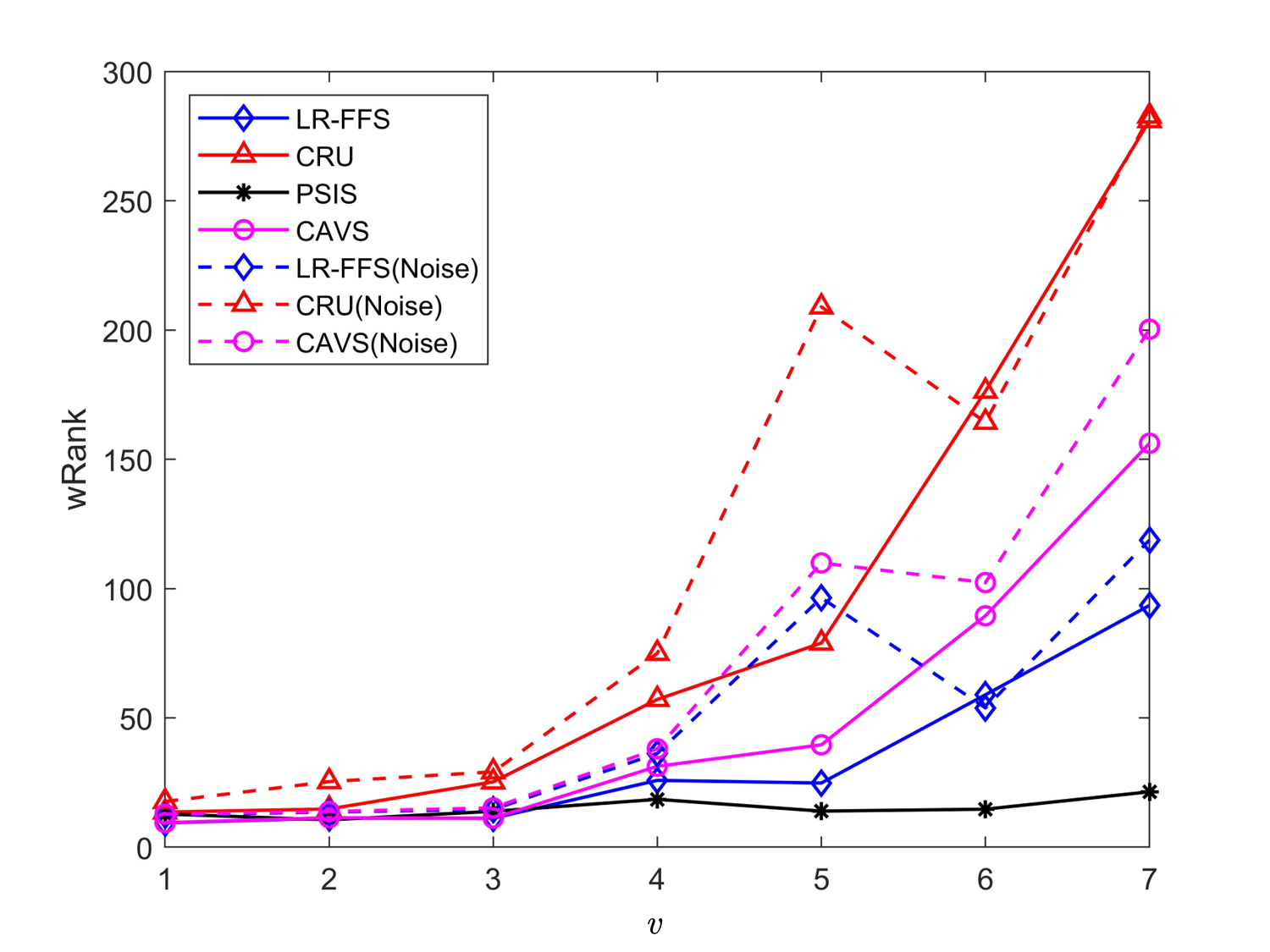}
		\end{minipage}
	}
	\subfigure[$R=7$]
	{
		\begin{minipage}[b]{.21\linewidth}
			\centering
			\includegraphics[scale=0.145]{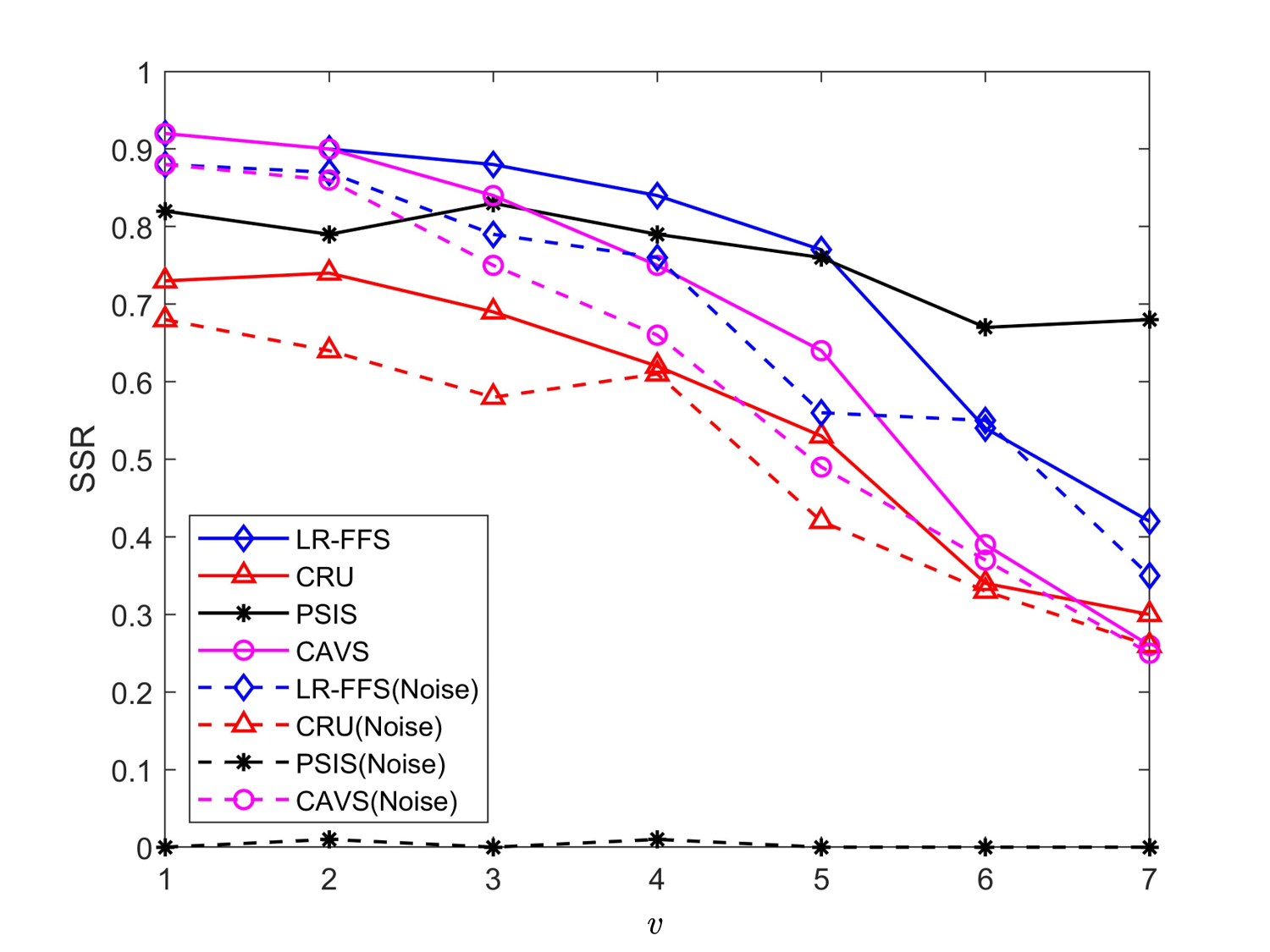} \\
			\includegraphics[scale=0.145]{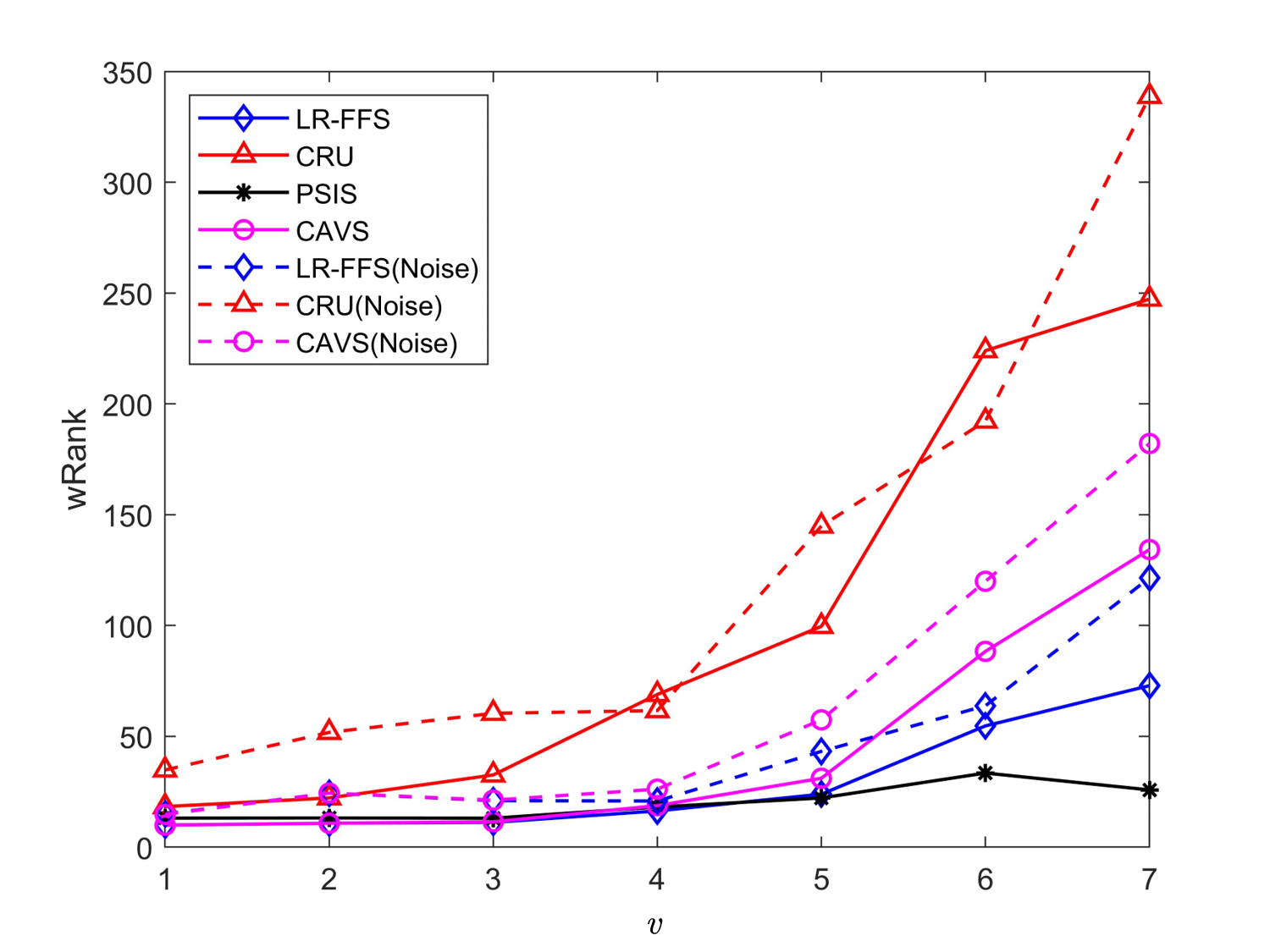}
		\end{minipage}
	}
	\caption{The simulation results for setting (a), where the first row represents SSR and the second row represents wRank.}\label{fig:settinga}
\end{figure}

Figure \ref{fig:settinga} depicts the experimental results for Example \ref{exa:example2}, setting (a), the simplest heterogeneity setting.

In the presence of noise, the wRank of the PSIS method exceeded 7000. To ensure the clarity of the visualization, the PSIS (Noise) is not shown here.

\begin{figure}[H]
	\centering
	\subfigure[Result of example \ref{exa:example4} setting(f)]
	{
		\begin{minipage}[b]{.45\linewidth}
			\centering
			\includegraphics[scale=0.045]{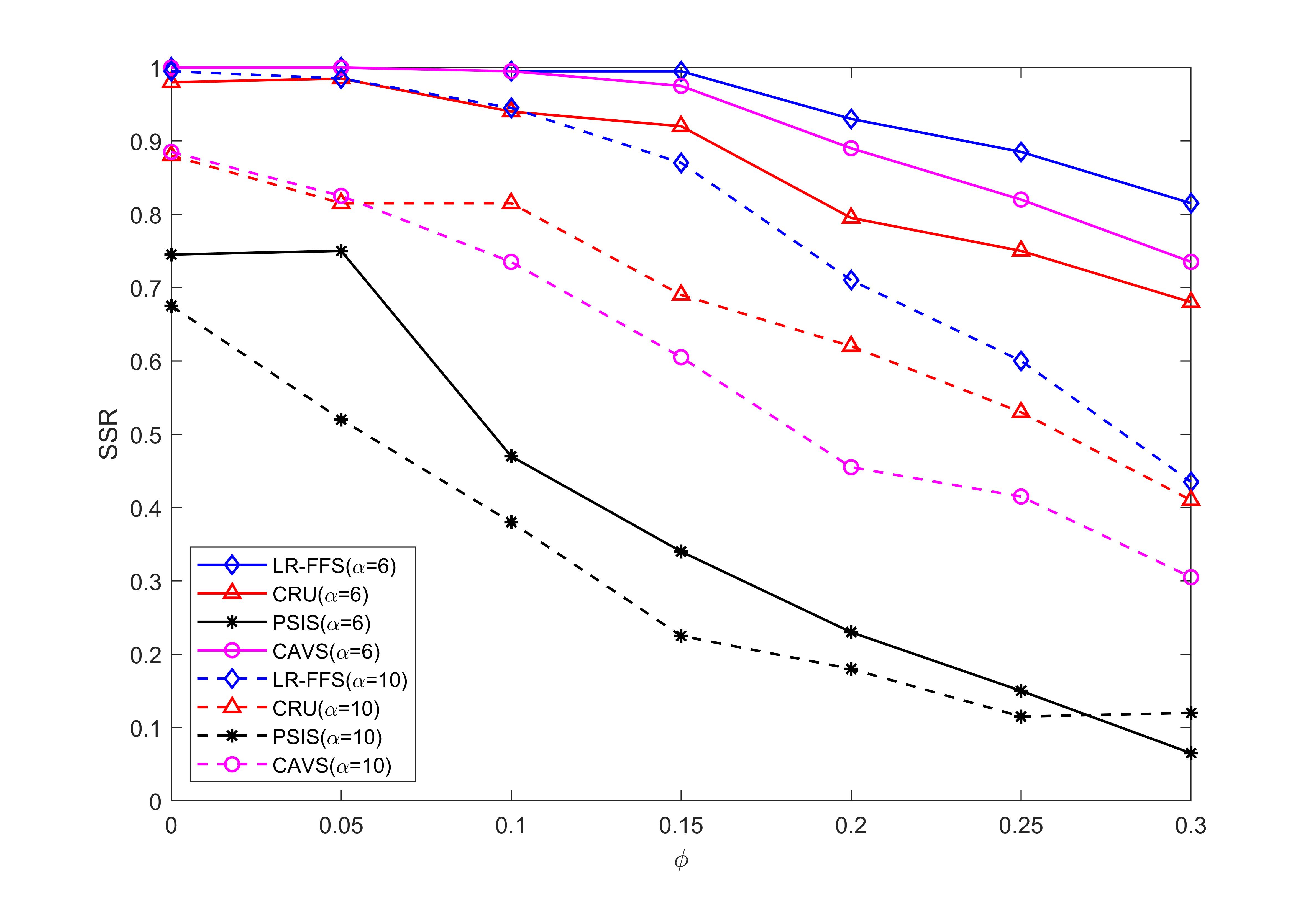} \\
			\includegraphics[scale=0.045]{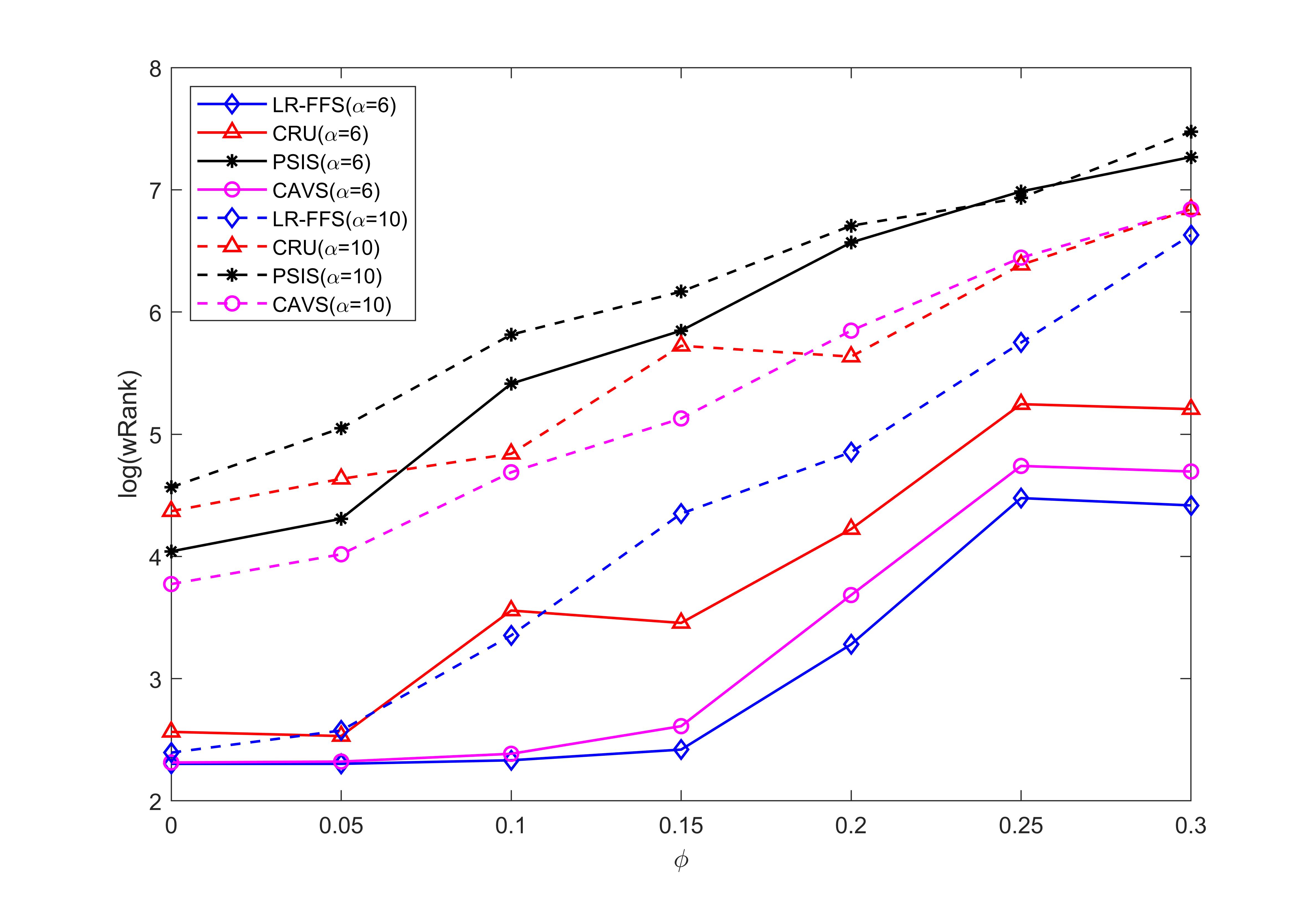}
		\end{minipage}
	}
	\subfigure[Result of example \ref{exa:example4} setting(g)]
	{
		\begin{minipage}[b]{.45\linewidth}
			\centering
			\includegraphics[scale=0.045]{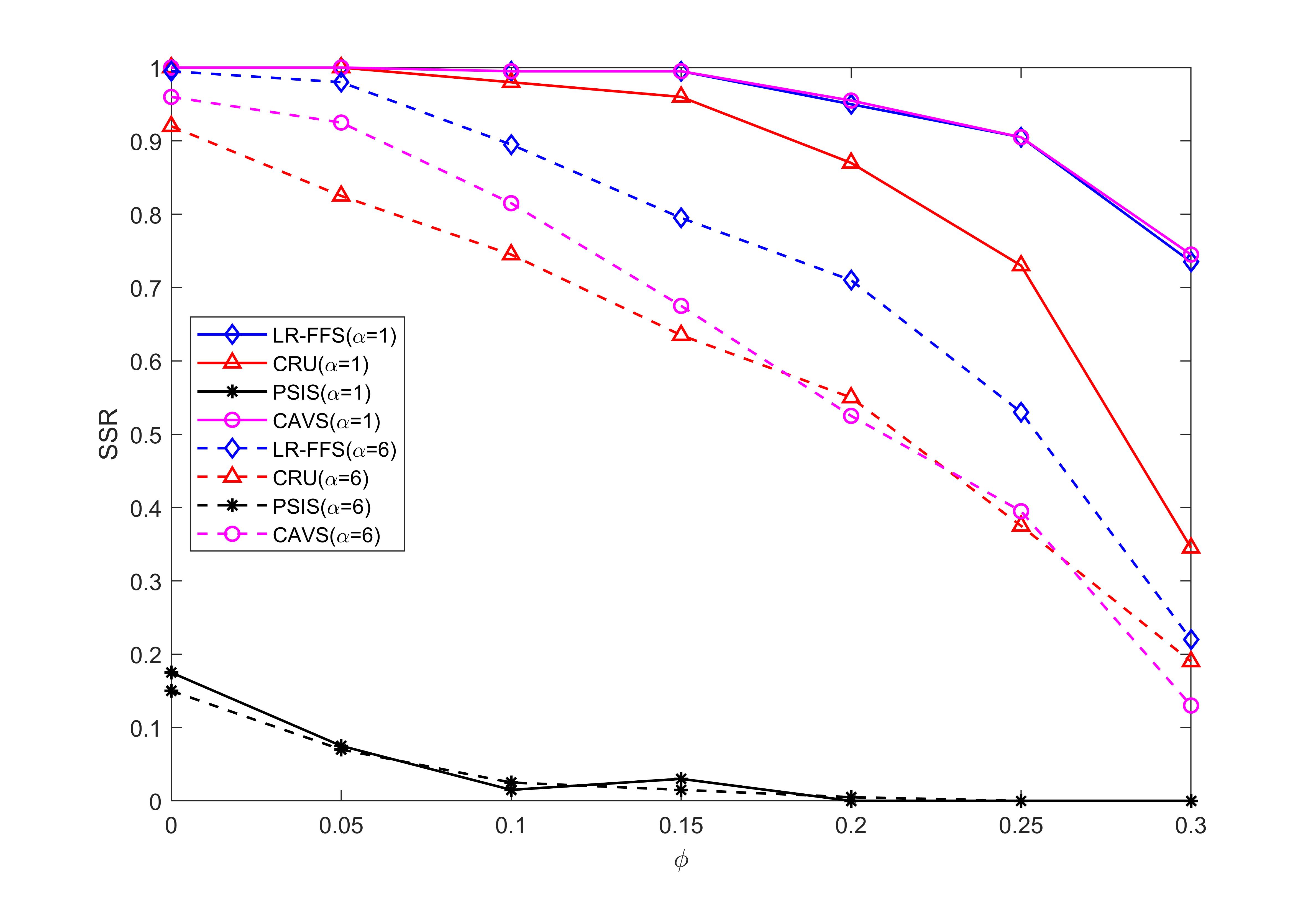} \\
			\includegraphics[scale=0.045]{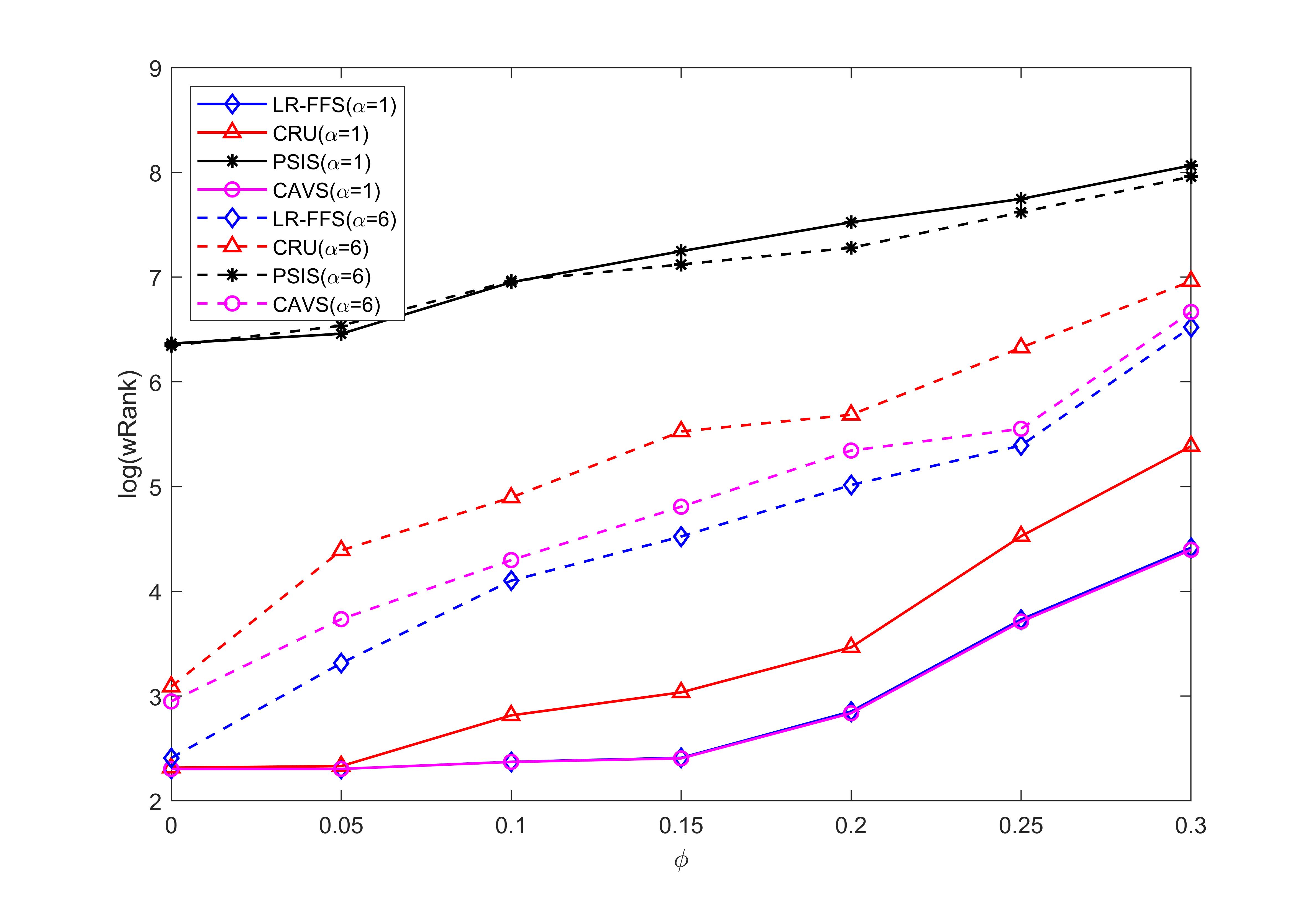}
		\end{minipage}
	}
	\caption{The simulation results for setting (f) and (g), where the first row represents SSR and the second row represents log(wRank).}\label{fig:settingfg}
\end{figure}

\begin{table}[!htbp]
  \centering
  \caption{Result for example \ref{exa:example2} Setting (d) without label shifting}
  \resizebox{\textwidth}{!}{
    \begin{tabular}{cccccccccccccrcccccccccccc}
    \toprule
    \multicolumn{13}{c}{The control of False Discovery Rate without noise}                                &       & \multicolumn{12}{c}{The control of False Discovery Rate with noise} \\
\cmidrule{1-13}\cmidrule{15-26}          & $\alpha$     & $X_1$    & $X_2$    & $X_3$   & $X_4$   & $X_5$   & $X_6$ & $X_7$  & $X_8$   & SSR   & FDR   & size  &       & $\alpha$     & $X_1$    & $X_2$    & $X_3$   & $X_4$   & $X_5$   & $X_6$ & $X_7$  & $X_8$    & SSR   & FDR   & size \\
\cmidrule{1-13}\cmidrule{15-26}    \multirow{7}[2]{*}{LR-FFS} & 0.1   & 1.00  & 1.00  & 1.00  & 1.00  & 1.00  & 1.00  & 1.00  & 1.00  & 1.00  & 0.08  & 8.85  &       & 0.1   & 1.00  & 0.99  & 1.00  & 1.00  & 1.00  & 1.00  & 1.00  & 1.00  & 0.98  & 0.10  & 9.12 \\
          & 0.15  & 1.00  & 0.99  & 1.00  & 1.00  & 0.99  & 1.00  & 1.00  & 1.00  & 0.98  & 0.12  & 9.40  &       & 0.15  & 0.99  & 1.00  & 1.00  & 1.00  & 0.99  & 1.00  & 1.00  & 1.00  & 0.98  & 0.12  & 9.33 \\
          & 0.2   & 1.00  & 1.00  & 1.00  & 1.00  & 1.00  & 1.00  & 1.00  & 1.00  & 1.00  & 0.16  & 10.15 &       & 0.2   & 1.00  & 0.99  & 1.00  & 1.00  & 1.00  & 0.99  & 1.00  & 0.99  & 0.98  & 0.18  & 10.71 \\
          & 0.25  & 1.00  & 1.00  & 1.00  & 1.00  & 1.00  & 1.00  & 1.00  & 1.00  & 1.00  & 0.22  & 11.18 &       & 0.25  & 1.00  & 1.00  & 1.00  & 1.00  & 1.00  & 1.00  & 1.00  & 1.00  & 0.99  & 0.23  & 11.15 \\
          & 0.3   & 1.00  & 1.00  & 1.00  & 1.00  & 1.00  & 1.00  & 1.00  & 1.00  & 1.00  & 0.28  & 12.12 &       & 0.3   & 1.00  & 1.00  & 1.00  & 1.00  & 1.00  & 1.00  & 1.00  & 1.00  & 1.00  & 0.28  & 13.09 \\
          & 0.35  & 1.00  & 1.00  & 1.00  & 1.00  & 1.00  & 1.00  & 1.00  & 1.00  & 1.00  & 0.32  & 13.53 &       & 0.35  & 1.00  & 1.00  & 1.00  & 1.00  & 1.00  & 1.00  & 1.00  & 1.00  & 1.00  & 0.34  & 15.50 \\
          & 0.4   & 1.00  & 1.00  & 1.00  & 1.00  & 1.00  & 1.00  & 1.00  & 1.00  & 1.00  & 0.39  & 15.77 &       & 0.4   & 1.00  & 1.00  & 1.00  & 1.00  & 1.00  & 1.00  & 1.00  & 1.00  & 0.99  & 0.40  & 18.37 \\
\cmidrule{1-13}\cmidrule{15-26}    \multirow{7}[2]{*}{CRU} & 0.1   & 1.00  & 1.00  & 1.00  & 1.00  & 1.00  & 1.00  & 1.00  & 1.00  & 1.00  & 0.10  & 9.11  &       & 0.1   & 1.00  & 0.99  & 1.00  & 1.00  & 1.00  & 1.00  & 1.00  & 1.00  & 0.99  & 0.10  & 9.21 \\
          & 0.15  & 1.00  & 1.00  & 1.00  & 1.00  & 1.00  & 1.00  & 1.00  & 1.00  & 0.99  & 0.11  & 9.28  &       & 0.15  & 1.00  & 1.00  & 1.00  & 1.00  & 1.00  & 1.00  & 1.00  & 1.00  & 1.00  & 0.12  & 9.67 \\
          & 0.2   & 1.00  & 1.00  & 1.00  & 1.00  & 1.00  & 1.00  & 1.00  & 1.00  & 1.00  & 0.16  & 10.12 &       & 0.2   & 1.00  & 1.00  & 1.00  & 1.00  & 1.00  & 1.00  & 1.00  & 1.00  & 1.00  & 0.17  & 10.84 \\
          & 0.25  & 1.00  & 1.00  & 1.00  & 1.00  & 1.00  & 1.00  & 1.00  & 1.00  & 1.00  & 0.23  & 11.26 &       & 0.25  & 1.00  & 1.00  & 1.00  & 1.00  & 1.00  & 1.00  & 1.00  & 1.00  & 0.99  & 0.23  & 11.62 \\
          & 0.3   & 1.00  & 1.00  & 1.00  & 1.00  & 1.00  & 1.00  & 1.00  & 1.00  & 1.00  & 0.28  & 12.39 &       & 0.3   & 1.00  & 1.00  & 1.00  & 1.00  & 1.00  & 1.00  & 1.00  & 1.00  & 1.00  & 0.31  & 14.50 \\
          & 0.35  & 1.00  & 1.00  & 1.00  & 1.00  & 1.00  & 1.00  & 1.00  & 1.00  & 1.00  & 0.33  & 13.92 &       & 0.35  & 1.00  & 1.00  & 1.00  & 1.00  & 1.00  & 1.00  & 1.00  & 1.00  & 1.00  & 0.32  & 15.15 \\
          & 0.4   & 1.00  & 1.00  & 1.00  & 1.00  & 1.00  & 1.00  & 1.00  & 1.00  & 1.00  & 0.41  & 16.09 &       & 0.4   & 1.00  & 1.00  & 1.00  & 1.00  & 1.00  & 1.00  & 1.00  & 1.00  & 1.00  & 0.40  & 19.62 \\
\cmidrule{1-13}\cmidrule{15-26}    \multirow{7}[2]{*}{PSIS} & 0.1   & 1.00  & 1.00  & 0.99  & 1.00  & 0.96  & 0.99  & 1.00  & 1.00  & 0.92  & 0.10  & 9.02  &       & 0.1   & 0.33  & 0.34  & 0.34  & 0.36  & 0.32  & 0.37  & 0.31  & 0.32  & 0.14  & 0.67  & 2809.24 \\
          & 0.15  & 0.99  & 0.98  & 0.98  & 0.99  & 0.99  & 0.99  & 0.99  & 0.98  & 0.88  & 0.11  & 9.16  &       & 0.15  & 0.32  & 0.35  & 0.31  & 0.31  & 0.34  & 0.34  & 0.29  & 0.30  & 0.12  & 0.69  & 2721.94 \\
          & 0.2   & 1.00  & 0.99  & 1.00  & 1.00  & 1.00  & 0.99  & 1.00  & 0.99  & 0.96  & 0.18  & 10.35 &       & 0.2   & 0.30  & 0.30  & 0.30  & 0.29  & 0.30  & 0.28  & 0.33  & 0.31  & 0.12  & 0.67  & 2725.71 \\
          & 0.25  & 0.99  & 1.00  & 1.00  & 0.99  & 1.00  & 1.00  & 1.00  & 1.00  & 0.97  & 0.23  & 11.14 &       & 0.25  & 0.32  & 0.33  & 0.32  & 0.35  & 0.34  & 0.35  & 0.33  & 0.31  & 0.12  & 0.64  & 2777.29 \\
          & 0.3   & 1.00  & 1.00  & 1.00  & 1.00  & 0.99  & 1.00  & 0.99  & 1.00  & 0.97  & 0.28  & 12.21 &       & 0.3   & 0.34  & 0.36  & 0.33  & 0.35  & 0.38  & 0.33  & 0.33  & 0.35  & 0.13  & 0.65  & 3026.97 \\
          & 0.35  & 1.00  & 1.00  & 1.00  & 1.00  & 0.99  & 0.99  & 0.99  & 1.00  & 0.96  & 0.32  & 13.11 &       & 0.35  & 0.41  & 0.39  & 0.39  & 0.37  & 0.37  & 0.40  & 0.41  & 0.41  & 0.21  & 0.66  & 3331.68 \\
          & 0.4   & 0.99  & 1.00  & 1.00  & 1.00  & 0.99  & 1.00  & 1.00  & 1.00  & 0.97  & 0.39  & 15.38 &       & 0.4   & 0.38  & 0.37  & 0.37  & 0.34  & 0.35  & 0.37  & 0.38  & 0.38  & 0.18  & 0.62  & 3384.30 \\
\cmidrule{1-13}\cmidrule{15-26}    \multirow{7}[1]{*}{CAVS} & 0.1   & 1.00  & 1.00  & 1.00  & 1.00  & 1.00  & 1.00  & 1.00  & 1.00  & 1.00  & 0.08  & 8.85  &       & 0.1   & 1.00  & 0.99  & 1.00  & 1.00  & 1.00  & 1.00  & 1.00  & 1.00  & 0.98  & 0.10  & 9.12 \\
          & 0.15  & 1.00  & 0.99  & 1.00  & 1.00  & 0.99  & 1.00  & 1.00  & 1.00  & 0.98  & 0.12  & 9.40  &       & 0.15  & 0.99  & 1.00  & 1.00  & 1.00  & 0.99  & 1.00  & 1.00  & 1.00  & 0.98  & 0.12  & 9.33 \\
          & 0.2   & 1.00  & 1.00  & 1.00  & 1.00  & 1.00  & 1.00  & 1.00  & 1.00  & 1.00  & 0.16  & 10.15 &       & 0.2   & 1.00  & 0.99  & 1.00  & 1.00  & 1.00  & 0.99  & 1.00  & 0.99  & 0.98  & 0.18  & 10.71 \\
          & 0.25  & 1.00  & 1.00  & 1.00  & 1.00  & 1.00  & 1.00  & 1.00  & 1.00  & 1.00  & 0.23  & 11.19 &       & 0.25  & 1.00  & 1.00  & 1.00  & 1.00  & 1.00  & 1.00  & 1.00  & 1.00  & 0.99  & 0.22  & 11.15 \\
          & 0.3   & 1.00  & 1.00  & 1.00  & 1.00  & 1.00  & 1.00  & 1.00  & 1.00  & 1.00  & 0.28  & 12.13 &       & 0.3   & 1.00  & 1.00  & 1.00  & 1.00  & 1.00  & 1.00  & 1.00  & 1.00  & 1.00  & 0.28  & 13.10 \\
          & 0.35  & 1.00  & 1.00  & 1.00  & 1.00  & 1.00  & 1.00  & 1.00  & 1.00  & 1.00  & 0.32  & 13.53 &       & 0.35  & 1.00  & 1.00  & 1.00  & 1.00  & 1.00  & 1.00  & 1.00  & 1.00  & 1.00  & 0.34  & 15.51 \\
          & 0.4   & 1.00  & 1.00  & 1.00  & 1.00  & 1.00  & 1.00  & 1.00  & 1.00  & 1.00  & 0.39  & 15.89 &       & 0.4   & 1.00  & 1.00  & 1.00  & 1.00  & 1.00  & 1.00  & 1.00  & 1.00  & 0.99  & 0.40  & 18.38 \\
          \bottomrule
    \end{tabular}}
  \label{result:FDR1}
\end{table}

\begin{table}[!htbp]
  \centering
  \caption{Result for Example \ref{exa:example2} Setting(d) with label shifting}
  \resizebox{\textwidth}{!}{
    \begin{tabular}{cccccccccccccrcccccccccccc}
    \toprule
    \multicolumn{13}{c}{The control of False Discovery Rate without noise}                                &       & \multicolumn{12}{c}{The control of False Discovery Rate with noise} \\
\cmidrule{1-13}\cmidrule{15-26}          & $\alpha$     & $X_1$    & $X_2$    & $X_3$   & $X_4$   & $X_5$   & $X_6$ & $X_7$  & $X_8$   & SSR   & FDR   & size  &       & $\alpha$     & $X_1$    & $X_2$    & $X_3$   & $X_4$   & $X_5$   & $X_6$ & $X_7$  & $X_8$    & SSR   & FDR   & size \\
\cmidrule{1-13}\cmidrule{15-26}    \multirow{7}[2]{*}{LR-FFS} & 0.10  & 0.99  & 0.98  & 0.98  & 0.97  & 0.99  & 0.99  & 1.00  & 0.99  & 0.89  & 0.10  & 8.90  &       & 0.10  & 0.96  & 0.98  & 0.96  & 0.98  & 0.98  & 0.97  & 0.98  & 0.96  & 0.78  & 0.13  & 9.17 \\
          & 0.15  & 0.98  & 0.99  & 0.99  & 0.98  & 0.99  & 0.97  & 0.99  & 1.00  & 0.87  & 0.11  & 9.12  &       & 0.15  & 0.97  & 0.98  & 0.96  & 0.98  & 0.97  & 0.98  & 0.97  & 0.99  & 0.82  & 0.12  & 9.39 \\
          & 0.20  & 0.99  & 0.99  & 0.98  & 0.98  & 1.00  & 0.99  & 0.99  & 1.00  & 0.91  & 0.18  & 10.29 &       & 0.20  & 0.98  & 0.98  & 0.99  & 0.98  & 0.98  & 0.98  & 0.96  & 0.99  & 0.83  & 0.17  & 10.26 \\
          & 0.25  & 0.99  & 0.99  & 0.99  & 0.99  & 1.00  & 0.99  & 0.99  & 1.00  & 0.93  & 0.23  & 11.32 &       & 0.25  & 0.99  & 0.99  & 0.99  & 0.99  & 1.00  & 0.99  & 0.99  & 0.99  & 0.92  & 0.25  & 14.39 \\
          & 0.30  & 1.00  & 1.00  & 1.00  & 0.98  & 1.00  & 1.00  & 0.99  & 1.00  & 0.94  & 0.28  & 12.27 &       & 0.30  & 0.98  & 0.98  & 0.98  & 1.00  & 0.99  & 1.00  & 0.99  & 0.99  & 0.89  & 0.28  & 12.38 \\
          & 0.35  & 1.00  & 1.00  & 1.00  & 1.00  & 1.00  & 1.00  & 0.99  & 0.99  & 0.96  & 0.31  & 12.93 &       & 0.35  & 0.99  & 0.99  & 1.00  & 0.99  & 0.99  & 1.00  & 0.99  & 1.00  & 0.96  & 0.34  & 15.04 \\
          & 0.40  & 1.00  & 0.99  & 1.00  & 1.00  & 1.00  & 0.99  & 1.00  & 0.99  & 0.96  & 0.33  & 13.32 &       & 0.40  & 1.00  & 0.99  & 1.00  & 1.00  & 0.98  & 0.99  & 0.99  & 1.00  & 0.95  & 0.39  & 17.29 \\
\cmidrule{1-13}\cmidrule{15-26}    \multirow{7}[2]{*}{CRU} & 0.10  & 0.99  & 0.99  & 0.98  & 0.98  & 0.98  & 0.98  & 0.99  & 0.99  & 0.89  & 0.09  & 8.92  &       & 0.10  & 0.96  & 0.98  & 0.95  & 0.97  & 0.97  & 0.98  & 0.97  & 0.93  & 0.82  & 0.11  & 8.97 \\
          & 0.15  & 0.98  & 1.00  & 0.99  & 0.98  & 0.97  & 0.96  & 0.98  & 0.99  & 0.87  & 0.11  & 9.03  &       & 0.15  & 0.98  & 0.99  & 0.98  & 0.98  & 0.98  & 0.98  & 0.98  & 0.98  & 0.88  & 0.10  & 9.08 \\
          & 0.20  & 0.99  & 0.99  & 0.99  & 0.99  & 0.99  & 0.98  & 0.99  & 0.99  & 0.91  & 0.18  & 10.30 &       & 0.20  & 0.97  & 0.99  & 0.98  & 0.97  & 0.98  & 0.98  & 0.97  & 0.99  & 0.85  & 0.18  & 10.33 \\
          & 0.25  & 0.99  & 1.00  & 0.99  & 0.99  & 0.99  & 0.99  & 0.99  & 0.99  & 0.93  & 0.23  & 10.97 &       & 0.25  & 0.99  & 0.99  & 0.97  & 1.00  & 0.97  & 0.98  & 0.98  & 0.96  & 0.89  & 0.27  & 12.83 \\
          & 0.30  & 1.00  & 0.99  & 0.99  & 0.99  & 1.00  & 1.00  & 0.98  & 1.00  & 0.92  & 0.29  & 12.55 &       & 0.30  & 0.98  & 0.97  & 0.97  & 0.98  & 0.98  & 1.00  & 0.99  & 0.98  & 0.91  & 0.31  & 13.75 \\
          & 0.35  & 1.00  & 1.00  & 1.00  & 0.99  & 0.99  & 0.99  & 0.99  & 1.00  & 0.95  & 0.33  & 13.49 &       & 0.35  & 0.98  & 1.00  & 1.00  & 0.99  & 0.99  & 1.00  & 1.00  & 1.00  & 0.96  & 0.36  & 15.42 \\
          & 0.40  & 1.00  & 0.99  & 0.99  & 1.00  & 0.99  & 0.98  & 1.00  & 0.99  & 0.94  & 0.36  & 14.65 &       & 0.40  & 0.99  & 0.99  & 1.00  & 0.99  & 0.99  & 0.99  & 0.98  & 0.99  & 0.92  & 0.40  & 17.46 \\
\cmidrule{1-13}\cmidrule{15-26}    \multirow{7}[2]{*}{PSIS} & 0.10  & 0.96  & 0.95  & 0.94  & 0.93  & 0.97  & 0.95  & 0.97  & 0.95  & 0.71  & 0.10  & 8.67  &       & 0.10  & 0.32  & 0.36  & 0.31  & 0.28  & 0.29  & 0.28  & 0.33  & 0.31  & 0.11  & 0.71  & 2585.78 \\
          & 0.15  & 0.98  & 0.96  & 0.97  & 0.96  & 0.99  & 0.93  & 0.95  & 0.97  & 0.75  & 0.12  & 9.15  &       & 0.15  & 0.29  & 0.31  & 0.30  & 0.27  & 0.28  & 0.28  & 0.30  & 0.28  & 0.10  & 0.62  & 2534.87 \\
          & 0.20  & 0.97  & 0.95  & 0.95  & 0.95  & 0.98  & 0.96  & 0.97  & 0.96  & 0.77  & 0.17  & 9.92  &       & 0.20  & 0.33  & 0.34  & 0.31  & 0.33  & 0.33  & 0.34  & 0.33  & 0.33  & 0.14  & 0.67  & 2850.35 \\
          & 0.25  & 0.99  & 0.98  & 0.98  & 0.98  & 0.99  & 0.95  & 0.98  & 0.98  & 0.85  & 0.24  & 11.27 &       & 0.25  & 0.31  & 0.35  & 0.32  & 0.34  & 0.32  & 0.33  & 0.34  & 0.32  & 0.12  & 0.63  & 3037.08 \\
          & 0.30  & 0.98  & 0.98  & 0.94  & 0.98  & 0.99  & 0.97  & 0.97  & 1.00  & 0.85  & 0.29  & 12.20 &       & 0.30  & 0.38  & 0.33  & 0.37  & 0.31  & 0.37  & 0.31  & 0.32  & 0.33  & 0.16  & 0.64  & 2943.98 \\
          & 0.35  & 0.99  & 0.99  & 0.98  & 0.98  & 0.99  & 0.98  & 1.00  & 0.99  & 0.88  & 0.34  & 13.63 &       & 0.35  & 0.36  & 0.39  & 0.37  & 0.33  & 0.33  & 0.36  & 0.35  & 0.34  & 0.13  & 0.69  & 3180.96 \\
          & 0.40  & 1.00  & 0.98  & 1.00  & 0.98  & 0.99  & 0.98  & 0.98  & 0.98  & 0.88  & 0.40  & 15.63 &       & 0.40  & 0.30  & 0.36  & 0.32  & 0.34  & 0.36  & 0.31  & 0.32  & 0.34  & 0.16  & 0.58  & 2902.57 \\
\cmidrule{1-13}\cmidrule{15-26}    \multirow{7}[1]{*}{CAVS} & 0.10  & 0.95  & 0.95  & 0.96  & 0.96  & 0.97  & 0.96  & 0.98  & 0.96  & 0.77  & 0.10  & 8.72  &       & 0.10  & 0.97  & 0.95  & 0.92  & 0.95  & 0.93  & 0.95  & 0.95  & 0.94  & 0.67  & 0.13  & 8.95 \\
          & 0.15  & 0.96  & 0.96  & 0.98  & 0.97  & 0.98  & 0.95  & 0.94  & 0.97  & 0.76  & 0.11  & 9.01  &       & 0.15  & 0.95  & 0.94  & 0.94  & 0.97  & 0.95  & 0.97  & 0.95  & 0.97  & 0.74  & 0.13  & 9.30 \\
          & 0.20  & 0.96  & 0.96  & 0.98  & 0.97  & 0.98  & 0.96  & 0.98  & 0.97  & 0.82  & 0.17  & 9.86  &       & 0.20  & 0.97  & 0.95  & 0.95  & 0.95  & 0.97  & 0.94  & 0.92  & 0.94  & 0.69  & 0.16  & 9.89 \\
          & 0.25  & 0.98  & 0.97  & 0.98  & 0.98  & 0.99  & 0.97  & 0.97  & 0.99  & 0.86  & 0.23  & 11.09 &       & 0.25  & 0.96  & 0.96  & 0.97  & 0.98  & 0.98  & 0.97  & 0.98  & 0.97  & 0.84  & 0.25  & 15.09 \\
          & 0.30  & 1.00  & 0.98  & 0.98  & 0.96  & 0.97  & 0.97  & 0.98  & 0.98  & 0.87  & 0.28  & 11.93 &       & 0.30  & 0.96  & 0.97  & 0.97  & 0.96  & 0.98  & 0.97  & 0.98  & 0.96  & 0.79  & 0.28  & 12.97 \\
          & 0.35  & 0.98  & 0.97  & 0.97  & 0.98  & 0.98  & 0.95  & 0.98  & 0.97  & 0.82  & 0.31  & 13.12 &       & 0.35  & 0.98  & 0.96  & 0.97  & 0.99  & 0.96  & 0.96  & 0.97  & 0.97  & 0.83  & 0.33  & 14.52 \\
          & 0.40  & 0.97  & 0.98  & 0.98  & 0.99  & 0.98  & 0.97  & 0.99  & 0.98  & 0.88  & 0.34  & 13.50 &       & 0.40  & 0.96  & 0.95  & 0.98  & 0.98  & 0.98  & 0.97  & 0.96  & 0.96  & 0.84  & 0.38  & 18.39 \\
          \bottomrule
    \end{tabular}}
  \label{result:FDR2}
\end{table}%

\begin{example}\label{exa:example extend}
    In this example, we simulate based on setting (a) in example \ref{exa:example2} where $R=6$, considering simulation results for different weight selections. Specifically, for the CRU weight, $\zeta_r = [\pi_r(1-\pi_r)]^2$, for CAVS weight, $\zeta_r=1-\pi_r$, for MV-SIS weight, $\zeta_r = \pi_r^2(1-\pi_r)$, for equal weight, $\zeta_r=1$. The simulation results are shown in Table \ref{tab:table_extend}, with results in parentheses indicating outcomes under noise scenarios.
\end{example}

\begin{table}[!htbp]
	\centering
	\caption{Result for Example \ref{exa:example extend}.}
	\resizebox{\textwidth}{!}{
	\begin{tabular}{cclllllll}
		\toprule
		&       & \multicolumn{1}{c}{1} & \multicolumn{1}{c}{2} & \multicolumn{1}{c}{3} & \multicolumn{1}{c}{4} & \multicolumn{1}{c}{5} & \multicolumn{1}{c}{6} & \multicolumn{1}{c}{7} \\
		\midrule
		\multirow{6}[2]{*}{SSR$\uparrow$} & LR-FFS & \textbf{0.86(0.88)} & \textbf{0.92(0.96)} & \textbf{0.88(0.82)} & \textbf{0.74(0.78)} & \textbf{0.74(0.62)} & \textbf{0.54(0.58)} & \textbf{0.48(0.41)} \\
		& LR-FFS(CRU weight) & 0.76(0.76) & 0.82(0.8) & 0.6(0.64) & 0.52(0.48) & 0.6(0.32) & 0.28(0.42) & 0.32(0.38) \\
		& LR-FFS-PAIR & 0.7(0.64) & 0.8(0.74) & 0.46(0.46) & 0.32(0.22) & 0.2(0.08) & 0.04(0) & 0(0) \\
		& LR-FFS(CAVS weight) & 0.26(0.22) & 0.38(0.28) & 0.18(0.22) & 0.2(0.08) & 0.16(0.1) & 0.08(0.02) & 0.1(0.02) \\
		& LR-FFS(MV-SIS weight) & 0.76(0.74) & 0.82(0.8) & 0.74(0.7) & 0.62(0.58) & 0.64(0.32) & 0.36(0.36) & 0.32(0.36) \\
		& LR-FFS(equal weight) & 0.26(0.22) & 0.38(0.28) & 0.18(0.26) & 0.22(0.1) & 0.22(0.1) & 0.1(0.04) & 0.1(0.02) \\
		\midrule
		\multirow{6}[2]{*}{PSR$\uparrow$} & LR-FFS & \textbf{0.98(0.99)} & \textbf{0.99(0.99)} & \textbf{0.99(0.98)} & \textbf{0.96(0.97)} & \textbf{0.95(0.94)} & \textbf{0.9(0.9)} & \textbf{0.87(0.84)} \\
		& LR-FFS(CRU weight) & 0.96(0.97) & 0.97(0.97) & 0.93(0.94) & 0.89(0.9) & 0.89(0.79) & 0.73(0.81) & 0.72(0.76) \\
		& LR-FFS-PAIR & 0.95(0.94) & 0.97(0.96) & 0.89(0.89) & 0.82(0.78) & 0.73(0.67) & 0.52(0.48) & 0.29(0.32) \\
		& LR-FFS(CAVS weight) & 0.83(0.8) & 0.85(0.83) & 0.76(0.79) & 0.73(0.71) & 0.75(0.64) & 0.59(0.54) & 0.53(0.53) \\
		& LR-FFS(MV-SIS weight) & 0.96(0.96) & 0.98(0.97) & 0.96(0.95) & 0.93(0.92) & 0.92(0.85) & 0.84(0.84) & 0.8(0.79) \\
		& LR-FFS(equal weight) & 0.83(0.8) & 0.85(0.83) & 0.75(0.79) & 0.73(0.71) & 0.76(0.64) & 0.58(0.57) & 0.54(0.54) \\
		\midrule
		\multirow{6}[2]{*}{FDR$\downarrow$} & LR-FFS & 0.43(0.43) & 0.46(0.39) & \textbf{0.37(0.44)} & 0.47(0.46) & \textbf{0.45(0.4)} & 0.42(0.49) & \textbf{0.44(0.46)} \\
		& LR-FFS(CRU weight) & \textbf{0.42(0.43)} & 0.43(0.46) & 0.39(0.49) & 0.52(0.49) & 0.46(0.47) & 0.46(0.47) & 0.5(0.54) \\
		& LR-FFS-PAIR & 0.43(0.39) & 0.46(0.44) & 0.42(0.47) & \textbf{0.47(0.44)} & 0.55(0.49) & 0.55(0.61) & 0.69(0.74) \\
		& LR-FFS(CAVS weight) & 0.47(0.43) & 0.44(0.5) & 0.47(0.57) & 0.56(0.51) & 0.5(0.52) & 0.52(0.49) & 0.52(0.56) \\
		& LR-FFS(MV-SIS weight) & 0.44(0.43) & \textbf{0.43(0.45)} & 0.39(0.51) & 0.52(0.46) & 0.47(0.43) & \textbf{0.4(0.45)} & 0.45(0.5) \\
		& LR-FFS(equal weight) & 0.46(0.43) & 0.44(0.5) & 0.47(0.57) & 0.56(0.51) & 0.5(0.52) & 0.53(0.48) & 0.52(0.56) \\
		\midrule
		\multirow{6}[2]{*}{Size} & LR-FFS & 17.02(20.14) & 19.34(16.36) & 16.34(16.94) & 19.56(18.12) & 17.84(16.16) & 17.48(19.48) & 15.52(16.34) \\
		& LR-FFS(CRU weight) & 17.5(17.36) & 17.46(19.26) & 15.22(18.7) & 20.38(18.38) & 17.12(18.98) & 14.64(17.12) & 14.52(17.92) \\
		& LR-FFS-PAIR & 17.08(15.86) & 19.24(18.78) & 15.58(17.78) & 17.06(14.5) & 17.16(16.92) & 14.64(16.92) & 10.76(14.22) \\
		& LR-FFS(CAVS weight) & 16.46(15.48) & 15.46(18.44) & 15.62(19.34) & 17.58(16.34) & 17.24(16.04) & 14.48(11.86) & 13.4(14.94) \\
		& LR-FFS(MV-SIS weight) & 17.88(17.52) & 17.24(18.96) & 15.82(20.22) & 20.1(18.18) & 18.2(17.32) & 14.32(16.16) & 14.8(17.32) \\
		& LR-FFS(equal weight) & 16.26(15.56) & 15.56(18.42) & 15.66(19.34) & 17.3(16.18) & 17.3(16.34) & 14.5(11.92) & 13.72(14.96) \\
		\midrule
		\multirow{6}[2]{*}{wRank$\downarrow$} & LR-FFS & \textbf{11.6(11.78)} & \textbf{11.8(8.66)} & \textbf{11.5(12.56)} & \textbf{17.26(30.38)} & \textbf{23.72(87.36)} & \textbf{37.96(36.76)} & \textbf{51.84(125.2)} \\
		& LR-FFS(CRU weight) & 15.52(20.46) & 12.94(13.38) & 24.44(29.5) & 68.58(98.82) & 92.5(231.76) & 302.58(133.26) & 321.76(243.3) \\
		& LR-FFS-PAIR & 16.92(31.46) & 15.76(17.36) & 46.2(40.8) & 111.82(121.26) & 225.54(453.18) & 461.32(485.64) & 2037.18(2418.8) \\
		& LR-FFS(CAVS weight) & 75.96(76.34) & 49.92(63.94) & 95.74(115.6) & 121.72(238.08) & 154.48(403.78) & 381.5(281.76) & 435.12(316.28) \\
		& LR-FFS(MV-SIS weight) & 16.52(21.42) & 12.76(12.96) & 22.24(25.44) & 41.18(82.56) & 49.24(195.14) & 159.04(93.24) & 174.8(118.3) \\
		& LR-FFS(equal weight) & 75.1(75.5) & 50.3(64.44) & 94.94(115.84) & 128.9(241.1) & 160.06(404.48) & 403.54(276.8) & 428.8(312.52) \\
		\bottomrule
	\end{tabular}
	}\label{tab:table_extend}
\end{table}

\begin{example}\label{exp:compareCAVS}
	It is noted that the majority of screening methods can be viewed as weighted averages of utility values across different categories. In this example, we conduct simulations with the proposed methods for screening category-specific active predictors in a more complex setting within a non-distributed framework. The experimental setup is identical to that of Example 2 in \citet{xie_category-adaptive_2020}.
	The data are generated from the similar model as in Example \ref{exa:example2},  the $i$-th sample vector of predictors $\boldsymbol{X}_i$ is generated from a mixture distribution $0.9\tilde{\boldsymbol{X}}_i+0.1\boldsymbol{Z}$, where $\tilde{\boldsymbol{X}}_i =\boldsymbol{\mu }_r+\boldsymbol{\varepsilon}_i$, $\boldsymbol{\varepsilon}_i$ follows standard normal distribution and $\boldsymbol{Z}$ is a random vector with each component being independent Student’s $t$-distribution with 1 degree of freedom.
	
	 Set $R=5$ and $\boldsymbol{\mu}_1=\left(1.5,1.5, \mathbf{0}_{p-2}^{\top}\right)$, $\boldsymbol{\mu}_2=\left(\mathbf{0}_7^{\top}, 1.5,1.5,1.5, \mathbf{0}_{p-10}^{\top}\right)$, $ \boldsymbol{\mu}_3=\left(\mathbf{0}_3^{\top}, 1.5,1.5,1.5, \mathbf{0}_{p-6}^{\top}\right)$, $ \boldsymbol{\mu}_4=\left(\mathbf{0}_{15}^{\top}, 1.5,1.5,1.5,1.5,1.5, \mathbf{0}_{p-20}^{\top}\right)$,  $\boldsymbol{\mu}_5=\left(\mathbf{0}_{30}^{\top}, 1.5,1.5,1.5,1.5,1.5, \mathbf{0}_{p-35}^{\top}\right)$.  Accordingly, the true active sets are $\mathcal{A}_1=\left\{X_1, X_2\right\}$, $ \mathcal{A}_2=\left\{X_8, X_9, X_{10}\right\}$, $\mathcal{A}_3=\left\{X_4, X_5, X_6\right\}$, $\mathcal{A}_4=\left\{X_{16}, X_{17}, X_{18}, X_{19}, X_{20}\right\}$, and $\mathcal{A}_5=\left\{X_{31}, X_{32}, X_{33}, X_{34}, X_{35}\right\}$, respectively. We consider the balanced and imbalanced design as follows:

     \begin{itemize}
         \item [Case 1]: $p_i=0.2, i=1, \ldots, 5, n=200, p=1000$ or $3000$;
         \item [Case 2]: $p_1=p_2=p_3=0.1, p_4=p_5=0.35, n=200$, $p=1000$ or $3000$.
     \end{itemize}

	In \citet{xie_category-adaptive_2020}, CAVS was compared with a modified version of MV-SIS, KF, and PSIS, demonstrating its optimal performance among them. In this paper, we compare the proposed LR-FFS with a modified version of CRU and CAVS, denoted by $CRU_r$ and $CAVS_r$, respectively. 
	
	The simulation experiments were repeated 400 times, and the median, IQR(interquartile range), mean, and standard deviation of the rank of the least correlated feature in the analysis (The indicator ``mean'' refers to the previously reported wRank) over these repetitions were reported. Additionally, we reported $R_a$, the average of the ranks of all active predictors among all candidate variables sorted by the screening procedure; and $P_a$, the proportion of all active predictors being selected into the submodel with size $\lfloor n/\log(n)\rfloor$. The simulation results are shown in Table \ref{tab:compare with CAVS}.
\end{example}

\begin{table}[H]
	\centering
	\caption{Screening results with different methods for Example \ref{exp:compareCAVS}.}
	\resizebox{\textwidth}{!}{
	\begin{tabular}{cccccccccccccc}
		\toprule
		\multirow{2}[4]{*}{Method} & \multicolumn{6}{c}{$p=1000$}                    &       & \multicolumn{6}{c}{$p=3000$} \\
		\cmidrule{2-7}\cmidrule{9-14}          & Median$\downarrow$ & IQR $\downarrow$  & Mean$\downarrow$  & SD$\downarrow$    &$R_a\uparrow$    & $P_a\uparrow$    &       & Median $\downarrow$& IQR$\downarrow$   & Mean$\downarrow$  & SD$\downarrow$    & $R_a\uparrow$    & $P_a\uparrow$ \\
		\midrule
		\multicolumn{14}{c}{Case 1} \\
		$LR-FFS_1$  & 2     & 0     & 2.00  & 0.00  & 1.50  & 1.00  &       & 2     & 0     & 2.00  & 0.05  & 1.50  & 1.00 \\
		$CRU_1$  & 2     & 0     & 2.00  & 0.00  & 1.50  & 1.00  &       & 2     & 0     & 2.00  & 0.05  & 1.50  & 1.00 \\
  $CAVS_1$ & 2     & 0     & 2.00  & 0.00  & 1.50  & 1.00  &       & 2     & 0     & 2.01  & 0.11  & 1.51  & 1.00 \\
		
		$LR-FFS_2$  & 3     & 0     & 3.01  & 0.07  & 2.00  & 1.00  &       & 3     & 0     & 3.02  & 0.18  & 2.01  & 1.00 \\
		$CRU_2$  & 3     & 0     & 3.01  & 0.07  & 2.00  & 1.00  &       & 3     & 0     & 3.02  & 0.18  & 2.01  & 1.00 \\
  $CAVS_2$ & 3     & 0     & 3.01  & 0.07  & 2.00  & 1.00  &       & 3     & 0     & 3.03  & 0.37  & 2.01  & 1.00 \\
		$LR-FFS_3$  & 3     & 0     & 3.01  & 0.07  & 2.00  & 1.00  &       & 3     & 0     & 3.01  & 0.11  & 2.01  & 1.00 \\
		$CRU_3$  & 3     & 0     & 3.01  & 0.07  & 2.00  & 1.00  &       & 3     & 0     & 3.01  & 0.11  & 2.01  & 1.00 \\
  $CAVS_3$ & 3     & 0     & 3.01  & 0.10  & 2.00  & 1.00  &       & 3     & 0     & 3.02  & 0.12  & 2.01  & 1.00 \\
		$LR-FFS_4$  & 5     & 0     & 5.00  & 0.00  & 3.00  & 1.00  &       & 5     & 0     & 5.02  & 0.18  & 3.01  & 1.00 \\
		$CRU_4$  & 5     & 0     & 5.00  & 0.00  & 3.00  & 1.00  &       & 5     & 0     & 5.02  & 0.18  & 3.01  & 1.00 \\
  $CAVS_4$ & 5     & 0     & 5.00  & 0.00  & 3.00  & 1.00  &       & 5     & 0     & 5.03  & 0.21  & 3.01  & 1.00 \\
		$LR-FFS_5$  & 5     & 0     & 5.02  & 0.26  & 3.01  & 1.00  &       & 5     & 0     & 5.04  & 0.44  & 3.01  & 1.00 \\
		$CRU_5$  & 5     & 0     & 5.02  & 0.26  & 3.01  & 1.00  &       & 5     & 0     & 5.04  & 0.44  & 3.01  & 1.00 \\
  $CAVS_5$ & 5     & 0     & 5.02  & 0.27  & 3.01  & 1.00  &       & 5     & 0     & 5.05  & 0.52  & 3.01  & 1.00 \\
		\midrule
		\multicolumn{14}{c}{Case 2} \\
		$LR-FFS_1$ & 2     & 0     & 3.10  & 6.86  & 2.05  & 1.00  &       & 2     & 0     & 4.26  & 12.68 & 2.69  & 1.00 \\
		$CRU_1$  & 2     & 0     & 3.09  & 6.84  & 2.05  & 1.00  &       & 2     & 0     & 4.25  & 12.68 & 2.69  & 1.00 \\
  $CAVS_1$ & 2     & 0     & 3.21  & 7.47  & 2.11  & 0.99  &       & 2     & 0     & 4.52  & 13.86 & 2.83  & 0.99 \\
		$LR-FFS_2$  & 3     & 1     & 5.00  & 7.56  & 2.72  & 1.00  &       & 3     & 1     & 8.41  & 24.08 & 3.87  & 0.99 \\
		$CRU_2$  & 3     & 1     & 5.00  & 7.57  & 2.72  & 1.00  &       & 3     & 1     & 8.41  & 24.08 & 3.87  & 0.99 \\
		$CAVS_2$ & 3     & 1     & 5.20  & 8.14  & 2.79  & 1.00  &       & 3     & 1     & 8.98  & 26.04 & 4.07  & 0.99 \\
		$LR-FFS_3$  & 3     & 0     & 4.94  & 10.17 & 2.69  & 1.00  &       & 3     & 1     & 9.64  & 32.90 & 4.36  & 0.99 \\
		$CRU_3$ & 3     & 0     & 4.93  & 10.12 & 2.68  & 1.00  &       & 3     & 1     & 9.64  & 32.93 & 4.36  & 0.99 \\
		$CAVS_3$ & 3     & 0     & 5.17  & 11.12 & 2.77  & 1.00  &       & 3     & 2     & 10.23 & 35.23 & 4.57  & 0.99 \\
		$LR-FFS_4$  & 5     & 0     & 5.03  & 0.18  & 3.01  & 1.00  &       & 5     & 0     & 5.02  & 0.14  & 3.00  & 1.00 \\
		$CRU_4$  & 5     & 0     & 5.03  & 0.18  & 3.01  & 1.00  &       & 5     & 0     & 5.02  & 0.14  & 3.00  & 1.00 \\
		$CAVS_4$ & 5     & 0     & 5.05  & 0.26  & 3.01  & 1.00  &       & 5     & 0     & 5.04  & 0.21  & 3.01  & 1.00 \\
		$LR-FFS_5$  & 5     & 0     & 5.02  & 0.19  & 3.01  & 1.00  &       & 5     & 0     & 5.04  & 0.24  & 3.01  & 1.00 \\
		$CRU_5$  & 5     & 0     & 5.02  & 0.19  & 3.01  & 1.00  &       & 5     & 0     & 5.04  & 0.24  & 3.01  & 1.00 \\
	$CAVS_5$ & 5     & 0     & 5.04  & 0.27  & 3.01  & 1.00  &       & 5     & 0     & 5.07  & 0.32  & 3.02  & 1.00 \\
		\bottomrule
	\end{tabular}}\label{tab:compare with CAVS}
\end{table}

From Table \ref{tab:compare with CAVS}, the simulation results for CAVS are consistent with those in \citet{xie_category-adaptive_2020}. We show that the LR-FFS method performs no worse than CAVS, illustrating that this method is also competitive in non-distributed scenarios.

\section{Algorithm}	\label{sec:alg}
\begin{algorithm}[H]
	\caption{Federated Feature Screening for PSIS}
	\label{alg:PSIS}
	\renewcommand{\algorithmicrequire}{\textbf{Input:}}
	\renewcommand{\algorithmicensure}{\textbf{Output:}}
	
	\begin{algorithmic}[1]
		\REQUIRE $\{(\boldsymbol{X}_i^l,Y_i^l)\}_{i=1}^{n_l}$ %%input
		\ENSURE  the estimated screening utilities $\{\omega_j\},j=1,\cdots,p$    %%output
		
		\FOR{each feature $j \in \{1,\cdots,p\}$ \textit{\textbf{in parallel}}}
		\FOR{each client $l \in \{1,\cdots,m\}$ \textit{\textbf{in parallel}}}
		\STATE Client $C_l$ does:
		\FOR{each category $r \in \{1,\cdots,R\}$}
		\STATE $\theta_{j,r,1}^l \leftarrow \sum_{i=1}^{n_l}I(Y_i^l = y_r) $
		\STATE $\theta_{j,r,2}^l \leftarrow \sum_{i=1}^{n_l}X^l_{ji}I(Y_i^l = y_r) $
		\ENDFOR
		\STATE upload$_{C_l \rightarrow S}$ $\{\theta_{j,r,1}^l,\theta_{j,r,2}^l\}$, $r=1,\cdots,R$
		\ENDFOR
		\STATE Central Server $S$ does:
		\FOR{each category $r \in \{1,\cdots,R\}$}
		\STATE $\theta_{j,r,1} \leftarrow \sum_{l=1}^m \theta_{j,r,1}^l $
		\STATE $\theta_{j,r,2} \leftarrow \sum_{l=1}^m \theta_{j,r,2}^l $
		\STATE $\theta_{j,r} \leftarrow \theta_{j,r,1}/\theta_{j,r,2}$
		\ENDFOR
		\STATE $\omega_{j} \leftarrow \max_r{\theta_{j,r}}-\min_r{\theta_{j,r}}$
		\ENDFOR
		\RETURN $\{\omega_{j}\}$, 	$j=1,\cdots,p$
	\end{algorithmic}
\end{algorithm}

\begin{algorithm}[H]
	\caption{Federated Feature Screening for CAVS}
	\label{alg:CAVS}
	\renewcommand{\algorithmicrequire}{\textbf{Input:}}
	\renewcommand{\algorithmicensure}{\textbf{Output:}}
	
	\begin{algorithmic}[1]
		\REQUIRE $\{(\boldsymbol{X}_i^l,Y_i^l)\}_{i=1}^{n_l}$ %%input
		\ENSURE  the estimated screening utilities $\{\omega_j\},j=1,\cdots,p$    %%output
		
		\FOR{each feature $j \in \{1,\cdots,p\}$ \textit{\textbf{in parallel}}}
		\FOR{each client $l \in \{1,\cdots,m\}$ \textit{\textbf{in parallel}}}
		\STATE Client $C_l$ does:
		\FOR{each category $r \in \{1,\cdots,R\}$}
		\STATE $numerator_r^l \leftarrow 0 $
		\STATE $percent_r^l \leftarrow \sum_{i=1}^{n_l}I(Y_{i}^l=y_r) $	
		\FOR{each sample $i_1 \in \{1,\cdots,n_l\}$}
		\STATE $numerator_r^l \leftarrow numerator_r^l +\sum_{i_2=1}^{n_l} I(Y_{i_1}^l=y_r)I(X_{j i_1}^l<X_{ji_2}^l)$
		\ENDFOR
		\ENDFOR
		\STATE $numerator_r^l \leftarrow numerator_r^l/[n_l(n_l-1)]$
		\STATE upload$_{C_l \rightarrow S}$ $\{numerator_r^l,percent_r^l,n_l\}$, $r=1,\cdots,R$
		\ENDFOR
		\STATE Central Server $S$ does:
		\STATE $\omega_{j} \leftarrow 0$
		\FOR{each category $r \in \{1,\cdots,R\}$}
		\STATE $\theta_{j,r} \leftarrow \sum_{l=1}^m (numerator_r^l\lfloor n_l/2 \rfloor)/\sum_{l=1}^m \lfloor n_l/2 \rfloor$
		\STATE $percent_{j,r} \leftarrow \sum_{l=1}^m percent_r^l/\sum_{l=1}^m n_l$
		\STATE $\omega_{j, r} \leftarrow |\theta_{j,r}/percent_{j,r}-1/2|$
		\STATE $\omega_{j} \leftarrow \max\{\omega_j,\omega_{j, r}\}$
		\ENDFOR
		\ENDFOR
		\RETURN $\{\omega_{j}\}$, 	$j=1,\cdots,p$
	\end{algorithmic}
\end{algorithm}

\begin{algorithm}[H]
	\caption{Federated Feature Screening for FKF}
	\label{alg:FKF}
	\renewcommand{\algorithmicrequire}{\textbf{Input:}}
	\renewcommand{\algorithmicensure}{\textbf{Output:}}
	
	\begin{algorithmic}[1]
		\REQUIRE $\{(\boldsymbol{X}_i^l,Y_i^l)\}_{i=1}^{n_l}$ %%input
		\ENSURE  the estimated screening utilities $\{\omega_j\},j=1,\cdots,p$    %%output
		
		\FOR{each feature $j \in \{1,\cdots,p\}$ \textit{\textbf{in parallel}}}
		\FOR{each client $l \in \{1,\cdots,m\}$ \textit{\textbf{in parallel}}}
		\STATE Client $C_l$ does:
		\FOR{each category $r \in \{1,\cdots,R\}$}
		\STATE vector $density_r^l \leftarrow 0_{n_l\times 1} $
		\FOR{each sample $i_1 \in \{1,\cdots,n_l\}$}
		\STATE $density_r^l(i_1) \leftarrow \sum_{i=1}^{n_l}I(X_{ji}^l<X_{j i_1}^l)I(Y_i^l=y_r) $	
		\ENDFOR
		\STATE $density_r^l \leftarrow density_r^l/\sum_{i}^{n_l}I(Y_i^l=y_r) $	
		\ENDFOR
		\STATE $\omega_{j}^l \leftarrow 0$
		\FOR{each category $r_1 \in \{1,\cdots,R\}$}
		\FOR{each category $r_2 \in \{1,\cdots,R\}$}
		\STATE $\omega_{j,r_1,r_2}^l \leftarrow \max_{i}{|density_{r_1}^l-density_{r_2}^l|}$
		\STATE $\omega_{j}^l \leftarrow \max_{i}\{\omega_{j}^l,\omega_{j,r_1,r_2}^l\}$
		\ENDFOR
		\ENDFOR
		\STATE upload$_{C_l \rightarrow S}$ $\{\omega_{j}^l,n_l\}$
		\ENDFOR
		\STATE Central Server $S$ does:
		\STATE $\omega_{j} \leftarrow \sum_{l=1}^m (n_l\omega_{j}^l)/\sum_{l=1}^m n_l$
		\ENDFOR
		\RETURN $\{\omega_{j}\}$, 	$j=1,\cdots,p$
	\end{algorithmic}
\end{algorithm}

Before introducing the distributed estimation algorithm for MV-SIS, we focus on the decomposition of the MV-SIS utility. From Equation \ref{equ:MV}, we find that $\mathbb{E}_{X_{j}^{\prime}}\left[F\left(X_{j}^{\prime}\right)^{2}\right] = \frac{1}{3}$. Therefore, only
\begin{align*}
\theta_{j, r, 1}&=\mathbb{E}_{X^{\prime}}\left[\mathbb{E}_{X_{j}, Y}\left(I\left(X_{j} \leq X_{j}^{\prime}, Y=y_{r}\right)\right)^{2}\right]\\
\theta_{j, r, 2}&=\mathbb{E}_{X_{j}^{\prime}}\left[\mathbb{E}_{X_{j}, Y}\left(I\left(X_{j} \leq X_{j}^{\prime}, Y=y_{r}\right)\right) \mathbb{E}_{X_{j}}\left(I\left(X_{j} \leq X_{j}^{\prime}\right)\right)\right].
\end{align*}need to be estimated.

The estimation method follows \citet{li_distributed_2020}, using the U-statistic.

\begin{algorithm}[H]
	\caption{Federated Feature Screening for MV-SIS}
	\label{alg:MV}
	\renewcommand{\algorithmicrequire}{\textbf{Input:}}
	\renewcommand{\algorithmicensure}{\textbf{Output:}}
	
	\begin{algorithmic}[1]
		\REQUIRE $\{(\boldsymbol{X}_i^l,Y_i^l)\}_{i=1}^{n_l}$ %%input
		\ENSURE  the estimated screening utilities $\{\omega_j\},j=1,\cdots,p$    %%output
		
		\FOR{each feature $j \in \{1,\cdots,p\}$ \textit{\textbf{in parallel}}}
		\FOR{each client $l \in \{1,\cdots,m\}$ \textit{\textbf{in parallel}}}
		\STATE Client $C_l$ does:
		\FOR{each category $r \in \{1,\cdots,R\}$}
		\STATE $percent_r^l \leftarrow \sum_{i=1}^{n_l}I(Y_{i}^l=y_r) $	
		\STATE $\theta_{j, r, 1}^l \leftarrow \sum_{i_1\ne i_2\ne i_3 \in \{1,\cdots,n_l\}}I(X_{j i_3}^l<X_{j i_1}^l)I(X_{j i_2}^l<X_{j i_1}^l)I(Y_{i_2}^l=y_r)I(Y_{i_3}^l=y_r)$
		\STATE $\theta_{j, r, 1}^l \leftarrow \theta_{j, r, 1}^l/[n_l(n_l-1)(n_l-2)]$	
		\STATE $\theta_{j, r, 2}^l \leftarrow \sum_{i_1\ne i_2\ne i_3 \in \{1,\cdots,n_l\}}I(X_{j i_3}^l<X_{j i_1}^l)I(X_{j i_2}^l<X_{j i_1}^l)I(Y_{i_2}^l=y_r)$
		\STATE $\theta_{j, r, 2}^l \leftarrow \theta_{j, r, 2}^l/[n_l(n_l-1)(n_l-2)]$	
		\ENDFOR
		\STATE upload$_{C_l \rightarrow S}$ $\{\theta_{j, r, 1}^l,\theta_{j, r, 2}^l,percent_r^l,n_l\}$, $r=1,\cdots,R$
		\ENDFOR
		\STATE Central Server $S$ does:
		\STATE $\omega_{j} \leftarrow 0$
		\FOR{each category $r \in \{1,\cdots,R\}$}
		\STATE $\theta_{j, r, 1} \leftarrow \sum_{l=1}^m (\theta_{j, r, 1}^l n_l)/\sum_{l=1}^m n_l$
		\STATE $\theta_{j, r, 2} \leftarrow \sum_{l=1}^m (\theta_{j, r, 2}^l n_l)/\sum_{l=1}^m n_l$
		\STATE $percent_{j,r} \leftarrow \sum_{l=1}^m percent_r^l/\sum_{l=1}^m n_l$
		\STATE $\omega_{j} \leftarrow \omega_{j}+\theta_{j, r, 1}/percent_{j,r}-2\theta_{j, r, 2}+percent_{j,r}/3  $
		\ENDFOR
		\ENDFOR
		\RETURN $\{\omega_{j}\}$, 	$j=1,\cdots,p$
	\end{algorithmic}
\end{algorithm}

\vskip 0.2in
\bibliography{screening}

\end{document}